\documentclass[10pt,twocolumn,letterpaper]{article}

\usepackage[pagenumbers]{cvpr}
\usepackage[utf8]{inputenc}
\usepackage[T1]{fontenc}
\usepackage{amsmath,amsfonts,amssymb,bm}
\usepackage{booktabs}
\usepackage{multirow}
\usepackage{array}
\usepackage[export]{adjustbox}
\usepackage{graphicx}
\usepackage{caption}
\usepackage{subcaption}
\usepackage{xcolor}
\usepackage{soul}
\usepackage{bbding}
\usepackage{cuted}
\usepackage[colorlinks,linkcolor=red,citecolor=blue,urlcolor=blue]{hyperref}

\graphicspath{{figure/}{./}}

\newcommand{\rev}[1]{#1}
\newcommand{\revcolor}{}
\newcommand{\linelabel}[1]{}

\newcommand{\appendixtitle}[1]{%
  \clearpage
  \begin{center}
    #1
  \end{center}
  \vspace{0.5em}
}
\renewcommand{\paragraph}[1]{\vspace{.5em}\noindent\textbf{#1.}}

\title{SatSplat: Geometrically-Accurate Gaussian Splatting for Satellite Imagery}

\author{
Shuang Song\thanks{Equal contribution.}\quad
Jiyong Kim\footnotemark[1]\quad
Rongjun Qin\\
The Ohio State University\\
{\tt\small \{song.1634, qin.324\}@osu.edu}
}

\begin{document}
\maketitle

\begin{abstract}
High-resolution satellite imagery demands 3D reconstruction methods that deliver both speed and geometric accuracy. Recent adaptations of 3D Gaussian Splatting (3DGS) to satellite imagery demonstrate strong efficiency, but reconstruction quality often degrades under diverse illumination across multi-date, high-altitude acquisitions (with small intersection angles), limiting applicability to remote sensing and vision tasks. We present SatSplat, the first framework to adapt 2D Gaussian Splatting (2DGS) to satellite photogrammetry, with online camera adjustment. We approximate satellite cameras with an affine model and learn a minimal delta parameterization for in-splat camera refinement from dense observations. The formulation is implemented with a 2DGS scene representation. To handle time-varying shadows and illumination changes, we integrate geometric shadow mapping and per-camera color correction during training. Across the evaluated DFC2019 and IARPA2016 benchmark sites, SatSplat achieves strong geometric accuracy while significantly outperforming prior 3DGS-based baselines. On our processed DFC2019 benchmark, SatSplat reduces mean absolute error by 11.93\% and peak video memory by 31\% relative to the previous state of the art. Our approach enables large-scale digital surface modeling with practical computational efficiency. The project page is available at \url{https://gdaosu.github.io/satsplat/}.
\end{abstract}

\section{Introduction}
\label{sec:intro}

Commercial high-resolution satellites\footnote{\url{https://earth.esa.int/eogateway/missions/worldview}} and emerging very low-Earth orbit (VLEO) constellations are pushing space imaging toward decimeter-level detail\footnote{\url{https://albedo.com/}}. This amplifies a core need: converting large-scale, multi-date satellite imagery into accurate 3D reconstructions at scale. We target this gap with an approach that is both efficient and precise.

Neural Radiance Fields (NeRF)~\cite{mildenhall2021nerf} offered a compelling early answer~\cite{mariMultiDateEarthObservation2023,billouardSATNGPUnleashingNeural2024}. 3D Gaussian Splatting (3DGS)~\cite{kerbl3DGaussianSplatting2023} further broke the efficiency barrier~\cite{Aira_2025_CVPR}. Integrations of 3DGS with diffusion priors further improve rendering quality for satellite imagery~\cite{lee2025skyfallgssynthesizingimmersive3d}. However, as a novel view synthesis approach, 3DGS is designed for photorealistic rendering rather than strict geometry. Its primitive, semi-transparent volumetric blobs, are ill-suited for representing solid surfaces, reintroducing issues that NeRFs had largely overcome: noisy depth, inconsistent normals, floating artifacts, and geometric ambiguities. For photogrammetric applications where accurate Digital Surface Models (DSMs) are the target, this trade-off is unacceptable.

In this work, we introduce SatSplat, a framework for multi-date satellite photogrammetry. We adapt 2D Gaussian Splatting to affine cameras for a geometrically faithful surface representation and incorporate an online camera-adjustment module that refines initial poses from dense observations—critical when initial poses are imperfect. We integrate geometric shadow mapping to handle time-varying shadows and per-camera color correction to model illumination variations across dates. We also propose a geometry-aware initialization compatible with the rough DSM priors commonly available in satellite imaging. \rev{\linelabel{rev:scope}We evaluate the method on public benchmark groups from DFC2019 and the 2016 IARPA Multi-View Stereo 3D Mapping Challenge (IARPA2016)~\cite{boschMultipleViewStereo2016}, and frame our conclusions around that scope.}

\section{Related Work}
\label{sec:related_works}

\subsection{NeRF and 3DGS}
NeRF~\cite{mildenhall2021nerf} introduces continuous volumetric scene representations that enable high-quality novel view synthesis, but at the cost of long training and rendering times. 3DGS~\cite{kerbl3DGaussianSplatting2023} dramatically improves efficiency via explicit Gaussian primitives and differentiable rasterization, achieving real-time rendering; however, its volumetric blobs can limit surface fidelity. \rev{\linelabel{rev:illumination}For unconstrained photo collections, NeRF in the Wild~\cite{Martin-Brualla_2021_CVPR} extends NeRF with appearance embeddings to separate scene structure from transient effects, and WildGaussians~\cite{kulhanekWildGaussians2024} brings a related appearance-aware formulation to 3DGS for in-the-wild image collections. Recent work has also studied reconstruction under illumination-inconsistent inputs. GS-I$^3$~\cite{wangGSI32025} addresses surface reconstruction from illumination-inconsistent images in standard perspective-camera settings by combining tone correction and normal compensation. Compared with these generic in-the-wild or illumination-robust formulations, our setting is more constrained geometrically but more demanding radiometrically: we target multi-date satellite images with known acquisition geometry and sun direction, and therefore use an explicit shadow-aware image formation model under affine cameras rather than relying solely on implicit appearance embeddings or empirical tone corrections.}

\subsection{Adaptations to Satellite Multi-view Stereo}

Neural differentiable reconstruction methods have transformed 3D reconstruction from multi-view satellite imagery. Pioneering works including S-NeRF~\cite{derksenShadowNeuralRadiance2021}, Sat-NeRF~\cite{mariSatNeRFLearningMultiView2022}, Sat-NGP~\cite{billouardSATNGPUnleashingNeural2024}, and EO-NeRF~\cite{mariMultiDateEarthObservation2023} employ "true" multi-view approaches that iteratively optimize a neural radiance field. A common characteristic of these methods is the modeling of shadow generation to adapt to satellite images captured under different illumination conditions; shadow reasoning is also an active topic in high-resolution remote sensing analysis~\cite{PERS_001283217800007}. However, they all suffer from slow optimization and rendering speeds inherited from NeRF~\cite{mildenhall2021nerf} (ranging from hours to days). EOGS~\cite{Aira_2025_CVPR} represents the first effort to adapt 3DGS~\cite{kerbl3DGaussianSplatting2023} for satellite imagery, achieving quality comparable to EO-NeRF with a 300-fold speedup. Skyfall-GS~\cite{lee2025skyfallgssynthesizingimmersive3d} fuses satellite-derived coarse geometry with a diffusion prior to generate city-block-scale 3D scenes with improved cross-view consistency and realistic textures, all without 3D annotations.

For context, classical multi-view stereo methods process satellite images pairwise~\cite{qinRPCSTEREOPROCESSOR2016,defranchisAutomaticModularStereo2014,beyerAmesStereoPipeline2018}. These approaches typically follow a sequential pipeline: first performing bundle adjustment or pairwise geometry correction using keypoint matching algorithms (e.g., SIFT~\cite{loweDistinctiveImageFeatures2004}, AKAZE~\cite{alcantarilla2013fast}), then selecting a few image pairs—ideally in-track stereo pairs captured within minutes of each other—based on heuristic criteria. After applying pairwise stereo matching algorithms (e.g., SGM~\cite{hirschmullerStereoProcessingSemiglobal2008}, MGM~\cite{faccioloMGMSignificantlyMore2015}), a fusion step generates the final DSM. Related recent studies continue to evaluate dense image matching, point-cloud accuracy, and satellite-based building reconstruction, highlighting the importance of geometric quality in photogrammetric 3D products~\cite{PERS_001431099700004,PERS_001446480900003}. While classical methods are effective for large-scale 3D reconstruction from satellite images with abundant in-track stereo pairs, they face significant limitations when handling multi-date acquisitions where seasonal changes and varying illumination conditions are present.

\subsection{Gaussian Surface Representation}
A major limitation of 3DGS lies in its surface representation. Because 3D Gaussians are volumetric primitives, surfaces are modeled as collections of semi-transparent "blobs." The research community has recognized this issue and attempted to address it by introducing 2D primitives that align more closely with the underlying geometry. SuGaR~\cite{Guedon_2024_CVPR} introduced a regularization term during 3DGS optimization to encourage Gaussians to align with surfaces. NeuSG~\cite{chen2023neusg} jointly optimizes a 3DGS and an MLP-based signed distance function (SDF), subsequently extracting surfaces from the neural SDF—a computationally expensive approach. In contrast to these methods, 2DGS~\cite{huang2DGaussianSplatting2024} directly models oriented disks and introduces a perspective‑accurate splatting process based on analytic ray–splat intersection, yielding cleaner, more accurate surfaces. However, existing 2DGS formulations and implementations target pinhole cameras and are incompatible with satellite imaging.

Our SatSplat contributes an affine reformulation of 2DGS with an analytic affine ray-splat intersection and efficient CUDA rasterization, together with an online delta-affine camera adjustment and a DSM-aware initialization, forming an end-to-end pipeline tailored to multi-date satellite photogrammetry.

\section{The SatSplat Framework}

Our method, SatSplat, adapts and extends the 2D Gaussian Splatting framework for the specific challenges of satellite photogrammetry. We build upon the physically-based image formation model of EOGS while introducing a geometrically superior surface representation, a robust initialization strategy, and a fine-grained camera optimization module. %The implementation details are provided in \cref{supsec:implementation_details}.

\subsection{Preliminaries} \label{sec:preliminaries}
We first establish preliminaries for the rest of this section. In particular, we describe the image formation model for satellite imagery and the affine camera approximation used in our framework.

\textbf{Image Formation Model.} Following EOGS~\cite{Aira_2025_CVPR} and EO-NeRF~\cite{mariMultiDateEarthObservation2023}, we model the image $\mathbf{I} \in \mathbb{R}^{C\times H \times W }$ of a pixel as a combination of the surface albedo $\mathbf{A} \in \mathbb{R}^{C \times H \times W}$, the sun visibility mask $\mathbf{V} \in \mathbb{R}^{1\times H \times W}$, and a global ambient light component $\mathbf{L}_a \in \mathbb{R}^{C\times 1 \times 1}$. The sun visibility equals 1 if the point is lit by the sun and 0 if it is in shadow. To model radiometric variation and sensor characteristics across different acquisitions, $\mathbf{C} \in \mathbb{R}^{C\times C}$ and $\mathbf{c} \in \mathbb{R}^{C}$ represent per-image camera gain and bias. These parameters define a per-image affine color transform initialized to identity. All images are kept in linear color space, which is important for accurate photometric modeling and optimization; we avoid non-linear transforms such as gamma correction in data preparation. The final rendered color is given by \cref{eq:image_formation}.

\begin{equation} \mathbf{I} =  \underbrace{(\mathbf{C} \cdot \mathbf{A}  + \mathbf{c} )}_{\text{Corrected albedo}} \odot \underbrace{(\mathbf{V} + (1 - \mathbf{V}) \mathbf{L}_a)}_{\text{Sun-sky shading}}\label{eq:image_formation}\end{equation}
where $\odot$ denotes element-wise multiplication, and $C$ is the number of color channels (i.e., 3 for RGB), $H$ is the image height, and $W$ is the image width. Aside from the sun visibility mask $\mathbf{V}$, which is determined at render time, all other components are learned during training. \rev{\linelabel{rev:esm}Following EOGS, we do not use an unfiltered binary shadow test; instead, the shadow map is softened with the same exponential shadow-mapping style filtering used in EOGS to reduce spurious depth-induced artifacts near discontinuities.}

\textbf{Affine camera model.} The pinhole camera model (perspective projection) is widely used in computer vision and graphics. However, satellite images are typically accompanied by Rational Polynomial Coefficient (RPC) models, which define a complex, non-linear mapping from 3D world coordinates (latitude, longitude, height) to 2D image coordinates. A large body of work in classical photogrammetry has shown that for high-resolution satellite sensors with a narrow field of view, the RPC projection over a local area can be accurately approximated by a simpler affine transformation. This simplification is the foundation of the camera model in EOGS~\cite{Aira_2025_CVPR} and our work. An affine camera maps a normalized 3D world point $\mathbf{p} \in \mathbb{R}^3$ to a 2D image point $\mathbf{x} \in \mathbb{R}^2$ via a $3 \times 4$ matrix $\mathbf{F}$ as \cref{eq:affine_camera}. Similar to EOGS, we use normalized UTM coordinates as the world coordinate system, and map to 2D NDC space as the image coordinate system.

\begin{equation}
\mathbf{x} = \mathbf{F} \cdot \begin{bmatrix}\mathbf{p}\\1\end{bmatrix} . \label{eq:affine_camera}
\end{equation}

\textbf{3D Gaussian Splatting.} 3DGS~\cite{kerbl3DGaussianSplatting2023} presents a scene as a collection of anisotropic 3D Gaussians, which are volumetric primitives. Specifically, each 3D Gaussian is defined by a center position $\mathbf{p}_k$, a 3D covariance matrix $\Sigma_k$, an opacity $\alpha_k$, and a feature vector for color, typically represented by spherical harmonics (SH). Rendering is performed via a differentiable rasterization pipeline that projects the 3D Gaussians onto the image plane using the camera model. The covariance matrix $\Sigma_k$ is factorized into a scaling matrix $\mathbf{S}$ and a rotation matrix $\mathbf{R}$.

\textbf{2D Gaussian Splatting.} The key innovation in 2DGS is the perspective-correct splatting process, which computes the exact intersection of rays with the oriented disks, rather than simply projecting the center of the primitive~\cite{huang2DGaussianSplatting2024}. A 2D Gaussian is defined in a local tangent plane in world space, parameterized by its center position $\mathbf{p}_k$, two tangent vectors $\mathbf{t}_u$ and $\mathbf{t}_v$, two scaling factors $s_u$ and $s_v$ along the tangent directions, and a normal vector $\mathbf{n}$ as shown in \cref{eq:2dgs_splat_param}.

\begin{equation}
\begin{aligned}
P(u,v) &= \mathbf{p}_k + s_u \mathbf{t}_u u + s_v \mathbf{t}_v v\\
&= \mathbf{H} \left[ u \quad v \quad 1 \quad 1 \right]^\top
\label{eq:2dgs_splat_param}
\end{aligned}
\end{equation}

\begin{equation}
\mathrm{where}\quad \mathbf{H} = \begin{bmatrix}
            s_u \mathbf{t}_u & s_v \mathbf{t}_v & \mathbf{0} & \mathbf{p}_k \\
            0 & 0 & 0 & 1
\end{bmatrix} = \begin{bmatrix}
            \mathbf{RS} & \mathbf{p}_k \\
            \mathbf{0} & 1
\end{bmatrix}
\label{eq:2dgs_splat_H}
\end{equation}

Projecting a 2D Gaussian onto an image plane can be described by a general 2D-to-2D mapping in homogeneous coordinates. The screen-space points $(x,y)$ and the intersected depth $z$ are given by \cref{eq:2dgs_uv2screen_space}.

\begin{equation}
\begin{aligned}
\left[xz \quad yz \quad z\right]^\top &= \mathbf{W} P(u,v)\\
&= \mathbf{WH} \left[ u \quad v \quad 1 \quad 1 \right]^\top\label{eq:2dgs_uv2screen_space}
\end{aligned}
\end{equation}
where $\mathbf{W}$ is a $4 \times 4$ transformation from world coordinates to screen space. $\left[xz \quad yz \quad z\right]^\top$ are the homogeneous coordinates in screen space with perspective projection.

In the rasterization stage, the key is to find the inverse mapping from screen space to the splat's local 2D coordinates. However, directly inverting the mapping $ \mathbf{M} = (\mathbf{WH})^{-1}$ is impossible since $\mathbf{WH}$ is singular. Huang et al.~\cite{huang2DGaussianSplatting2024} derived an explicit ray–splat intersection method by parameterizing a pixel ray as the intersection of two orthogonal planes. The method insightfully leverages the properties of homogeneous coordinates and plane transformations~\cite{blinnHomogeneousFormulationLines1977}: transforming points on a plane using a transformation matrix $\mathbf{M}$ is equivalent to transforming the homogeneous plane parameters using the inverse transpose $\mathbf{M}^{-\top}$. Therefore, applying $\mathbf{M}=(\mathbf{WH})^{-1}$ is equivalent to $(\mathbf{WH})^{\top}$. This is the cornerstone of 2DGS rasterization.

We note that in the affine camera model, without perspective projection, the screen-space coordinates $(x,y)$ are decoupled from the depth $z$. In other words, the assumption of homogeneous coordinates no longer holds. In the following sections, we describe our proposed 2D Gaussian Splatting with affine cameras.

\subsection{Affine 2D Gaussian Splatting} \label{sec:affine_2dgs}
The core challenge in applying 2DGS to satellite imagery is adapting its rendering pipeline from pinhole to affine cameras. For 3DGS rendering, since it follows the projection–then–rasterization paradigm, switching to affine cameras is straightforward by replacing the camera projection matrix. However, 2DGS rasterization relies on computing the ray–splat intersection in a perspective projection setting as described in \cref{sec:preliminaries}, which is not directly applicable to affine cameras. Fortunately, due to the linearity of the affine camera, we can derive an analytic solution for the inverse mapping from screen space to the splat's local 2D coordinates.

\textit{Ray-splat intersection.} Analogous to the derivation of 2DGS, we define the transformation from local 2D splat coordinates $(u,v)$ to screen space $(x,y)$ in \cref{eq:affine2dgs_uv2screen}. Because the affine camera is a linear projection without perspective division, we separately define depth $z$ from the screen-space coordinates $(x,y)$.

\begin{equation}
\begin{aligned}
\begin{bmatrix}x & y\end{bmatrix}^\top &= \mathbf{B}_{xy} P(u,v) + \mathbf{b}_{xy} \\
z &= \mathbf{B}_z P(u,v) + b_z
\end{aligned}
\label{eq:affine2dgs_uv2screen}
\end{equation}

\begin{equation}
\mathrm{where}\quad \mathbf{F} = \left[ \begin{array}{c|c} \mathbf{B}_{xy} & \mathbf{b}_{xy} \\ \hline
\mathbf{B}_z & b_z \end{array} \right]
\label{eq:decompose_affine_camera}
\end{equation}

Expanding \cref{eq:affine2dgs_uv2screen} with \cref{eq:2dgs_splat_param} and \cref{eq:2dgs_splat_H}, we take the first two columns of $\mathbf{S}$ as $\mathbf{S'}$ since the last column is always zero.

\begin{equation}
\begin{aligned}
\begin{bmatrix} x & y \end{bmatrix}^\top &= \mathbf{B}_{xy} \mathbf{RS} \begin{bmatrix}u &v&1\end{bmatrix}^\top + \mathbf{B}_{xy}\mathbf{p}_k + \mathbf{b}_{xy}
\\
&= \mathbf{B}_{xy} \mathbf{RS'} \begin{bmatrix}u & v\end{bmatrix}^\top + \mathbf{B}_{xy}\mathbf{p}_k + \mathbf{b}_{xy}
\end{aligned}
\end{equation}
where $\mathbf{B}_{xy}$ is a $2 \times 3$ matrix and $\mathbf{RS'}$ is a $3 \times 2$ matrix. The product $\mathbf{B}_{xy} \mathbf{RS'}$ is a $2 \times 2$ matrix, which is invertible as long as the splat is not edge-on to the camera.

\begin{equation}
\begin{bmatrix}u & v\end{bmatrix}^\top = \mathbf{Q} \begin{bmatrix} x & y \end{bmatrix}^\top + \mathbf{q}
\label{eq:affine2dgs_screen2uv}
\end{equation}

\begin{equation}
\begin{aligned}
\mathrm{where}\quad \mathbf{Q} &= (\mathbf{B}_{xy} \mathbf{RS'})^{-1} \\
\mathbf{q} &= -(\mathbf{B}_{xy} \mathbf{RS'}) ^{-1} (\mathbf{B}_{xy}\mathbf{p}_k + \mathbf{b}_{xy})
\end{aligned}
\label{eq:affine2dgs_Qq}
\end{equation}

\textit{Depth function.} The depth value $z$ is important for visibility sorting during rasterization and is also used for surface-normal computation as supervision for geometric regularization of 2DGS orientation. Due to the object–sensor relationship in satellite imagery, we use negative height as the depth value, as in EOGS~\cite{Aira_2025_CVPR}. In practice, this depth function is effective since the inverse transformation from screen space to real-world coordinates is straightforward.

\begin{equation}
\mathbf{B}_z = \begin{bmatrix} 0 & 0 & -1 \end{bmatrix} \quad\quad b_z = \text{max\_altitude}    
\label{eq:depth_affine_camera}
\end{equation}

\textit{Degenerate Solutions.} $\mathbf{B}_{xy} \mathbf{RS'}$ becomes singular when the splat is observed from a slanted viewpoint and the projection is a line in screen space. Therefore, it might be missed during rasterization. To address this, we detect such degenerate cases by checking the determinant of $\mathbf{B}_{xy} \mathbf{RS'}$. If it is below a small threshold—analogous to 2DGS~\cite{huang2DGaussianSplatting2024}—we use a lower-bounded Gaussian

\begin{equation}
\hat{\mathcal{G}}(\mathbf{x}) = \max(\mathcal{G}(\mathbf{u}(\mathbf{x})), \mathcal{G}(\frac{\mathbf{x}-\mathbf{c}}{\sigma}))
\end{equation}

where $\mathbf{u}$ is given by \cref{eq:affine2dgs_screen2uv} and $\mathbf{c}$ is the splat center $\mathbf{p}_k$ projected in screen space. $\sigma$ is a small constant to control the minimum size of the Gaussian in screen space.

\textit{Normal Definition.} The 2D splats naturally define a surface normal $\mathbf{n}$ in world space, which is the last column of the rotation matrix $\mathbf{R}$. In rasterization, the normal value at each pixel is integrated with volumetric alpha blending~\cite{kerbl3DGaussianSplatting2023}. By computing the gradient of the depth map, we can also obtain an estimate of the surface normal $\mathbf{N}$. Adapting the normal-consistency loss from 2DGS~\cite{huang2DGaussianSplatting2024}, we minimize the discrepancy between the rendered normal map and the depth gradient to align the splats' orientations with the underlying surface geometry.

Specifically, $\mathbf{N}$ is computed with \cref{eq:surface_normal}, where $\mathbf{p}_s$ is the 3D point map backprojected from the depth map. We note that surface-normal computation must be in world coordinates, since an affine transformation does not guarantee orthogonality in screen space.

\begin{equation}
\mathbf{N} = \frac{\nabla_x \mathbf{p}_s \times \nabla_y \mathbf{p}_s}{\|\nabla_x \mathbf{p}_s \times \nabla_y \mathbf{p}_s\|}
\label{eq:surface_normal}
\end{equation}

\textit{Rasterization.} We follow the rasterization procedure in 3DGS and 2DGS. For each splat, we compute the bounding box in screen space. Then the Gaussians are sorted based on the depth of their centers and organized into tiles (16$\times$16 pixels by default). \rev{\linelabel{rev:sorting}Under the affine camera model, the sorting key is the explicit height-based depth in \cref{eq:depth_affine_camera}, which is stable across our near-nadir satellite views and remains strongly correlated with the true front-to-back order. As with other center-based ordering schemes, extremely large splats can still be imperfect in principle, but in practice this is limited because the optimized surfels remain compact and the depth variation within each primitive is small.} Finally, volumetric alpha blending composites all features as in \cref{eq:rasterization}.

\begin{equation}
\mathbf{f}(\mathbf{x}) = \sum_{i=1}^{N} \mathbf{f}_i \alpha_i \hat{\mathcal{G}}_i(\mathbf{u}(\mathbf{x})) \prod_{j=1}^{i-1} (1 - \alpha_j\hat{\mathcal{G}}_j(\mathbf{u}(\mathbf{x})))
\label{eq:rasterization}
\end{equation}
where $\mathbf{f}(\mathbf{x})$ denotes the rasterized features at pixel $\mathbf{x}$, and $\mathbf{f}$ includes color, normal, and depth. Index $i$ orders Gaussian primitives by depth, and $\alpha$ denotes opacity. In this work, we do not use spherical harmonics (SH) to model view-dependent appearance due to the extreme sparsity of multi-date satellite images; instead, we directly optimize per-splat color.

\subsection{Cascade Camera Optimization}\label{sec:camera_optimization}
Accurate geometry requires precise camera poses. In common practice (Sat-NeRF~\cite{mariSatNeRFLearningMultiView2022} / Sat-NGP~\cite{billouardSATNGPUnleashingNeural2024} and EO-NeRF~\cite{mariMultiDateEarthObservation2023}), camera poses are initialized from a bundle adjustment system~\cite{mariGenericBundleAdjustment2021}. While bundle adjustment relies on sparse feature correspondences, pose quality is often suboptimal for multi-date satellite images. Large temporal appearance changes and small intersection angles make high-quality sparse keypoint matching challenging, which directly affects pose accuracy.

To address this, dense supervision for camera pose can be beneficial~\cite{huang3RGSBestPractice2025}. However, applying dense pose optimization to satellite images is non-trivial due to: (1) the affine camera model has 12 degrees of freedom, which is more challenging than the 6-DoF pinhole model; and (2) in the EOGS~\cite{Aira_2025_CVPR} image formation model \cref{eq:image_formation}, the sun visibility mask $\mathbf{V}$ depends on the depth difference between depth rendered from the satellite direction and depth reprojected from the sun direction.

As elaborated in \cref{sec:affine_2dgs}, the depth value is disentangled from image coordinates in the affine camera model. The depth function can be fixed since it does not depend on camera intrinsics. Therefore, we reduce camera optimization to a $2\times 4$ delta affine matrix $\Delta \mathbf{F}_{xy}$. 

\begin{figure}[htp]
\centering
\includegraphics[width=0.8\linewidth]{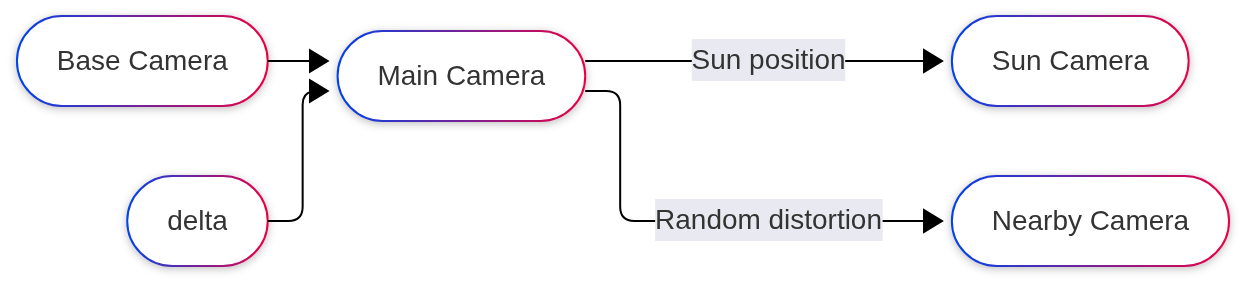}
\caption{Cascade camera update process. Each camera is a $3\times 4$ affine matrix. The sun position is obtained from metadata, and a nearby camera is randomly sampled around the primary camera.}
\label{fig:cascade_camera_update}
\end{figure}

% The initial affine camera parameters derived from RPC metadata contain small but significant errors due to inaccuracies in the satellite's recorded position and attitude. To correct for this, we introduce a camera optimization module. While one could optimize the full $3 \times 4$ affine matrix for each camera, this can be unstable. Instead, we opt for a more constrained approach inspired by bias-compensation methods in classical photogrammetry.We define the final camera projection Aj for camera j as the composition of the initial fixed camera matrix $\mathbf{F}_{xy}$ and a small, learnable delta transformation $\Delta \mathbf{F}_{xy}$:

\begin{equation}
    \mathbf{F}_{xy}' = \mathbf{F}_{xy} + \Delta \mathbf{F}_{xy}
\end{equation}
We parameterize $\Delta \mathbf{F}_{xy}$ as a $2\times 4$ matrix that represents a small 3D affine correction projected onto the 2D image plane, corresponding to the first two rows of $\mathbf{F}$. This formulation regularizes the optimization, preventing large deviations from the strong prior provided by the RPCs, while still allowing the fine-grained adjustments needed to align the images. This leads to significantly sharper reconstructions and better convergence.

\begin{figure}[htp]
    \centering
    \includegraphics[width=0.8\linewidth]{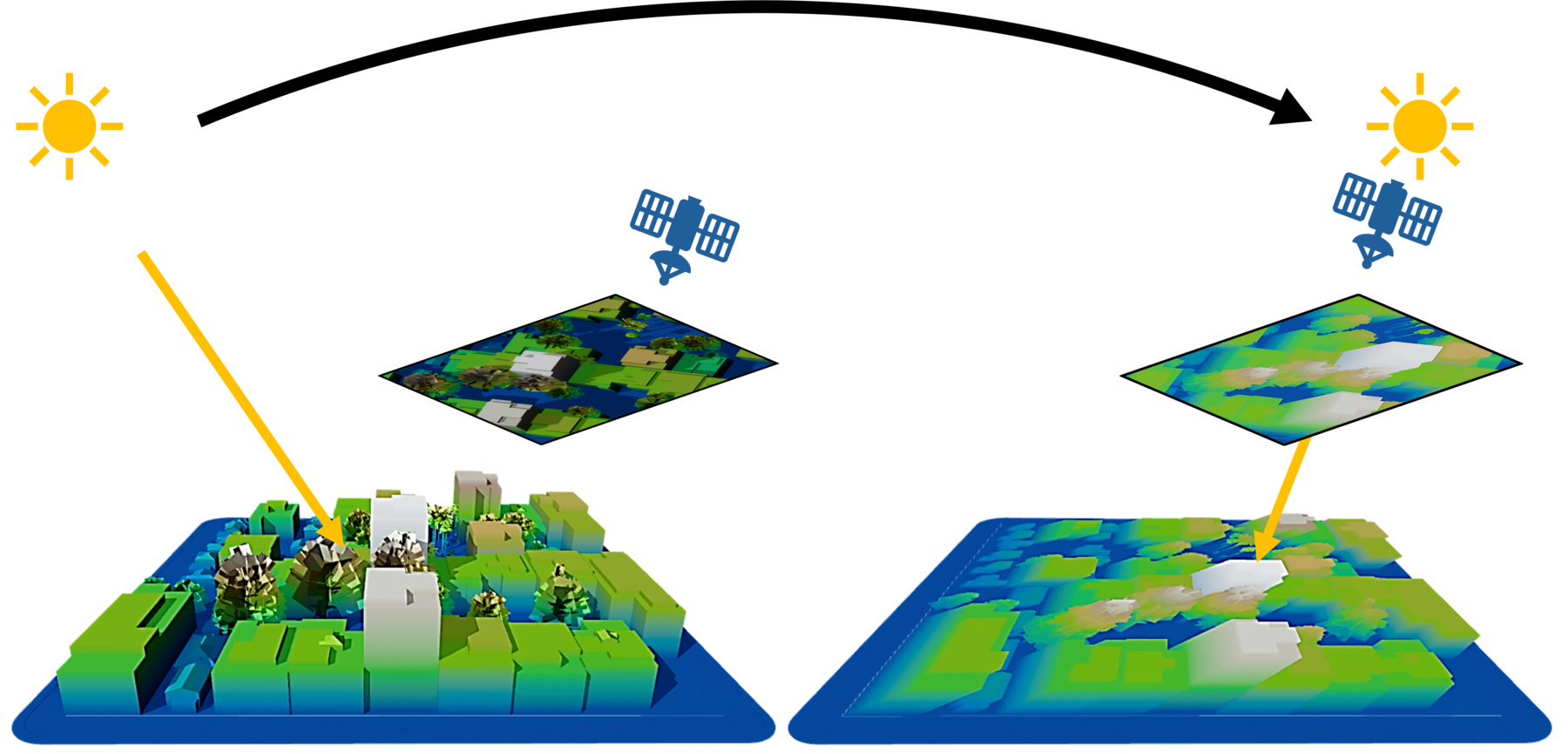}
\caption{Sun camera projection illustration. First, we project world points to the XOY plane along the sun direction, and then project the XOY plane to the image plane of the main camera.}
    \label{fig:sun_camera_formation}
\end{figure}

To generate the sun visibility mask $\mathbf{V}$, we need to render the depth from both the satellite camera and the sun direction. Unlike a pinhole camera where we can directly operate on the rotation matrix to change the viewing direction, the affine camera model does not have an explicit rotation component. Following EOGS~\cite{Aira_2025_CVPR}, we provide an intuitive way to acquire the sun camera pose. Changing depth along the viewing direction does not change the pixel location; therefore, the viewing direction is the nullspace of the matrix $\mathbf{B}_{xy}$ in \cref{eq:decompose_affine_camera}. We use a cross-product to find the direction vector $\mathbf{d}_{main} = \mathbf{B}_{x} \times \mathbf{B}_{y}$. For a sun camera with azimuth and elevation angles $(\phi, \theta)$, we flatten world points onto the XOY plane along the sun direction and then project with the main camera. This flattening would otherwise discard depth along the sun direction, so we preserve the original elevation as the sun depth (for shadow testing). The illustration is shown in \cref{fig:sun_camera_formation}. The sun camera's $\mathbf{B}_{xy}^{sun}$ is computed with \cref{eq:sun_camera_Bxy}, and the full derivation is provided in \cref*{supsec:sun_camera}.

\begin{equation}
\mathbf{B}_{xy}^{sun} 
%= \mathbf{B}_{xy} (\mathbf{I} - \left[ \mathbf{0} \quad \mathbf{0} \quad \mathbf{t} \right] ) = \mathbf{B}_{xy} \mathbf{T}
= \mathbf{B}_{xy} (\mathbf{I} - \begin{bmatrix}
    0 & 0 & \frac{\cos\phi}{\tan\theta} \\
    0 & 0 & \frac{\sin\phi}{\tan\theta} \\
    0 & 0 & 1
\end{bmatrix})
\label{eq:sun_camera_Bxy}
\end{equation}

Similarly, to generate a nearby camera, we compute the azimuth and elevation angle of the original camera viewing direction $\mathbf{d}_{main}$. We then add small random perturbations to the angles to create a new viewing direction. Then apply the same projection matrix transformation as \cref{eq:sun_camera_Bxy} to get the nearby camera's $\mathbf{B}_{xy}^{nearby}$.

% , we compute the sun position in world coordinates at a far distance along the sun direction from the satellite position. During training, for each primary camera, we randomly select a nearby secondary camera to form a stereo pair. The secondary camera's pose is also updated via its own delta affine matrix. We then render the depth maps from both cameras and compute the sun visibility mask $\mathbf{V}$ based on the depth difference along the sun direction. This dense geometric constraint provides strong supervision for optimizing the camera poses.

\subsection{Geometry-Aware Initialization}\label{sec:geometry_aware_initialization}
A good initialization is critical for fast convergence and high-quality results. Most Gaussian splatting methods utilize sparse point clouds from bundle adjustment~\cite{kerbl3DGaussianSplatting2023}, multi-view stereo~\cite{kotovenkoEDGSEliminatingDensification2025}, or random sampling~\cite{Aira_2025_CVPR}. Recent works have further investigated better initialization strategies based on monocular depth estimation or feed-forward 3D networks~\cite{huang3RGSBestPractice2025}. In this work, we propose a two-stage process that leverages a robust, open-source photogrammetry pipeline and a novel sampling strategy.

\textit{Coarse Initialization.} We first use ASP~\cite{sheanAutomatedOpensourcePipeline2016,beyerAmesStereoPipeline2018} to perform stereo reconstruction on pairs of input satellite images. ASP is a mature and widely-used tool in the remote sensing community that robustly produces a coarse DSM and a corresponding orthophoto map of the scene. This provides a dense 3D point cloud with initial color information.

\begin{figure}[htp]
    \centering
    \includegraphics[width=0.7\linewidth]{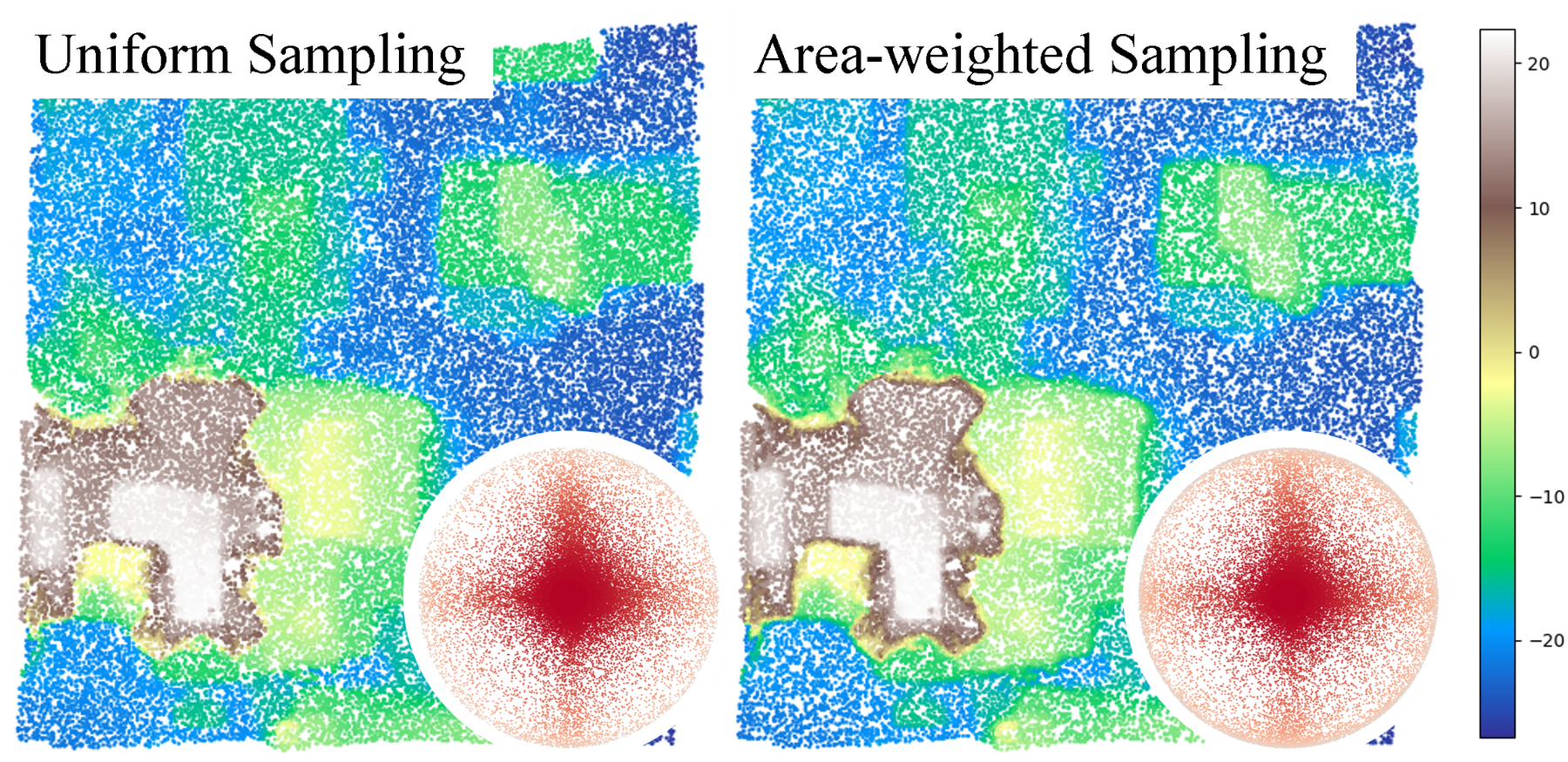}
    \caption{Point distribution with the same number of points. Normal distribution of splats are projected on a unit sphere at the right-bottom.}
    \label{fig:compare_sampling_methods}
\end{figure}

\textit{Area-Weighted Sampling.} Uniformly sampling points from the DSM to initialize our surfels would lead to a poor distribution of geometric primitives. Due to the predominantly nadir (top-down) viewing angles of satellites, flat surfaces such as roads and rooftops would be heavily oversampled, while geometrically important but less visible surfaces such as building facades would be undersampled. To address this, we introduce an area-weighted sampling strategy. For each pixel in the DSM, we estimate the local surface area to serve as a weight, which is $a = \| dx \times dy \|$ where $dx$ and $dy$ are the local tangent vectors computed from neighboring points. Due to the nature of 2.5D DSM, the tangent vectors can be easily computed via finite differences, guaranteeing that the cross product is valid. Intuitively, this weight reflects how much a small patch of surface contributes to the overall 3D structure. The area-based weight is then normalized as a probability distribution over all pixels. We sample splat centers according to this distribution, ensuring that larger surface areas (e.g., building facades) are more likely to be selected. This results in a more balanced and geometry-aware distribution of splats that better captures the scene's structure. 
\cref{fig:compare_sampling_methods} compares the splat distributions between uniform sampling and our method, demonstrating the effectiveness of our approach in capturing vertical structures.

\textit{Splat Orientation.} 
The initial orientation of each splat is set according to the local surface normal computed from the DSM. We construct an orthonormal basis for each splat where the normal vector $\mathbf{n}$ aligns with the splat's normal, and the tangent vectors $\mathbf{t}_u$ and $\mathbf{t}_v$ are derived via the Gram-Schmidt process to ensure orthogonality.

\textit{Splat Scaling.} We set the initial splat scales based on the average point spacing estimated by the kNN algorithm on the sampled points. This provides a reasonable starting size for the splats to cover the local surface without excessive overlap or gaps. During training, the splat scales are further optimized to better fit the underlying geometry.

\subsection{Implementation Details}
We build SatSplat based on the open-source framework gsplat~\cite{ye2025gsplat} and extend it to support the satellite image rendering pipeline as described in \cref{sec:preliminaries}. We implement the affine 2D Gaussian splatting module in CUDA for efficient rasterization. The cascade camera optimization is implemented in PyTorch and integrated into the training loop. The entire system is end-to-end differentiable, allowing joint optimization of splat parameters, camera deltas, and image formation components. In contrast to EOGS, which initializes from a dense volume and gradually prunes Gaussians, we found that the number of splats is difficult to estimate across different scenes. We directly sample from the DSM to initialize 2D Gaussians and use the 3DGS-MCMC~\cite{kheradmand3DGaussianSplatting} strategy to relocate Gaussians during training. 

\section{Experimental Evaluation}
We conduct a comprehensive set of experiments to validate the performance of SatSplat, focusing on its geometric accuracy. We evaluate on public benchmark groups from the 2019 IEEE GRSS Data Fusion Contest (DFC2019)~\cite{boschSemanticStereoIncidental2019}, including both JAX and OMA sites, \rev{as well as the 2016 IARPA Multi-View Stereo 3D Mapping Challenge (IARPA2016)~\cite{boschMultipleViewStereo2016}}. Each image covers an area of approximately 256$\times$256 meters, at a resolution of 30--50 cm per pixel. \rev{We report results by benchmark group rather than treating different tiles from the same benchmark as independent datasets.} \rev{\linelabel{rev:google_earth}We do not evaluate on the Google Earth benchmark used by Skyfall-GS because it does not provide the acquisition-time sun metadata required by our shadow-aware image formation model.} 
Detailed data preprocessing, optimization, and runtime settings are provided in \cref*{supsec:experiment_settings}.

\subsection{Accuracy Evaluation}

For the main experimental evaluation, we compare our method with several existing methods: s2p~\cite{defranchisAutomaticModularStereo2014,defranchisStereorectificationPushbroomImages2014,defranchisAutomaticSensorOrientation2014}, SAT-NGP~\cite{billouardSATNGPUnleashingNeural2024}, Skyfall-GS~\cite{lee2025skyfallgssynthesizingimmersive3d}, and EOGS~\cite{Aira_2025_CVPR}. Following common practice~\cite{mariSatNeRFLearningMultiView2022,billouardSATNGPUnleashingNeural2024,Aira_2025_CVPR} and established geospatial accuracy-assessment principles~\cite{PERS_001208095100005}, the metric used to evaluate geometric accuracy is the mean absolute error with registration ($\mathrm{MAE}_{reg}$), as defined in \cref{eq:maereg}. The registration process eliminates the systematic bias between the GCP-free (ground control point) bundle adjustment and the LiDAR ground truth DSM. To adapt the OMA dataset for the EOGS code, we modified the maximum height parameter in their CUDA implementation, changing it from 200 m to 1000 m for valid depth index sorting. This adjustment does not affect performance on the original JAX and IARPA datasets.

\begin{equation}
\mathrm{MAE}_{reg}=\min_{\mathbf{o}}\sum_{i=1}^{N}\frac{\left|\hat{M}(\hat{\mathbf{x}}_i)-o_z-M(\mathbf{x}_i)\right|}{N}
\label{eq:maereg}
\end{equation}
where $M$ is the ground truth DSM, $\hat{M}$ is the reconstructed DSM, $\mathbf{x}_i = (x_i, y_i)$ is a pixel location, $\hat{\mathbf{x}}_i$ is the offset location, and $\mathbf{o}=(o_x, o_y, o_z)$ is the offset.

\rev{\cref{tab:dataset_summary_mae} summarizes the dataset-level average $\mathrm{MAE}_{reg}$ on the JAX, OMA, and IARPA benchmark groups for both all classes and building-only regions, while the full per-site breakdown is provided in \cref*{tab:supp_jax_mae,tab:supp_oma_iarpa_mae}.}\ SatSplat achieves the best average accuracy on JAX and OMA for all classes, the best building-only accuracy on all three benchmark groups, and the second-best all-class average on IARPA. EOGS remains slightly stronger on the IARPA all-class average, but SatSplat is consistently competitive across benchmark groups. \rev{For Skyfall-GS~\cite{lee2025skyfallgssynthesizingimmersive3d}, their Iterative Dataset Update (IDU) is a key contribution that delivers compelling rendering quality but is computationally time-consuming. Furthermore, IDU degrades the final geometric accuracy; the reliance on an image diffusion prior during IDU introduces uncertainty and cross-view inconsistencies that perturb the underlying 3D geometry.}

As shown in \cref{fig:novel_view}, while EOGS suffers from artifacts like "floaters" and oversized Gaussians that degrade novel views, our SatSplat optimizes for a more compact Gaussian representation. This results in a high-fidelity model that is tightly anchored to the terrain surface, enabling consistent, high-quality rendering from any viewpoint. \cref{fig:ourdataset_comparison} presents the visual quality of SatSplat. Compared with other methods, it generates more detail and superior smoothness in the reconstructed surfaces. On the OMA-212 site, SatSplat best recovers the ring structures around the sphere. Extended errormap comparisons for all tiles are provided in \cref*{supsec:additional_results}.

\begin{table}[!ht]
\revcolor
\renewcommand{\baselinestretch}{1}\selectfont
\renewcommand{\arraystretch}{1}
\centering
\small
\caption{\rev{Dataset-level average $\mathrm{MAE}_{reg}$ (m; lower is better) across the JAX, OMA, and IARPA benchmark groups. Best and second-best results in each column are marked in bold and underline, respectively.}}
\label{tab:dataset_summary_mae}
\resizebox{\linewidth}{!}{
\begin{tabular}{lccccccc}
\toprule
& \multicolumn{3}{c}{\textbf{All Classes}} & \multicolumn{3}{c}{\textbf{Building Only}} & \multirow{2}{*}{\textbf{Time (min)}} \\
\cmidrule(lr){2-4}\cmidrule(lr){5-7}
\textbf{Method} & \textbf{JAX} & \textbf{OMA} & \textbf{IARPA} & \textbf{JAX} & \textbf{OMA} & \textbf{IARPA} & \\
\midrule
ASP & 2.09 & 1.00 & 2.38 & 1.88 & 2.12 & 1.90 & \textbf{2} \\
s2p & 2.98 & 1.30 & 3.28 & 3.45 & 2.13 & 3.20 & 20 \\
SAT-NGP & 3.01 & 2.01 & 2.37 & 3.40 & 3.99 & 2.70 & 23 \\
EOGS & \underline{1.50} & 1.47 & \textbf{1.86} & \underline{1.25} & 2.64 & \underline{1.37} & \underline{3} \\
Skyfall-GS (w/o IDU) & 1.78 & \underline{1.22} & 2.55 & 1.45 & \underline{1.69} & 2.17 & 30 \\
Skyfall-GS (w/ IDU) & 1.88 & 1.40 & 2.60 & 1.58 & 1.81 & 2.05 & 120 \\
SatSplat (Ours) & \textbf{1.33} & \textbf{0.97} & \underline{1.92} & \textbf{0.93} & \textbf{1.37} & \textbf{1.19} & 14 \\
\bottomrule
\end{tabular}
}
\end{table}

\begin{figure}[!htbp]
  \renewcommand{\baselinestretch}{1}\selectfont
  \renewcommand{\arraystretch}{1}
    \centering
    \begin{subfigure}[b]{0.48\linewidth}
        \includegraphics[width=\linewidth, trim=0 0 0 200, clip]{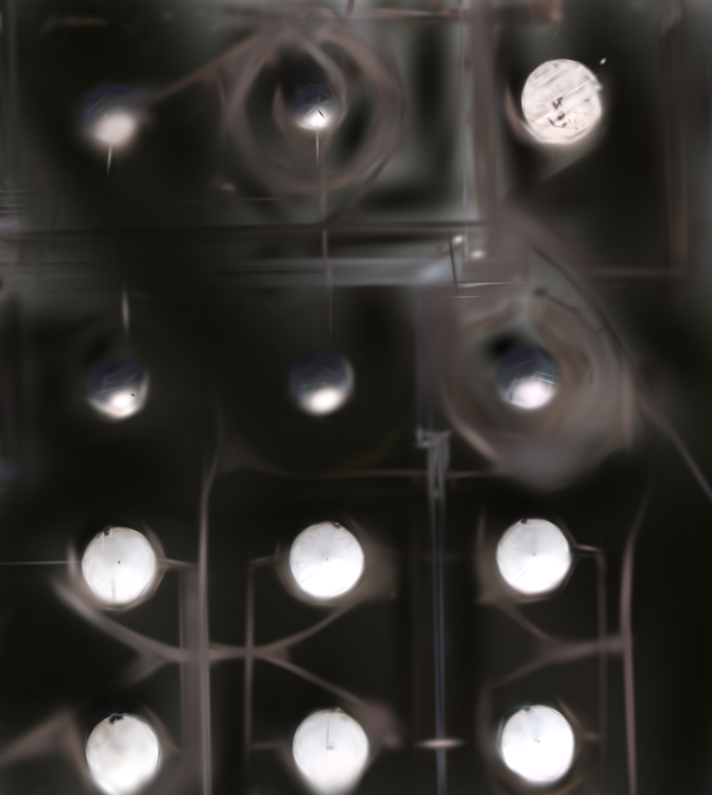}
        \caption{EOGS (Nadir)}
    \end{subfigure}
    \hfill
    \begin{subfigure}[b]{0.48\linewidth}
        \includegraphics[width=\linewidth, trim=0 0 0 200, clip]{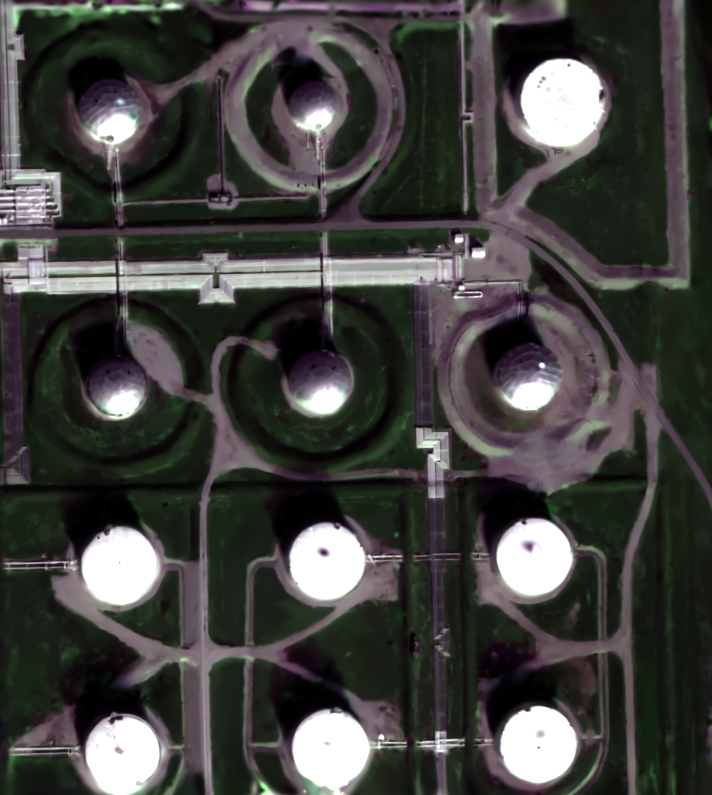}
        \caption{SatSplat (Nadir)}
    \end{subfigure}
    \\ \vspace{1mm}
    \begin{subfigure}[b]{0.48\linewidth}
        \includegraphics[width=\linewidth]{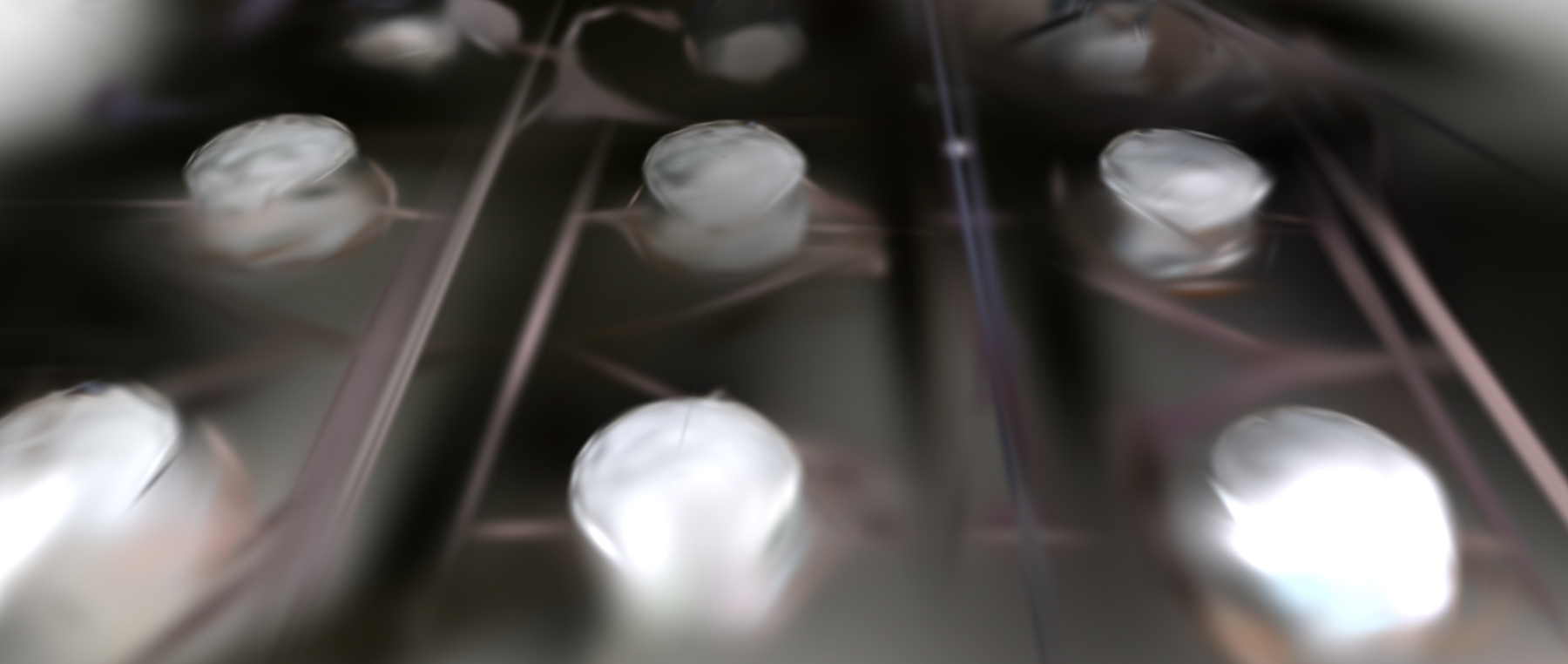}
        \caption{EOGS (Tilt)}
    \end{subfigure}
    \hfill
    \begin{subfigure}[b]{0.48\linewidth}
        \includegraphics[width=\linewidth]{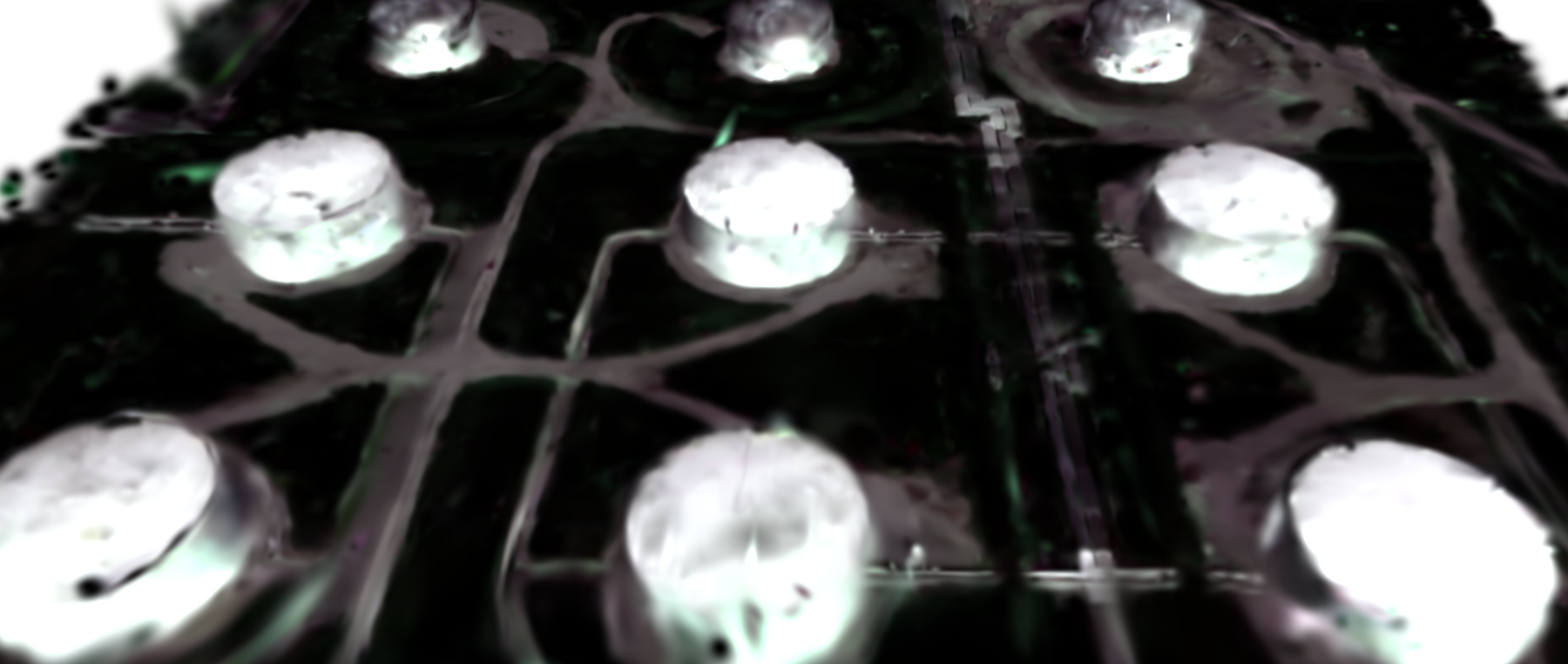}
        \caption{SatSplat (Tilt)}
    \end{subfigure}

    \caption{\rev{Novel view rendering comparison between EOGS and SatSplat from nadir and tilt perspectives. EOGS reconstruction is prone to artifacts, including oversized Gaussians and floaters that compromise visual quality. In contrast, SatSplat produces a cleaner reconstruction with compact Gaussians accurately anchored to the terrain surface, enabling high-fidelity rendering from varied viewpoints.}}
    \label{fig:novel_view}
\end{figure}

\begin{figure*}[!htbp]
\renewcommand{\baselinestretch}{1}\selectfont
\renewcommand{\arraystretch}{1}
\centering
\vspace{-10px}
\setlength{\tabcolsep}{0pt}
\begin{tabular}{@{}ccccccccc@{}}
% Row 1: JAX-168
\includegraphics[width=0.110\textwidth]{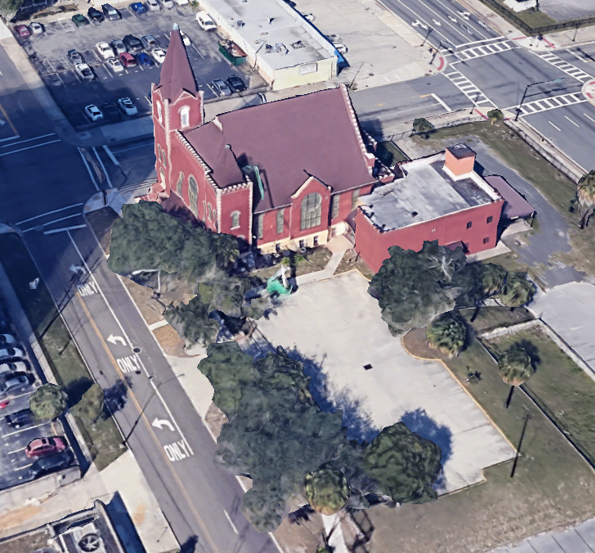} &
\includegraphics[width=0.110\textwidth]{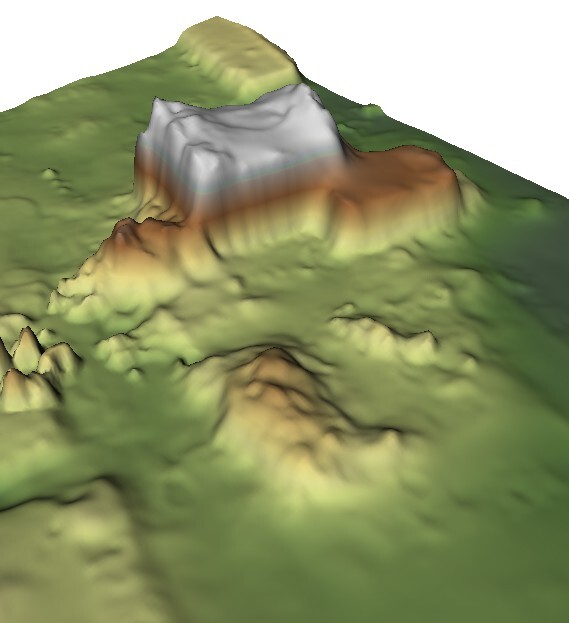} &
\includegraphics[width=0.110\textwidth]{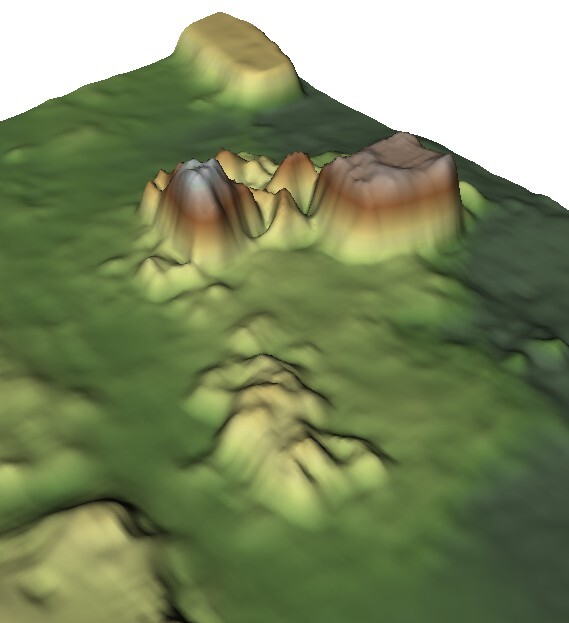} &
\includegraphics[width=0.110\textwidth]{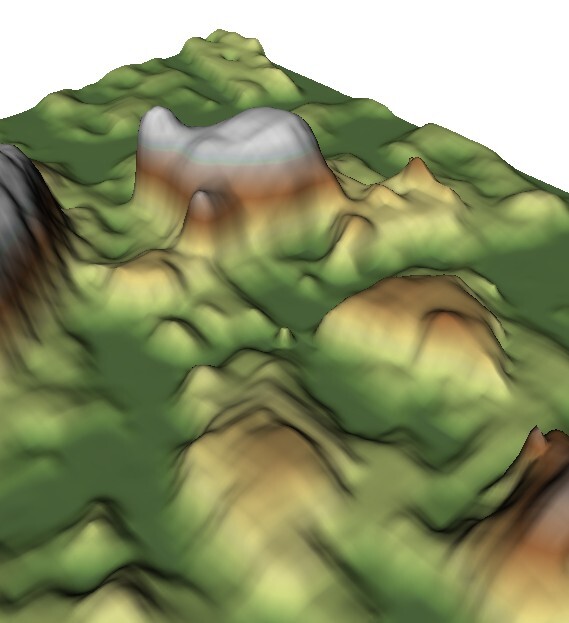} &
\includegraphics[width=0.110\textwidth]{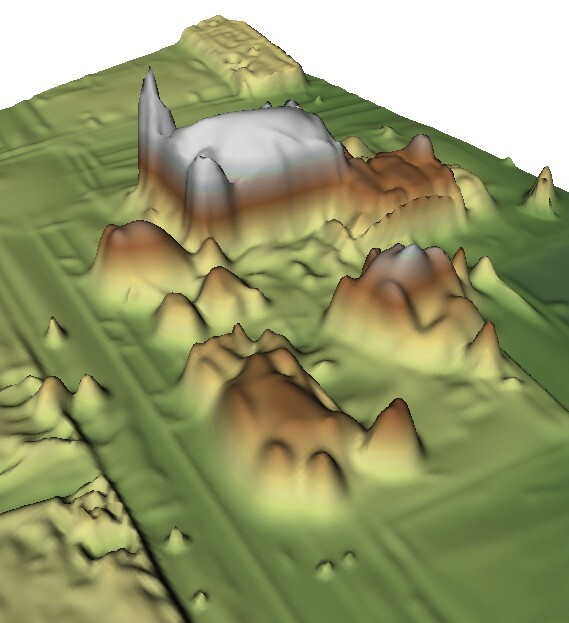} &
\includegraphics[width=0.110\textwidth]{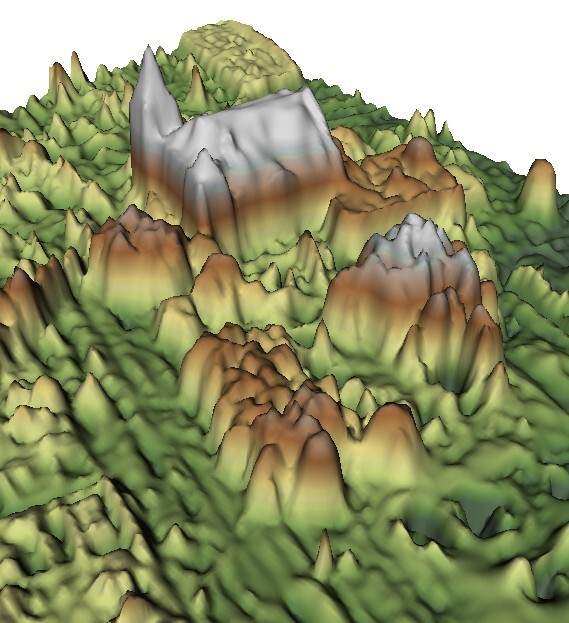} &
\includegraphics[width=0.110\textwidth]{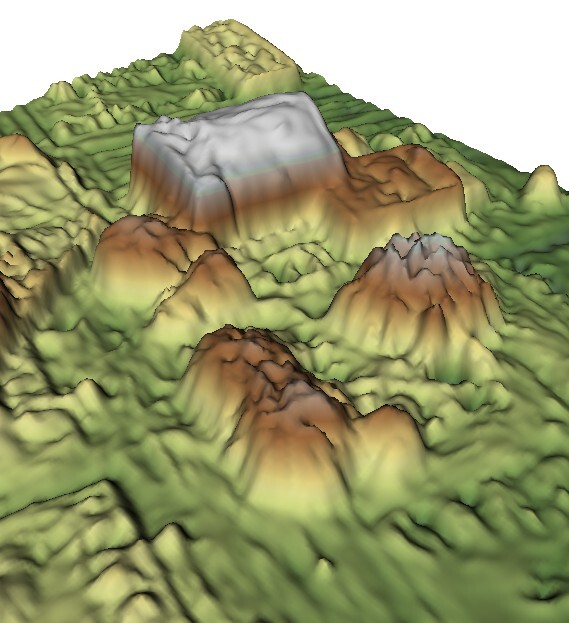} &
\includegraphics[width=0.110\textwidth]{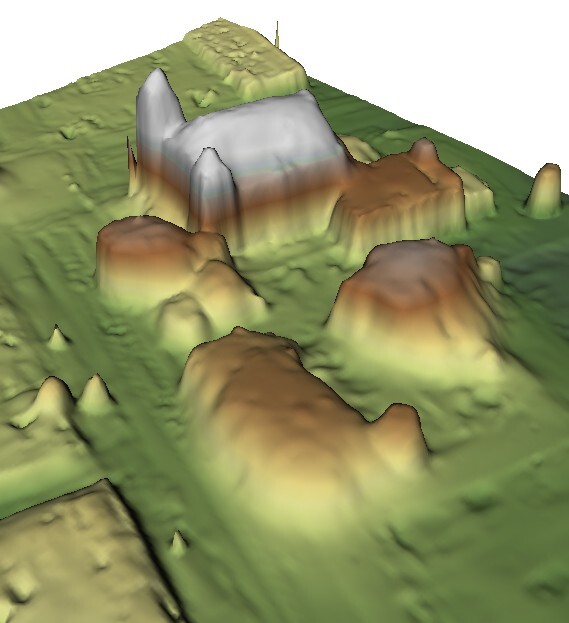} &
\includegraphics[width=0.110\textwidth]{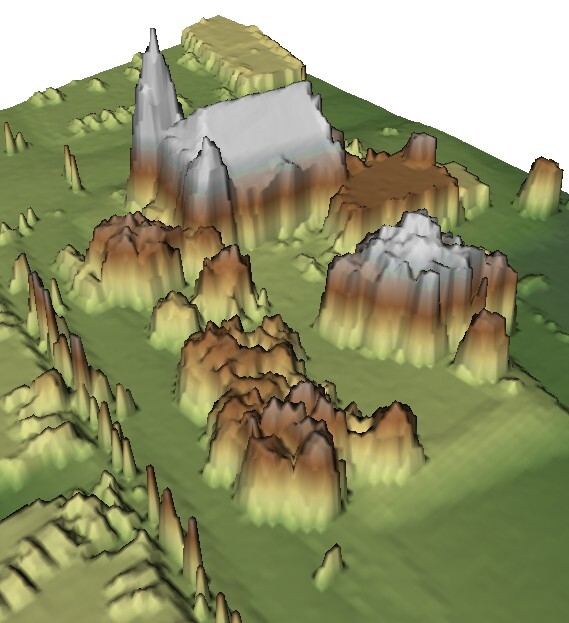} \\

% Row 2: OMA-212
\includegraphics[width=0.110\textwidth]{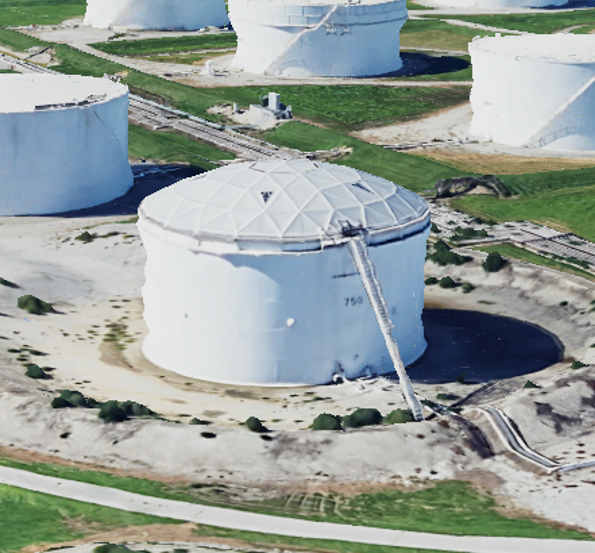} &
\includegraphics[width=0.110\textwidth]{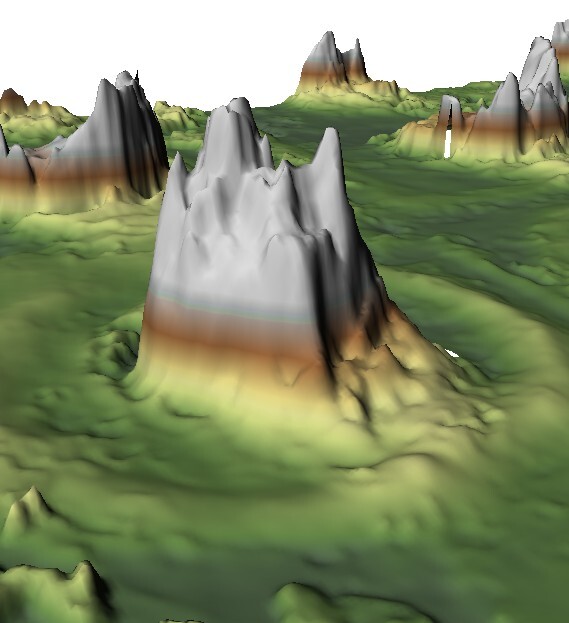} &
\includegraphics[width=0.110\textwidth]{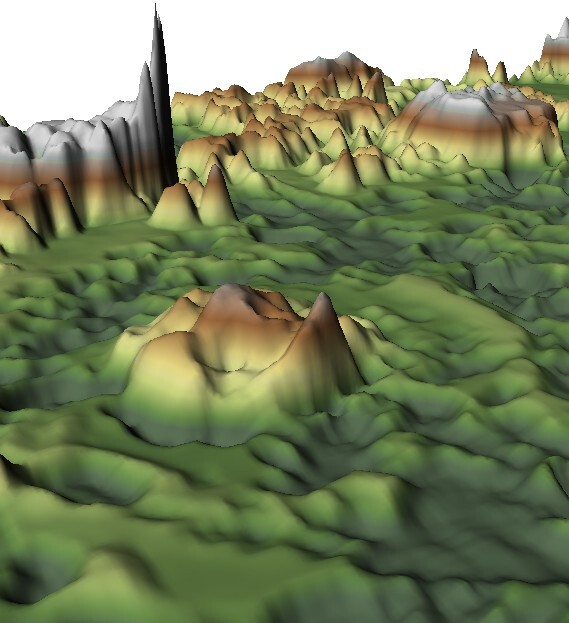} &
\includegraphics[width=0.110\textwidth]{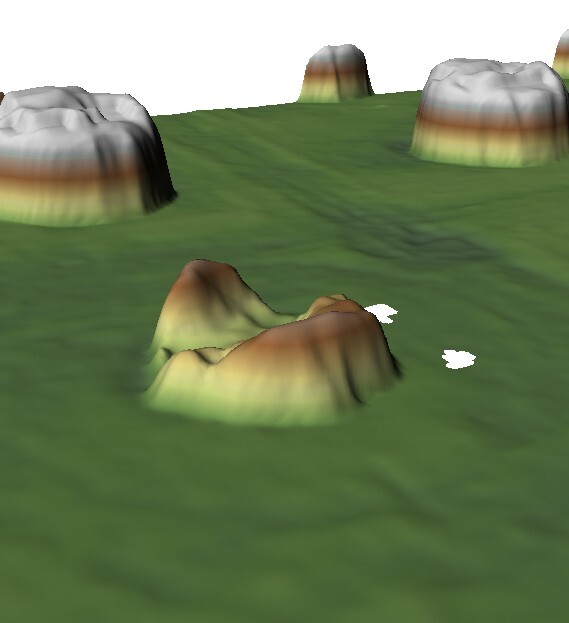} &
\includegraphics[width=0.110\textwidth]{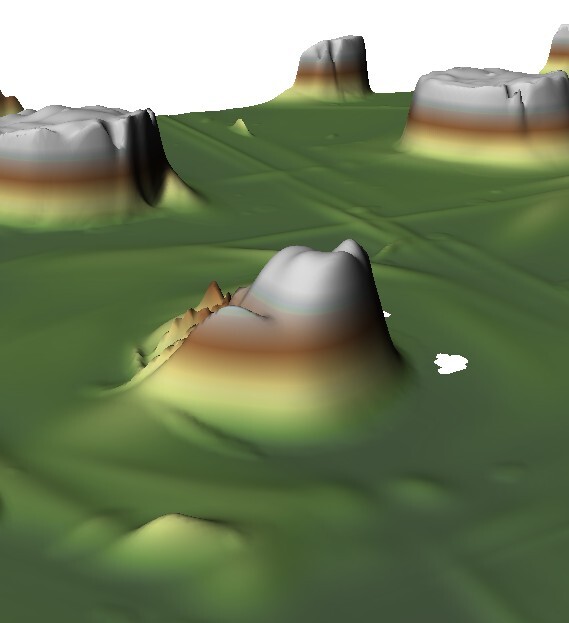} &
\includegraphics[width=0.110\textwidth]{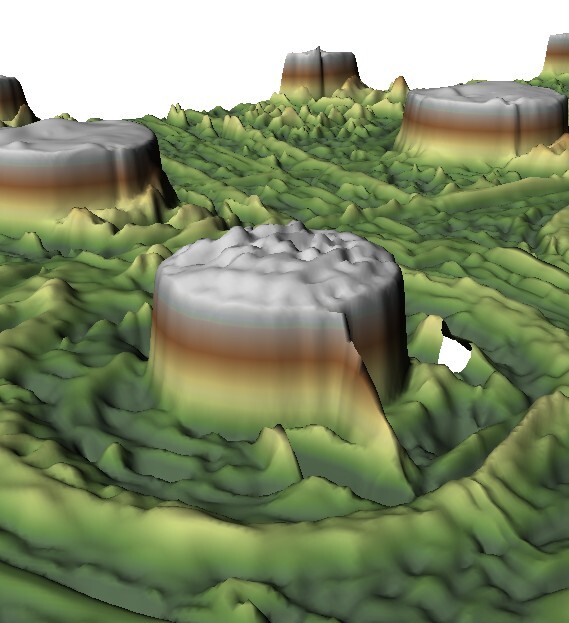} &
\includegraphics[width=0.110\textwidth]{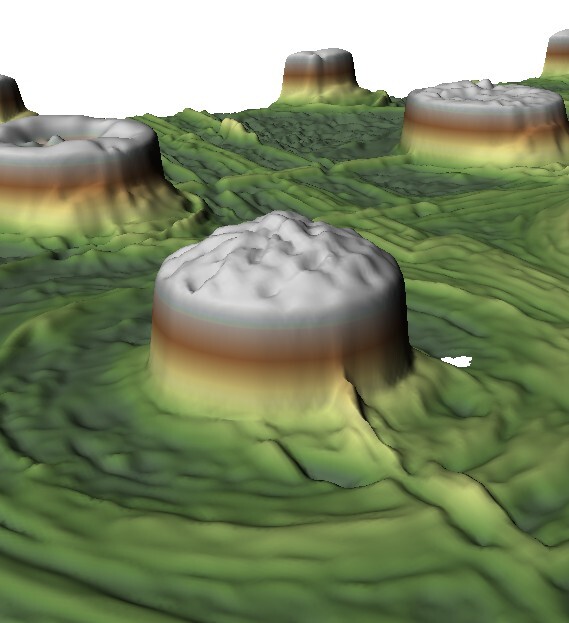} &
\includegraphics[width=0.110\textwidth]{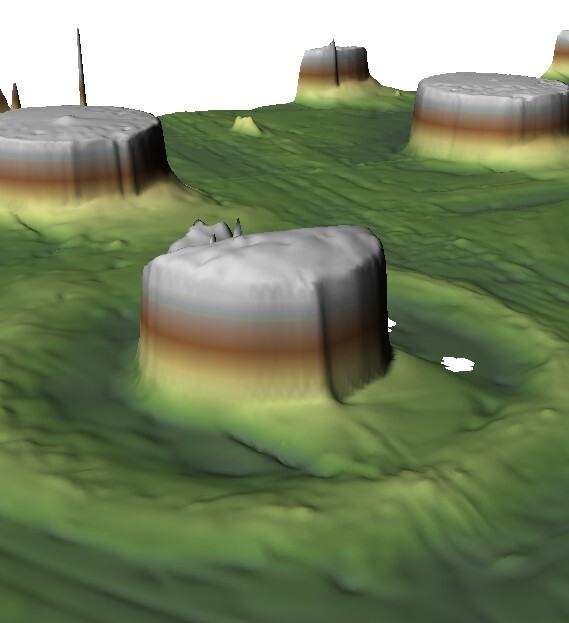} &
\includegraphics[width=0.110\textwidth]{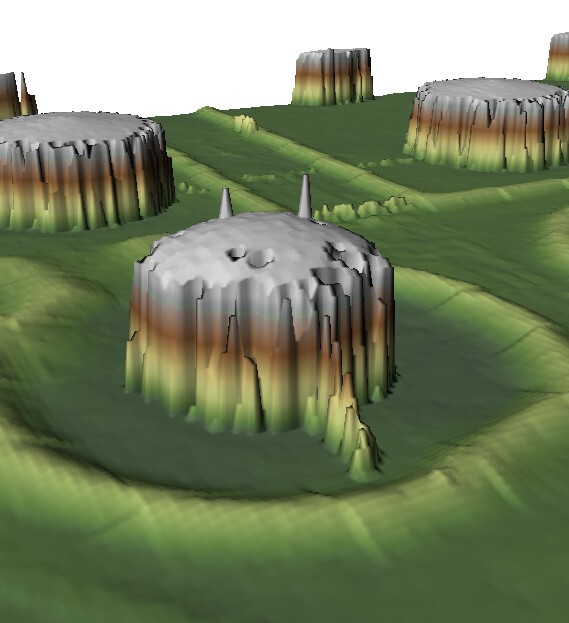} \\
% Row 3: OMA-315
\includegraphics[width=0.110\textwidth]{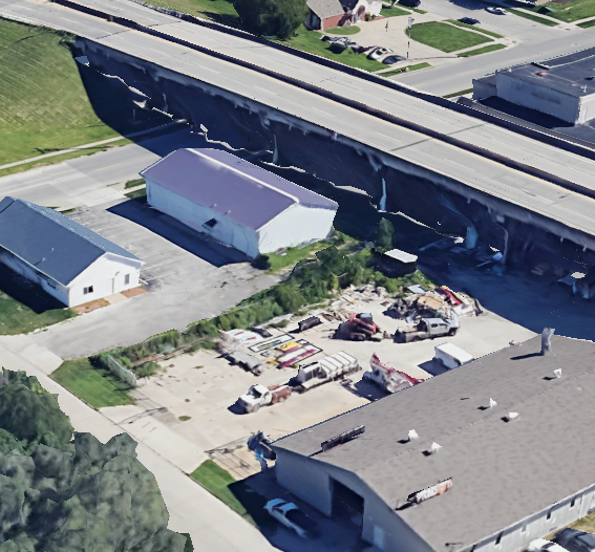} &
\includegraphics[width=0.110\textwidth]{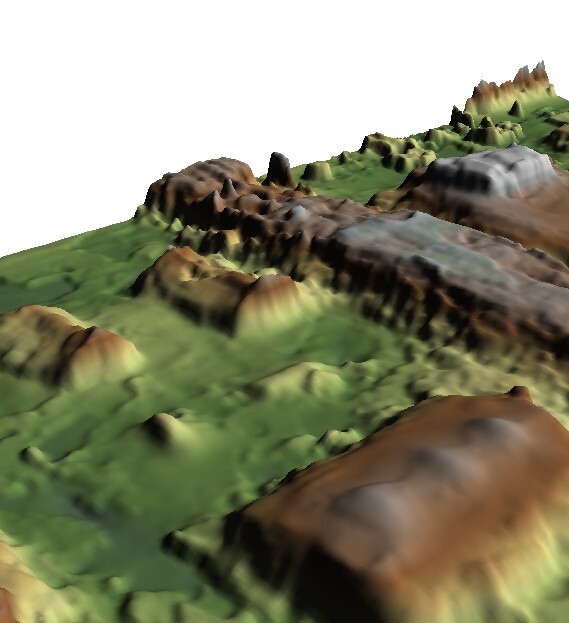} &
\includegraphics[width=0.110\textwidth]{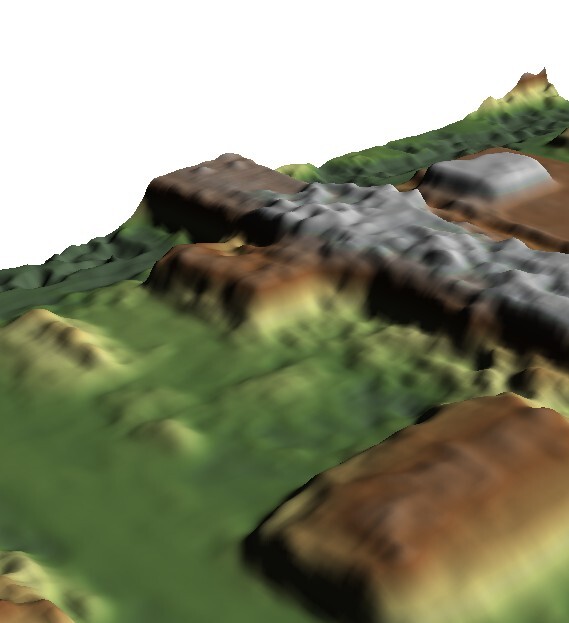} &
\includegraphics[width=0.110\textwidth]{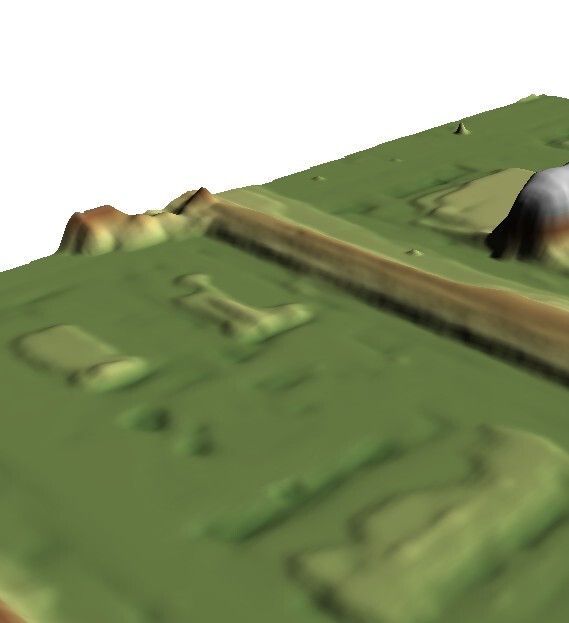} &
\includegraphics[width=0.110\textwidth]{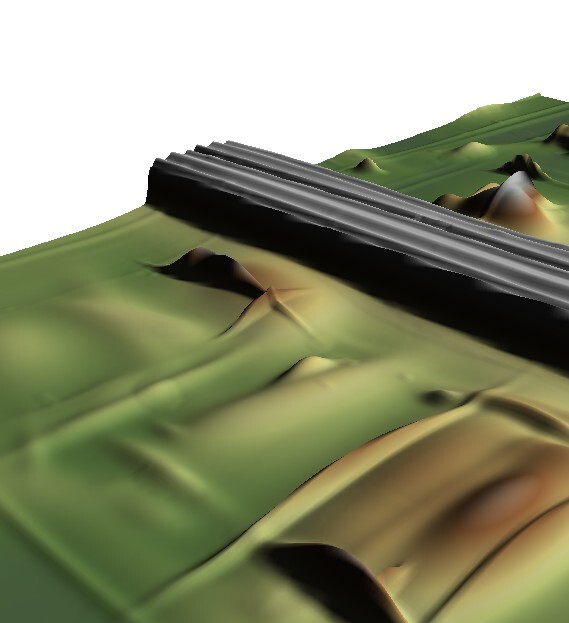} &
\includegraphics[width=0.110\textwidth]{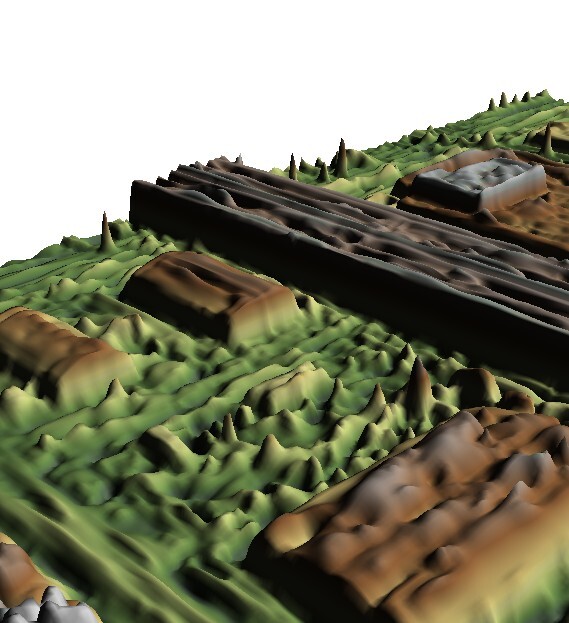} &
\includegraphics[width=0.110\textwidth]{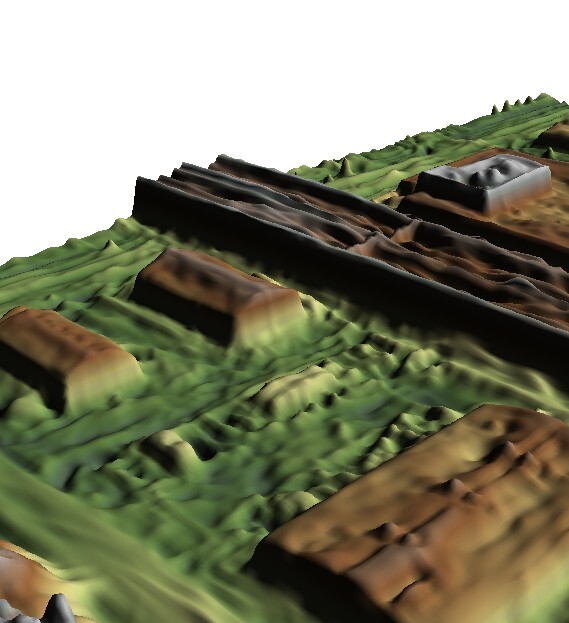} &
\includegraphics[width=0.110\textwidth]{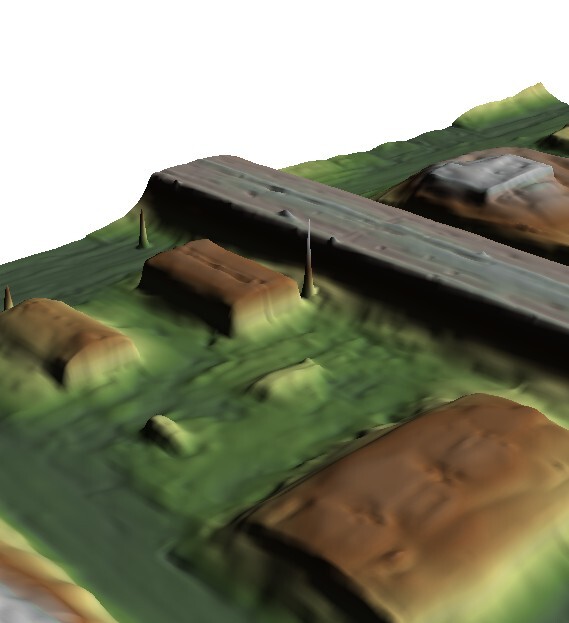} &
\includegraphics[width=0.110\textwidth]{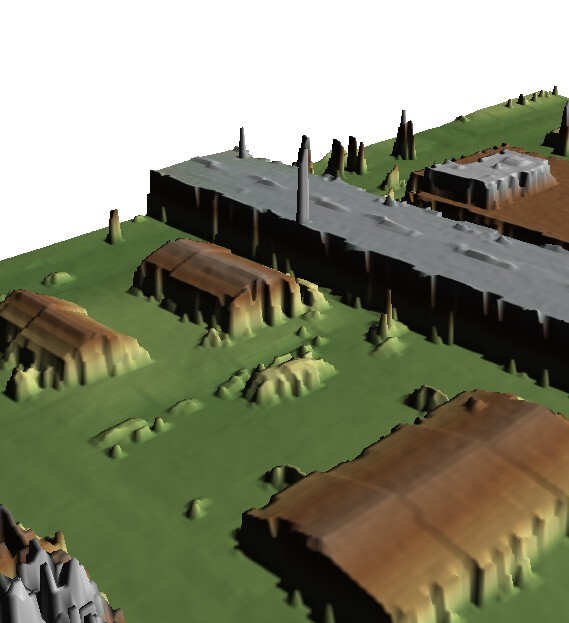} \\

\small (a) & \small (b) & \small (c) & \small (d) & \small (e) & \small (f) & \small (g) & \small (h) & \small (i) \\
\end{tabular}

\caption{\rev{Qualitative comparison on the JAX-168 (top), OMA-212 (middle), and OMA-315 (bottom) sites, where each column represents a reconstruction method: (a) Google Earth, (b) ASP, (c) s2p, (d) Sat-NGP, (e) EOGS, (f) Skyfall-GS (w/o IDU), (g) Skyfall-GS (w/ IDU), (h) SatSplat (Ours), and (i) GT. SatSplat reconstructs fine-grained details, such as trees and small structures on buildings while maintaining a smooth flat surface, which are missed or poorly reconstructed by other methods.}}
\label{fig:ourdataset_comparison}
\vspace{-10px}
\end{figure*}

\subsection{Ablation Studies}
\begin{figure}[!htbp]
    \centering
    \begin{subfigure}[b]{0.45\linewidth}
        \includegraphics[width=\linewidth]{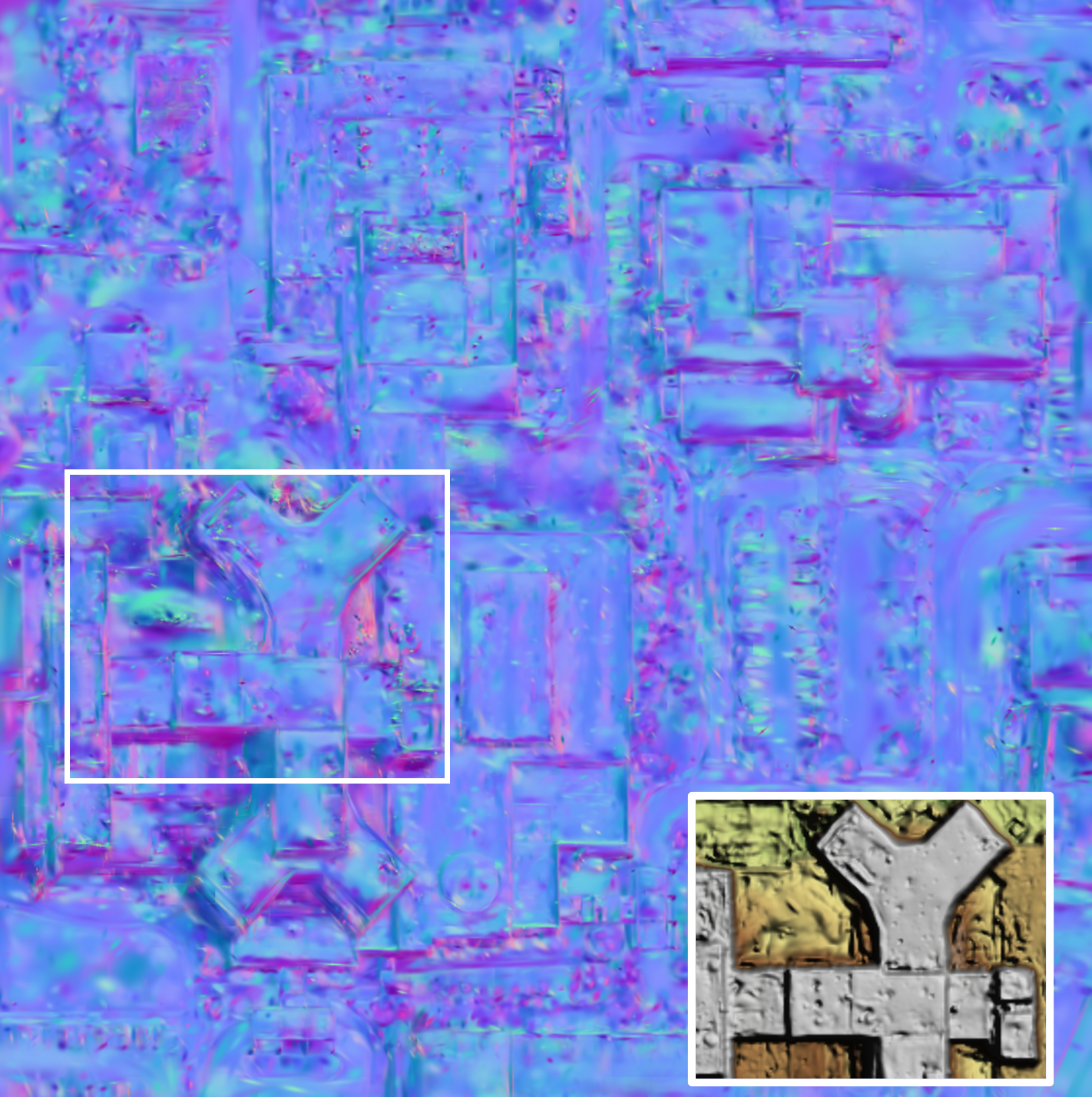}
        \caption{Without normal loss (C2).}
        \label{fig:without_normal}
    \end{subfigure}
    \begin{subfigure}[b]{0.45\linewidth}
        \includegraphics[width=\linewidth]{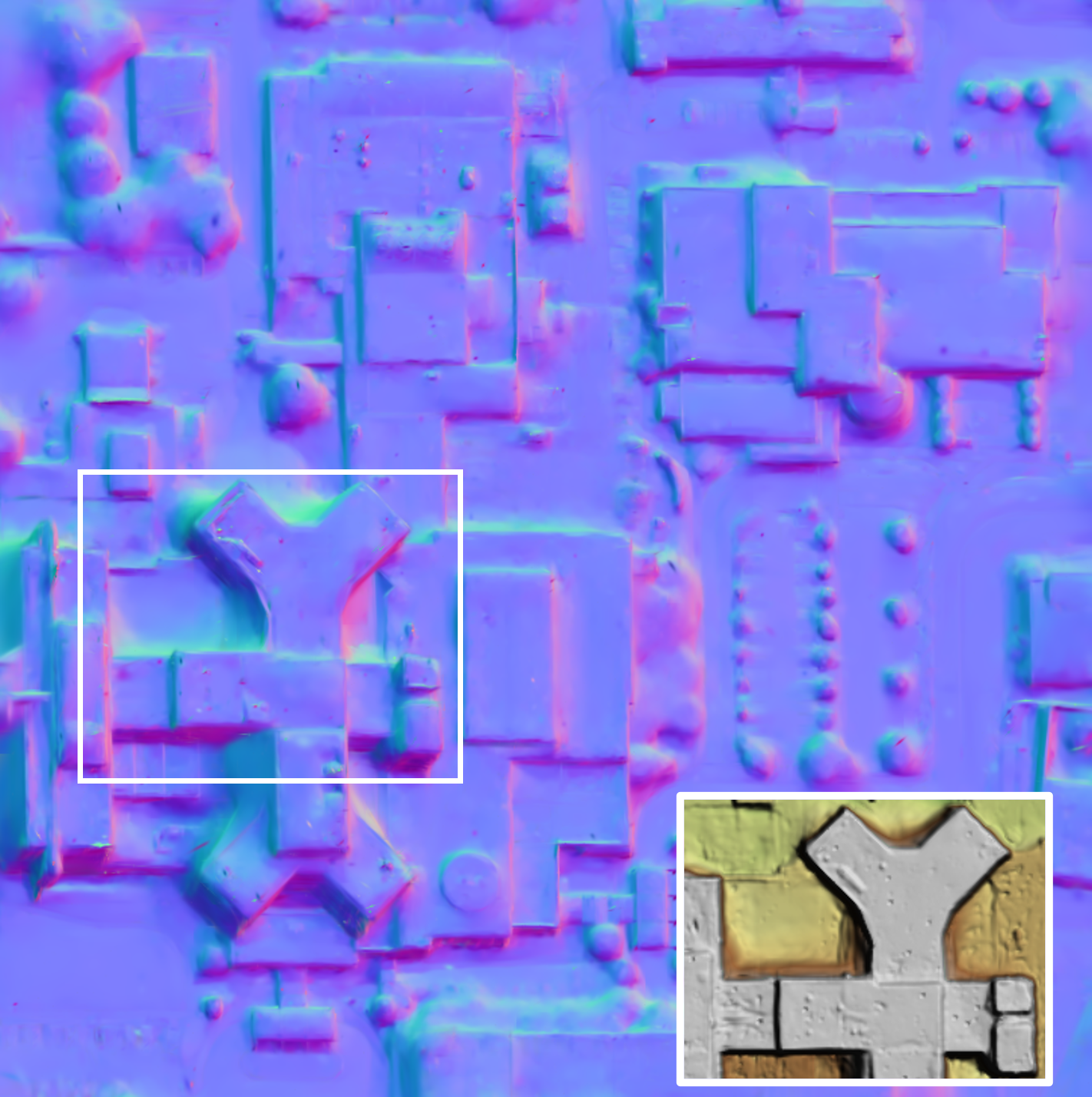}
        \caption{With normal loss (C5)}
        \label{fig:with_normal}
    \end{subfigure}
    \caption{SatSplat normal maps with and without the normal loss. The inclusion of the normal loss generates a more accurate surface as shown in the DSM crop at bottom-right.}
    \label{fig:normal}
\end{figure}

To assess the contribution of each proposed component in SatSplat, we conduct ablation studies on all reconstructed tiles and report the average $\mathrm{MAE}_{reg}$ in \cref{tab:ablation}. \rev{\linelabel{rev:3dgs_control}We explicitly include a 3DGS control (C4) alongside the 2DGS variants to separate the impact of the scene representation from that of the auxiliary training components.} As shown in the table, the normal loss provides the most significant contribution, reducing the $\mathrm{MAE}_{reg}$ by 30.22\% (C2$\rightarrow$C5, C3$\rightarrow$C6). The second-largest contribution stems from the coarse DSM initialization, which reduces the $\mathrm{MAE}_{reg}$ by 21.26\% (C1$\rightarrow$C2). \cref{fig:normal} illustrates the improved surface quality achieved with the normal loss, a component that effectively leverages the 2DGS representation. \rev{We mark the normal-loss entry for the 3DGS control as N/A because this primitive does not provide an explicit surface or surfel parameterization for defining the same surface-normal regularization. Under the same geometry-aware initialization, the 3DGS control (C4, 1.60) achieves an 8.04\% lower $\mathrm{MAE}_{reg}$ than the corresponding 2DGS control (C2, 1.74), suggesting that in this reduced setting, where only geometry-aware initialization is enabled and no explicit surface regularization is applied, the more flexible 3DGS primitive adapts better than 2DGS in our satellite setup. This also indicates that the final performance gain should be attributed to the full formulation rather than to the representation alone.}

The contribution of the delta affine camera optimization appears dependent on data quality. It provides a marginal contribution on the DFC2019 dataset, resulting in a 1.98\% $\mathrm{MAE}_{reg}$ reduction (C2$\rightarrow$C3 and C5$\rightarrow$C6). We further analyze its robustness to camera errors in a controlled disturbance experiment in \cref{sec:camopt_capacity}.

\begin{table}[!ht]
\revcolor
\renewcommand{\baselinestretch}{1}\selectfont
\renewcommand{\arraystretch}{1}
\centering
\small
\caption{\rev{Ablation study of each proposed component in SatSplat. C1-C6 represent different combinations of components. C4 uses 3DGS for comparison, while all other configurations use 2DGS. N/A indicates that the normal loss is not applicable to the 3DGS control because it does not provide an explicit surface or surfel representation.}}
\label{tab:ablation}
\resizebox{\columnwidth}{!}{%
\begin{tabular}{lcccccc}
\toprule
 & \textbf{C1} & \textbf{C2} & \textbf{C3} & \textbf{C4} & \textbf{C5} & \textbf{C6} \\
\midrule
3DGS &  &  &  & \Checkmark &  &  \\
2DGS & \Checkmark & \Checkmark & \Checkmark &  & \Checkmark & \Checkmark  \\
Normal Loss &  &  &  & N/A & \Checkmark & \Checkmark  \\
Cascade Camera Optimization  &  &  & \Checkmark &  &  & \Checkmark  \\
Geometry-Aware Initialization &  & \Checkmark & \Checkmark & \Checkmark & \Checkmark & \Checkmark \\
\midrule
$\mathrm{MAE}_{reg}$ & 2.21 & 1.74 & 1.70 & 1.60 & 1.21 & \textbf{1.19} \\
Train Time (minutes) & 10.25 & 13.53 & 12.12 & 11.60 & 13.58 & 13.07 \\
Peak VRAM (GB) & 1.99 & 1.62 & 1.64 & 1.70 & 1.63 & 1.65 \\
\bottomrule
\end{tabular}
}
\end{table}

\subsection{Camera Optimization Robustness} \label{sec:camopt_capacity}
\begin{figure}
\centering
\includegraphics[width=\linewidth]{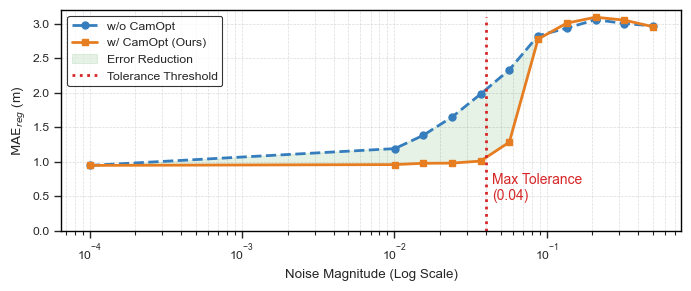}
\caption{\rev{Camera optimization tolerance analysis. The horizontal axis shows the magnitude of injected error on a logarithmic scale.}}
\label{fig:camopt_capacity}
\end{figure}

While DFC2019 has good data quality, we designed an experiment to verify our method's ability to handle imperfect camera poses by manually controlling the magnitude of applied errors. We randomly generated a $2\times4$ matrix from a normal distribution for each camera, normalized it with the Frobenius norm, and then scaled it by a chosen magnitude to simulate error. \rev{We evaluate the reconstruction both with and without camera optimization. The results, shown in \cref{fig:camopt_capacity}, indicate that with camera optimization our method can tolerate up to $1e^{-1}$ error in the affine camera matrix and is nearly unaffected by errors below $4e^{-2}$. In contrast, without optimization, the $\mathrm{MAE}_{reg}$ degrades immediately and monotonically with the injected noise.} This robust behavior is critical when processing data where the initial camera pose is not ideal.

\subsection{Scalability}

\begin{figure}[!htbp]
\centering
\includegraphics[width=\linewidth]{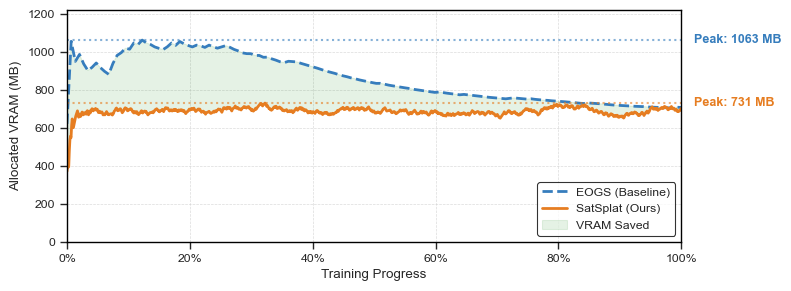}
\caption{\rev{VRAM usage analysis during training. The horizontal axis shows the progress of the entire training process. Our method consistently optimizes on 100K splats, while EOGS ends up with 30K--80K chosen by the opacity-based pruning.}}
\label{fig:memory}
\end{figure}

EOGS initializes Gaussians by randomly sampling the entire volumetric space and then pruning them based on an empirical opacity threshold. We found this to be less efficient when scaling to large scenes because it has a high peak in video RAM (VRAM) consumption at startup (\cref{fig:memory}), and the final number of points is hard to control (on DFC2019 with default parameters, it initializes around 2M Gaussians and eventually obtains about 30K--80K). Although SatSplat requires more training epochs and yields more Gaussians in the end, its VRAM consumption is more stable and can be explicitly bounded. 

\begin{figure}[!ht]
\centering
\includegraphics[width=\linewidth]{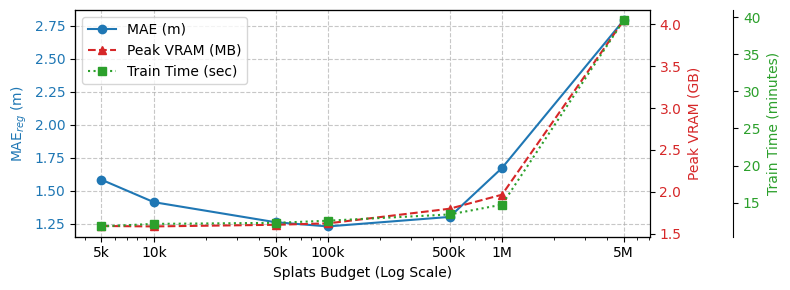}
\caption{\rev{Accuracy-efficiency trade-off of SatSplat under different maximum splat budgets. As the splat quota increases from 5K to 5M, peak VRAM usage and training time increase monotonically. However, the geometric accuracy ($\mathrm{MAE}_{reg}$) reaches its optimal performance at 100K before degrading due to over-fitting with excessively large budgets.}}
\label{fig:accuracy_efficiency}
\end{figure}

\rev{To further investigate the relationship between accuracy and resource consumption, we vary the maximum splat quota from 5K to 5M on the DFC2019 dataset (\cref{fig:accuracy_efficiency}). As the budget increases, both peak VRAM usage and training time grow monotonically. However, the geometric accuracy ($\mathrm{MAE}_{reg}$) behaves non-monotonically, reaching its optimum (lowest $\mathrm{MAE}_{reg}$) at a budget of 100K. We attribute this to the fact that excessively large budgets provide too much representational capacity, allowing the model to over-fit to photometric variations or noise by placing floaters in empty spaces rather than adhering strictly to the true surface geometry. This insight justifies our default setting of 100K, which strikes an optimal balance between precision and computational efficiency, making SatSplat highly suitable for large-scale satellite reconstruction.}

\subsection{\rev{Limitations and Failure Cases}\label{sec:limitations}}
\begin{figure}[!htbp]
    \centering
    \begin{subfigure}[b]{0.24\linewidth}
        \includegraphics[width=\linewidth]{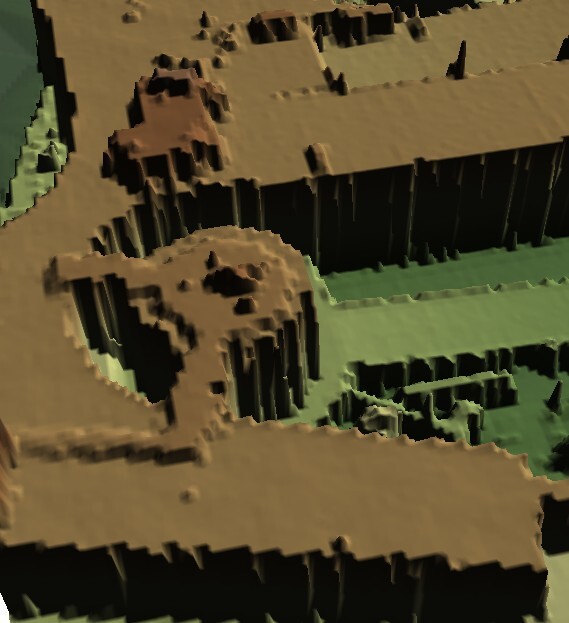}
        \caption{\rev{Ground Truth}}
    \end{subfigure}
    \begin{subfigure}[b]{0.24\linewidth}
        \includegraphics[width=\linewidth]{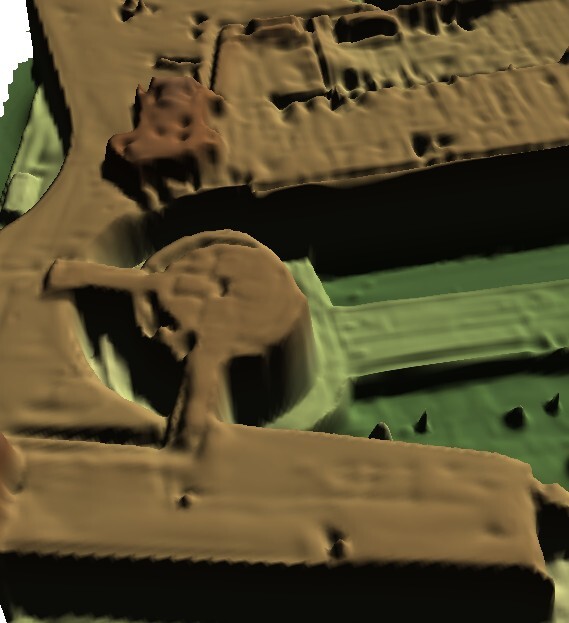}
        \caption{\rev{EOGS}}
    \end{subfigure}
    \begin{subfigure}[b]{0.24\linewidth}
        \includegraphics[width=\linewidth]{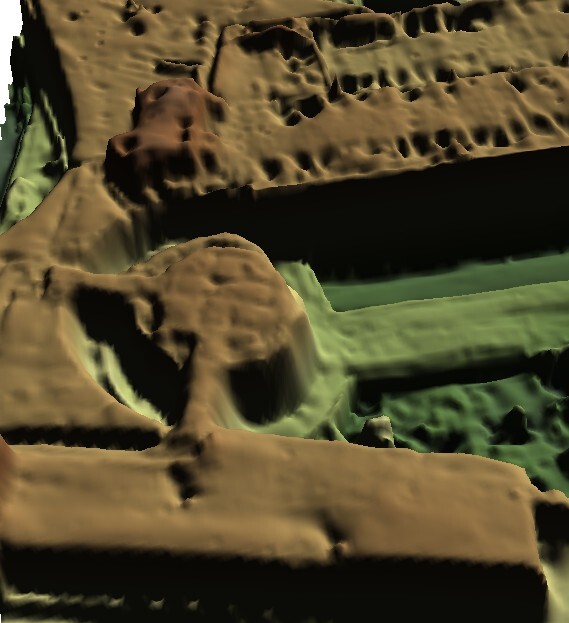}
        \caption{\rev{SatSplat (3DGS)}}
    \end{subfigure}
    \begin{subfigure}[b]{0.24\linewidth}
        \includegraphics[width=\linewidth]{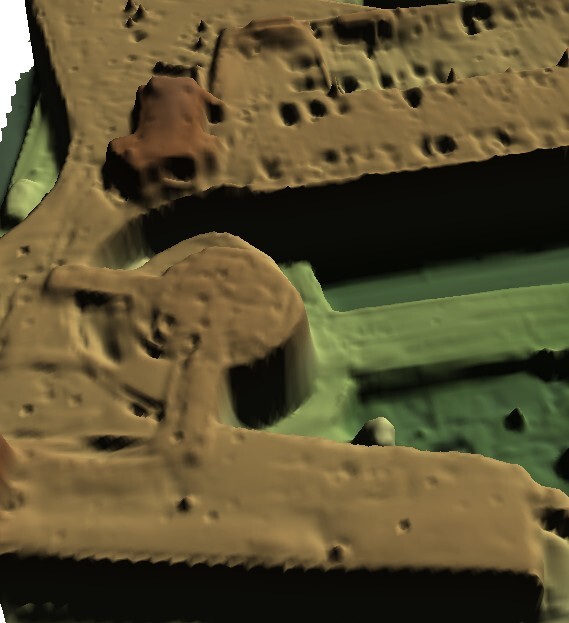}
        \caption{\rev{SatSplat (2DGS)}}
    \end{subfigure}
    \caption{\rev{A failure case on the JAX-214 site. The explicit piecewise planar approximations of the 2DGS representation lead to an over-smoothed surface, failing to recover the narrow gap between the buildings—an issue not present when using the volumetric 3DGS primitive.}}
    \label{fig:limitation_jax214}
\end{figure}

\rev{Beyond common satellite reconstruction challenges—such as non-Lambertian surfaces (e.g., water), extreme temporal variations, and image pairs with minimal intersection angles—our method has two notable limitations. First, as illustrated in \cref{fig:limitation_jax214}, the strict piecewise planar assumptions of the 2DGS representation can lead to over-smoothing in challenging narrow concavities, failing to recover deep architectural gaps. This issue is absent when using the more flexible, volumetric 3DGS primitive, indicating that it is a direct consequence of the 2DGS surface assumptions struggling to represent high-frequency depth discontinuities.}

\rev{Second, the benefit of our online camera optimization is marginal on standard public benchmarks (like DFC2019 and IARPA2016) because they already provide highly accurate initial geometry. While the module serves as an effective robustness mechanism under heavily perturbed poses (\cref{sec:camopt_capacity}), its advantage is less pronounced when initial RPCs are near-optimal.}

\section{Conclusion}
We introduce SatSplat, the first 2DGS framework for reconstructing very high-resolution satellite scenes. The planar representation of 2D Gaussians is inherently better suited for modeling expansive surfaces, such as the Earth's, than volumetric 3D Gaussians. Incorporating clearly defined normals further improves the geometric quality of the reconstruction. The framework also provides the infrastructure to integrate geometric priors from external sources, such as monocular depth estimation networks, to further enhance geometric accuracy.
\rev{Our analysis on the evaluated DFC2019 and IARPA2016 benchmark sites shows that SatSplat achieves strong geometric accuracy and produces fewer artifacts in novel view synthesis than prior satellite Gaussian-splatting baselines.} In addition, SatSplat is memory-efficient and scalable to larger areas than previous NeRF-based and EOGS methods. \rev{We believe SatSplat represents a meaningful step forward in large-scale, satellite-based 3D reconstruction while also highlighting the remaining challenges posed by water, weak parallax, and strong temporal change.}

\section*{Acknowledgements}
This work is supported by the Office of Naval Research (Award No. N000142312670) and Intelligence Advanced Research Projects Activity (IARPA) via Department of Interior/Interior Business Center (DOI/IBC) contract number 140D0423C0075. This work is also supported by the U.S. National Science Foundation (NSF) under Award No. 2331104.

\setcounter{section}{0}
\setcounter{subsection}{0}
\setcounter{figure}{0}
\setcounter{table}{0}
\setcounter{equation}{0}

\renewcommand\thesection{\Alph{section}}
\renewcommand\thesubsection{\thesection.\arabic{subsection}}
\renewcommand\thefigure{S\arabic{figure}}
\renewcommand\thetable{S\arabic{table}}
\renewcommand\theequation{S\arabic{equation}}

% Prefix hyperref anchors to avoid duplicates with main text
\renewcommand{\theHsection}{A.\thesection}
\renewcommand{\theHsubsection}{A.\thesubsection}
\renewcommand{\theHfigure}{A.\thefigure}
\renewcommand{\theHtable}{A.\thetable}
\renewcommand{\theHequation}{A.\theequation}

\appendixtitle{\Large SatSplat: Geometrically-Accurate Gaussian Splatting for Satellite Imagery \\[.5em] \large Supplementary Material}

\section{Affine Camera of Sun Direction}
\label{supsec:sun_camera}

We first construct a projector that flattens 3D points onto the $z=0$ plane along the sun direction. Given the sun azimuth and elevation angles $(\phi,\theta)$ and a 3D point $\mathbf{p}$, we flatten $\mathbf{p}$ along the sun direction as in \cref{eq:flatten_along_sun}:

\begin{equation}
	\mathbf{p}' = \mathbf{p} - p_z
 \begin{bmatrix}
 \frac{\cos\phi}{\tan\theta} \\
 \frac{\sin\phi}{\tan\theta} \\
 1
 \end{bmatrix}
 = \left(\mathbf{I} - \begin{bmatrix}
		0 & 0 & \frac{\cos\phi}{\tan\theta} \\
		0 & 0 & \frac{\sin\phi}{\tan\theta} \\
		0 & 0 & 1
	\end{bmatrix}\right)\mathbf{p},
	\label{eq:flatten_along_sun}
\end{equation}
where $p_z$ is the altitude (world $z$-coordinate) of $\mathbf{p}$.

For convenience, we denote this sun direction by
\begin{equation}
 \mathbf{s} =
 \begin{bmatrix}
 \frac{\cos\phi}{\tan\theta} \\
 \frac{\sin\phi}{\tan\theta} \\
 1
 \end{bmatrix},
\end{equation}
so that \cref{eq:flatten_along_sun} can be written compactly as
\begin{equation}
 \mathbf{p}' = \mathbf{p} - p_z\,\mathbf{s}.
\end{equation}

When we concatenate this projector with the affine camera matrix $\mathbf{B}_{xy}$ defined in \cref*{eq:decompose_affine_camera} in the main paper, we obtain the sun-camera mapping

\begin{equation}
\begin{aligned}
\mathbf{B}_{xy}^{sun}\mathbf{p} &= \mathbf{B}_{xy}\mathbf{p}' \\
& = \mathbf{B}_{xy} \left(\mathbf{I} - \begin{bmatrix}
	0 & 0 & \frac{\cos\phi}{\tan\theta} \\
	0 & 0 & \frac{\sin\phi}{\tan\theta} \\
	0 & 0 & 1
\end{bmatrix}\right) \mathbf{p}.
\end{aligned}
\label{eq:derive_B_xy_sun}
\end{equation}

Next, we show that this direction is the viewing direction of the sun camera, i.e.\ it lies in the nullspace of $\mathbf{B}_{xy}^{sun}$. From \cref{eq:flatten_along_sun}, evaluating the projector at $\mathbf{s}$ gives
\begin{equation}
 \mathbf{s}' = \mathbf{s} - s_z\,\mathbf{s}
 = \mathbf{s} - 1\cdot\mathbf{s} = \mathbf{0},
\end{equation}
since the third component of $\mathbf{s}$ equals $1$ by construction. Therefore
\begin{equation}
 \mathbf{B}_{xy}^{sun}\,\mathbf{s}
 = \mathbf{B}_{xy}\,\mathbf{s}'
 = \mathbf{B}_{xy}\,\mathbf{0}
 = \mathbf{0},
\end{equation}
so $\mathbf{s}$ lies in the nullspace of the sun camera matrix $\mathbf{B}_{xy}^{sun}$. Geometrically, any point $\mathbf{p}_t = \mathbf{p} + t\,\mathbf{s}$ is flattened to the same $\mathbf{p}'$ and therefore mapped to the same image position by $\mathbf{B}_{xy}^{sun}$, which is exactly the definition of a viewing direction.

\section{Experimental Settings}
\label{supsec:experiment_settings}

We generate RGB images from the DFC2019 MSI images and crop them to cover the ground-truth DSM, similar to EOGS. We then linearly stretch pixel values to [0,255], discarding the darkest and brightest 2\% of pixels. Subsequently, we refine the RPCs using the method from~\cite{mariGenericBundleAdjustment2021} and fit an affine camera model from a normalized 3D UTM coordinate system to image pixels, to remain consistent with EOGS~\cite{Aira_2025_CVPR}. The ASP~\cite{sheanAutomatedOpensourcePipeline2016,beyerAmesStereoPipeline2018} stereo tool takes around 2 minutes per stereo pair to generate a DSM and ortho photo. We then train for 30,000 iterations per site with a batch size of 1 on a single NVIDIA RTX 6000 Ada. We use the Adam optimizer with initial learning rates of 0.075 for splat means, 0.0025 for color, 0.001 for quaternions and scales, 0.0001 for camera delta matrices, and 0.001 for camera gain and bias. The shadow mapping process is applied throughout training; view-consistency loss~\cite{Aira_2025_CVPR} is disabled; camera optimization loss starts after 3,000 iterations. Our training loss is a combination of the following terms: image L1 (0.8), image SSIM (0.2), shadow translucency~\cite{Aira_2025_CVPR} (0.01), distortion~\cite{Barron_2022_CVPR} (0.05), normal~\cite{huang2DGaussianSplatting2024} (0.05), and opacity regularization~\cite{Aira_2025_CVPR} (0.01).

\section{Full Per-Site Accuracy Results}
\label{supsec:full_accuracy_results}

\cref{tab:supp_jax_mae,tab:supp_oma_iarpa_mae} report the full per-site $\mathrm{MAE}_{reg}$ values corresponding to the dataset-level averages summarized in the main paper. We omit runtime here and retain the complete per-site breakdown for both all-class and building-only evaluations.

\begin{table}[!htbp]
\renewcommand{\baselinestretch}{1}\selectfont
\renewcommand{\arraystretch}{1}
\centering
\caption{Full per-site $\mathrm{MAE}_{reg}$ (m; lower is better) on the DFC2019 JAX benchmark.}
\label{tab:supp_jax_mae}
\resizebox{\linewidth}{!}{
\begin{tabular}{lccccccc}
\toprule
\textbf{Method} & JAX-004 & JAX-068 & JAX-214 & JAX-260 & JAX-168 & JAX-251 & JAX-280 \\
\midrule
s2p                 & 3.14 & 1.81 & 4.85 & 2.52 & 3.12 & 2.40 & 2.99 \\
SAT-NGP             & 2.38 & 1.83 & 4.05 & 3.25 & 2.72 & 1.95 & 4.89 \\
EOGS                & \underline{1.64} & \textbf{0.94} & \underline{1.50} & \underline{1.59} & \underline{1.04} & \underline{1.21} & \underline{2.55} \\
Skyfall-GS (w/o IDU) & 2.40 & 1.49 & 2.10 & 1.55 & 1.37 & 1.73 & 1.79 \\
Skyfall-GS (w/ IDU) & 2.33 & 1.47 & 2.19 & 1.62 & 1.60 & 1.89 & 2.08 \\
SatSplat (Ours)     & \textbf{1.46} & \underline{1.04} & \textbf{1.49} & \textbf{1.48} & \textbf{0.94} & \textbf{1.07} & \textbf{1.85} \\
\midrule
{\textbf{Buildings}} \\
s2p                 & 5.29 & 2.13 & 3.95 & 2.08 & 6.40 & 2.08 & 2.22 \\
SAT-NGP             & 1.82 & 1.81 & 4.29 & 6.33 & 3.59 & 1.65 & 4.31 \\
EOGS                & 1.53 & \textbf{0.93} & \textbf{0.97} & \underline{1.04} & \underline{0.98} & \underline{0.87} & 2.43 \\
Skyfall-GS (w/o IDU) & \underline{1.03} & 1.47 & 1.88 & 1.95 & 1.49 & 1.46 & \underline{0.86} \\
Skyfall-GS (w/ IDU) & 1.29 & 1.31 & 1.96 & 2.01 & 1.86 & 1.55 & 1.07 \\
SatSplat (Ours)     & \textbf{0.82} & \underline{0.99} & \underline{1.19} & \textbf{1.01} & \textbf{0.87} & \textbf{0.79} & \textbf{0.83} \\
\bottomrule
\end{tabular}
}
\end{table}

\begin{table}[!htbp]
\renewcommand{\baselinestretch}{1}\selectfont
\renewcommand{\arraystretch}{1}
\centering
\caption{Full per-site $\mathrm{MAE}_{reg}$ (m; lower is better) on the OMA and IARPA benchmarks.}
\label{tab:supp_oma_iarpa_mae}
\resizebox{\linewidth}{!}{
\begin{tabular}{lcccccc}
\toprule
\textbf{Method}  & OMA-203 & OMA-212 & OMA-315 & IARPA-001 & IARPA-002 & IARPA-003 \\
\midrule
s2p                  & \underline{1.24} & 1.66 & \textbf{1.01} & 3.23 & 3.87 & 2.75\\
SAT-NGP              & 2.93 & 0.88 & 2.23 & 2.04 & 2.10 & 2.96\\
EOGS                 & 2.18 & \underline{0.83} & 1.40& 1.65 & \textbf{1.82} & 2.10 \\
Skyfall-GS (w/o IDU)  & 1.38 & 0.94 & 1.33 & 2.40 & 2.36 & 2.90\\
Skyfall-GS (w/ IDU)  & 1.82 & 0.96 & 1.43 & 2.56 & 2.51 & 2.73\\
SatSplat (Ours)      & \textbf{1.10} & \textbf{0.63} & \underline{1.19} & \textbf{1.63} & \underline{2.09} & \textbf{2.04}\\
\midrule
{\textbf{Buildings}} \\
s2p                  & \underline{0.60} & 4.58 & \textbf{1.20} & 2.59 & 4.06 & 2.96\\
SAT-NGP              & 3.08 & 4.13 & 4.75 & 1.86 & 2.28 & 3.96\\
EOGS                 & 1.61 & 3.68 & 2.64 & 1.27 & 1.05 & 1.78\\
Skyfall-GS (w/o IDU)  & 1.19 & \underline{2.24} & 1.64 & 2.10 & 2.23 & 2.18 \\
Skyfall-GS (w/ IDU)  & 1.47 & 2.33 & 1.62 & 2.01 & 1.77 & 2.37\\
SatSplat (Ours)     & \textbf{0.59} & \textbf{2.10} & \underline{1.42} & \textbf{0.88} & \textbf{1.02} & \textbf{1.68}  \\
\bottomrule
\end{tabular}
}
\end{table}

\section{Additional Results}
\label{supsec:additional_results}

\begin{figure*}[!htbp]
\centering
\newcommand{\leftlabeljax}[1]{\adjustbox{valign=m}{\rotatebox{90}{\small #1}}}
\setlength{\tabcolsep}{1.5pt}
\renewcommand{\arraystretch}{1.1}
\begin{tabular}{m{7pt}cccccccc}
\leftlabeljax{JAX-004} &
\includegraphics[width=0.11\linewidth, valign=m]{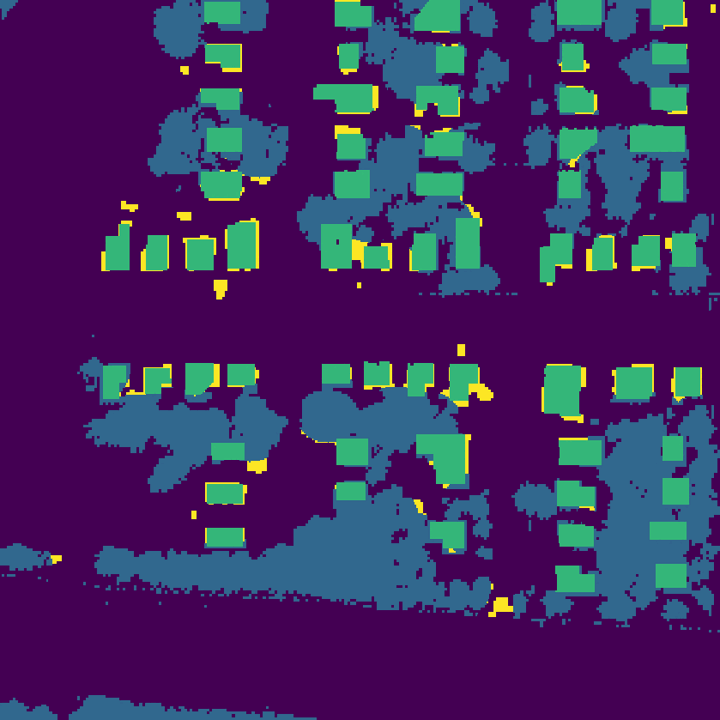} &
\includegraphics[width=0.11\linewidth, valign=m]{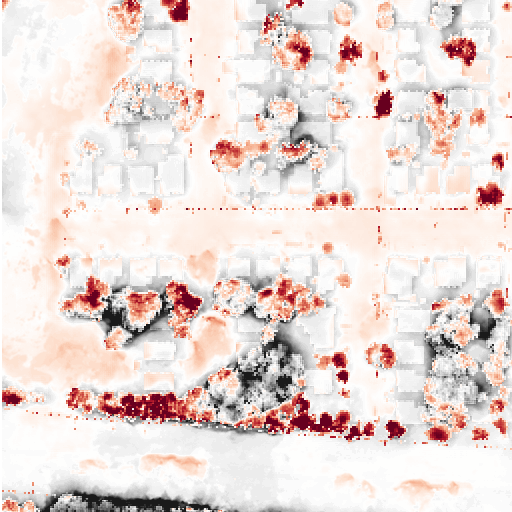} &
\includegraphics[width=0.11\linewidth, valign=m]{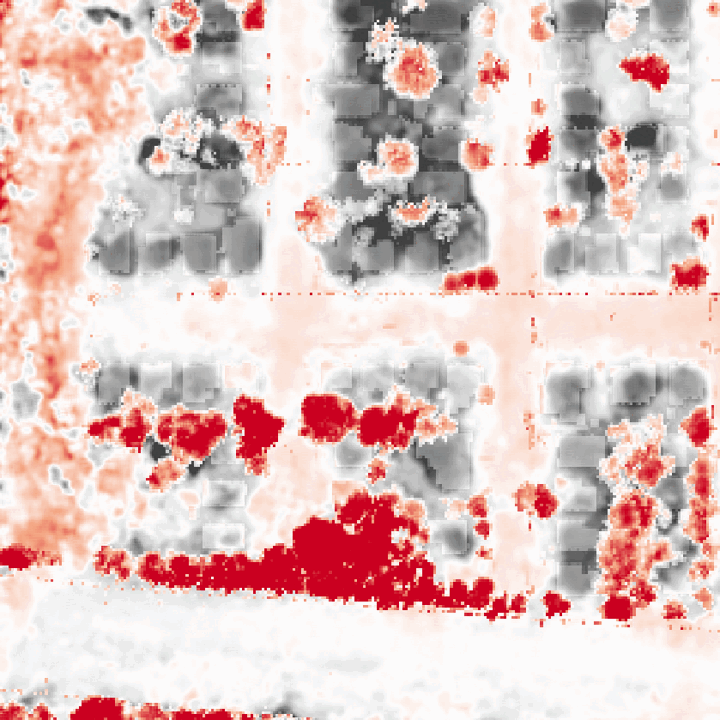} &
\includegraphics[width=0.11\linewidth, valign=m]{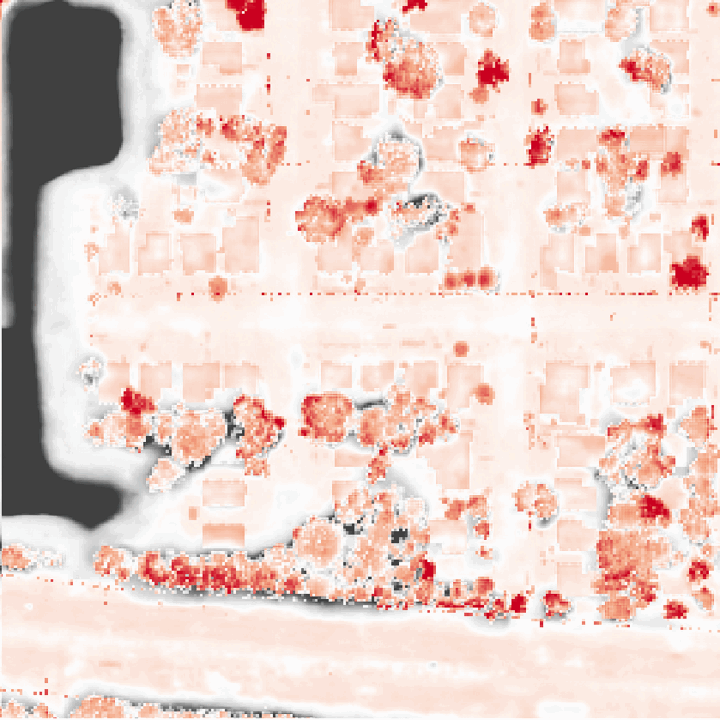} &
\includegraphics[width=0.11\linewidth, valign=m]{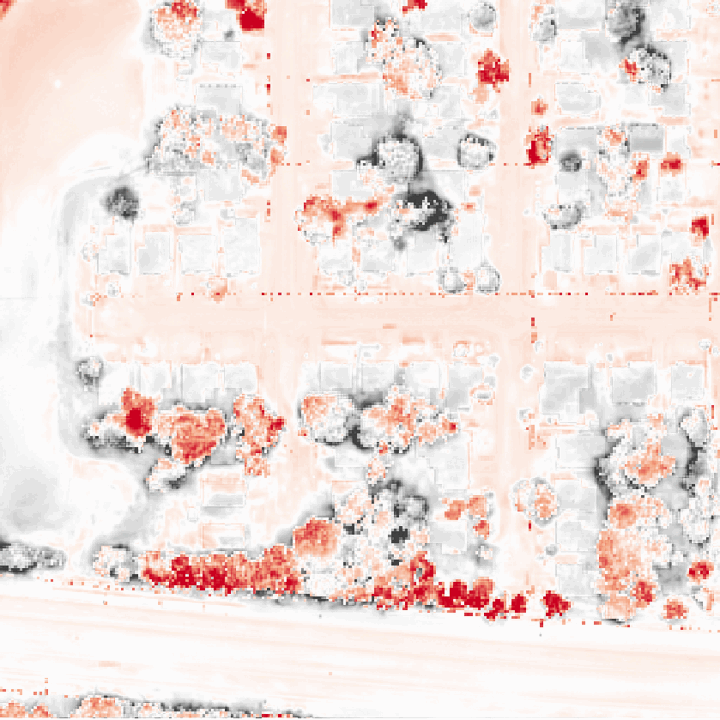} &
\includegraphics[width=0.11\linewidth, valign=m]{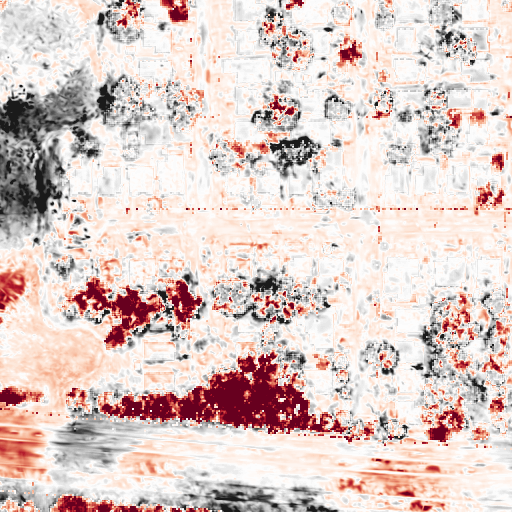} &
\includegraphics[width=0.11\linewidth, valign=m]{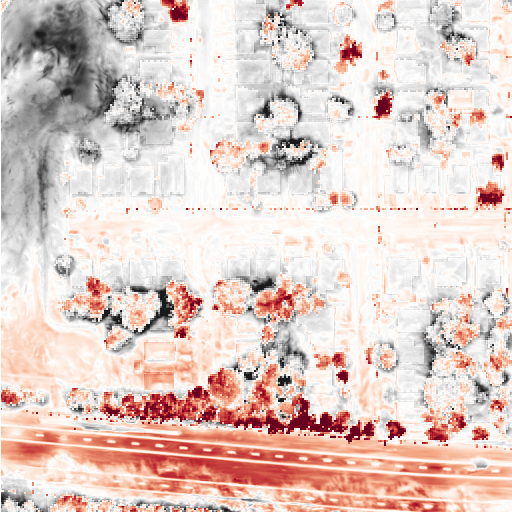} &
\includegraphics[width=0.11\linewidth, valign=m]{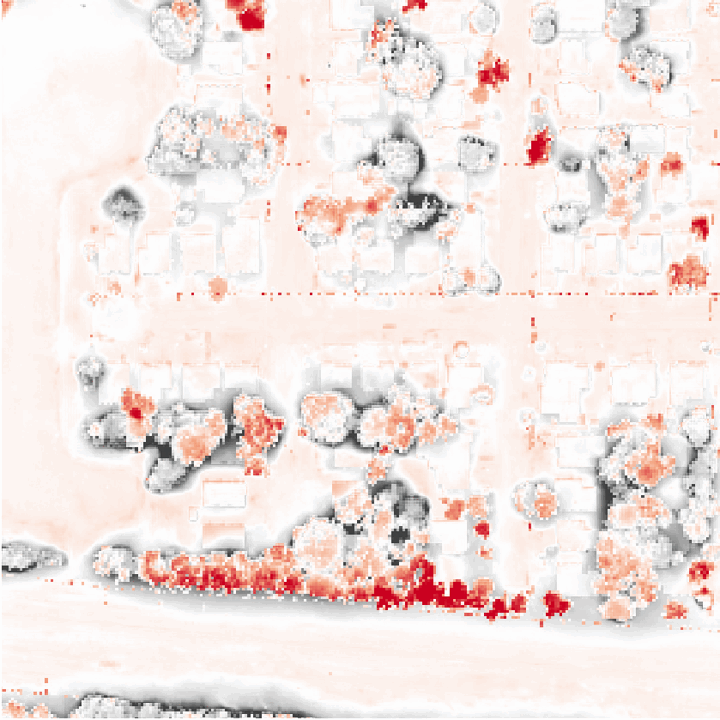} \\[2pt]

\leftlabeljax{JAX-068} &
\includegraphics[width=0.11\linewidth, valign=m]{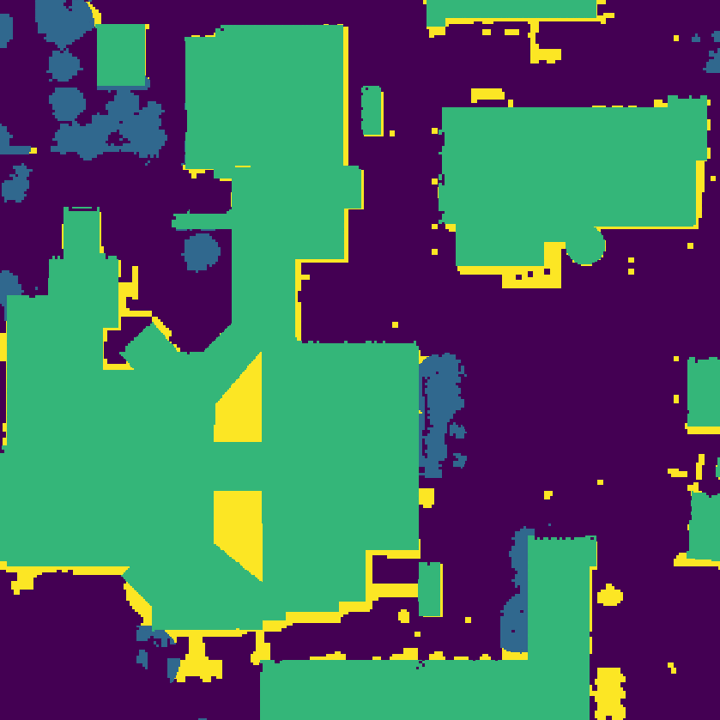} &
\includegraphics[width=0.11\linewidth, valign=m]{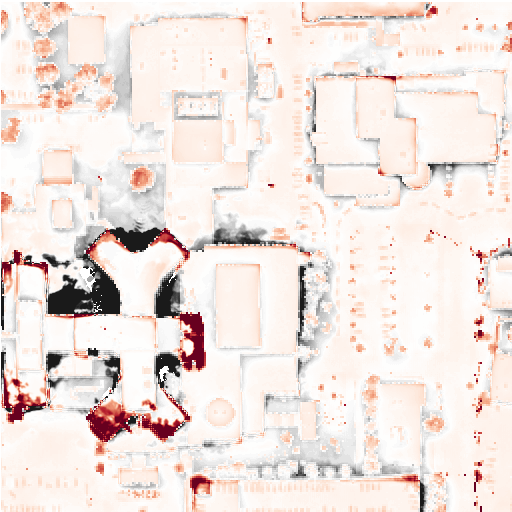} &
\includegraphics[width=0.11\linewidth, valign=m]{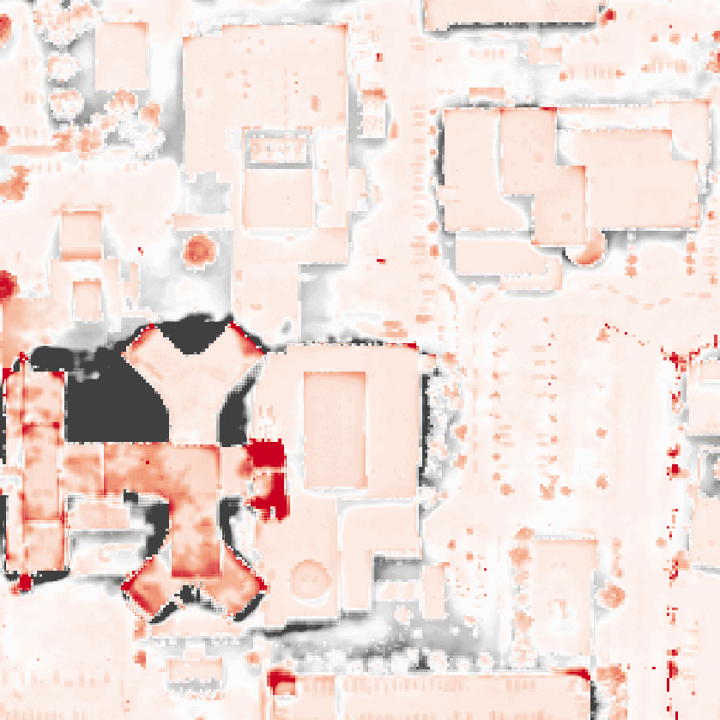} &
\includegraphics[width=0.11\linewidth, valign=m]{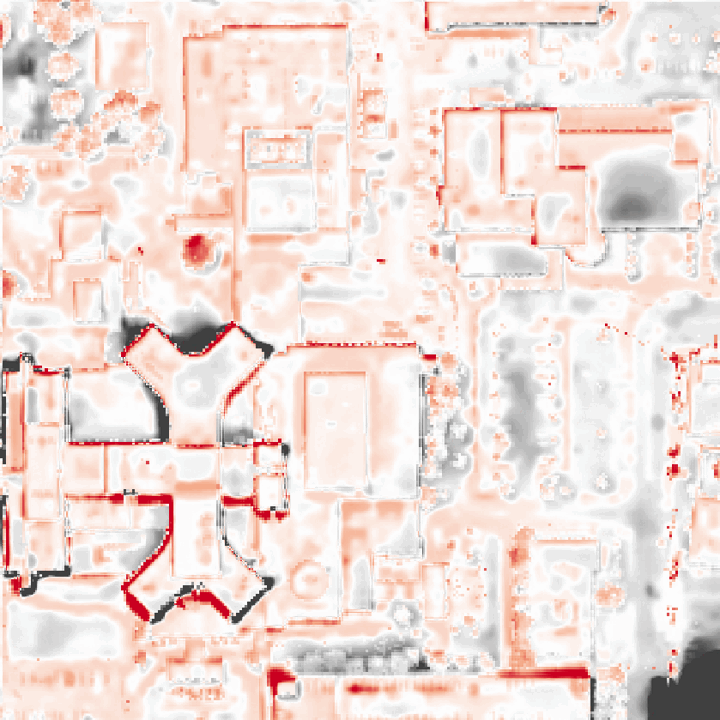} &
\includegraphics[width=0.11\linewidth, valign=m]{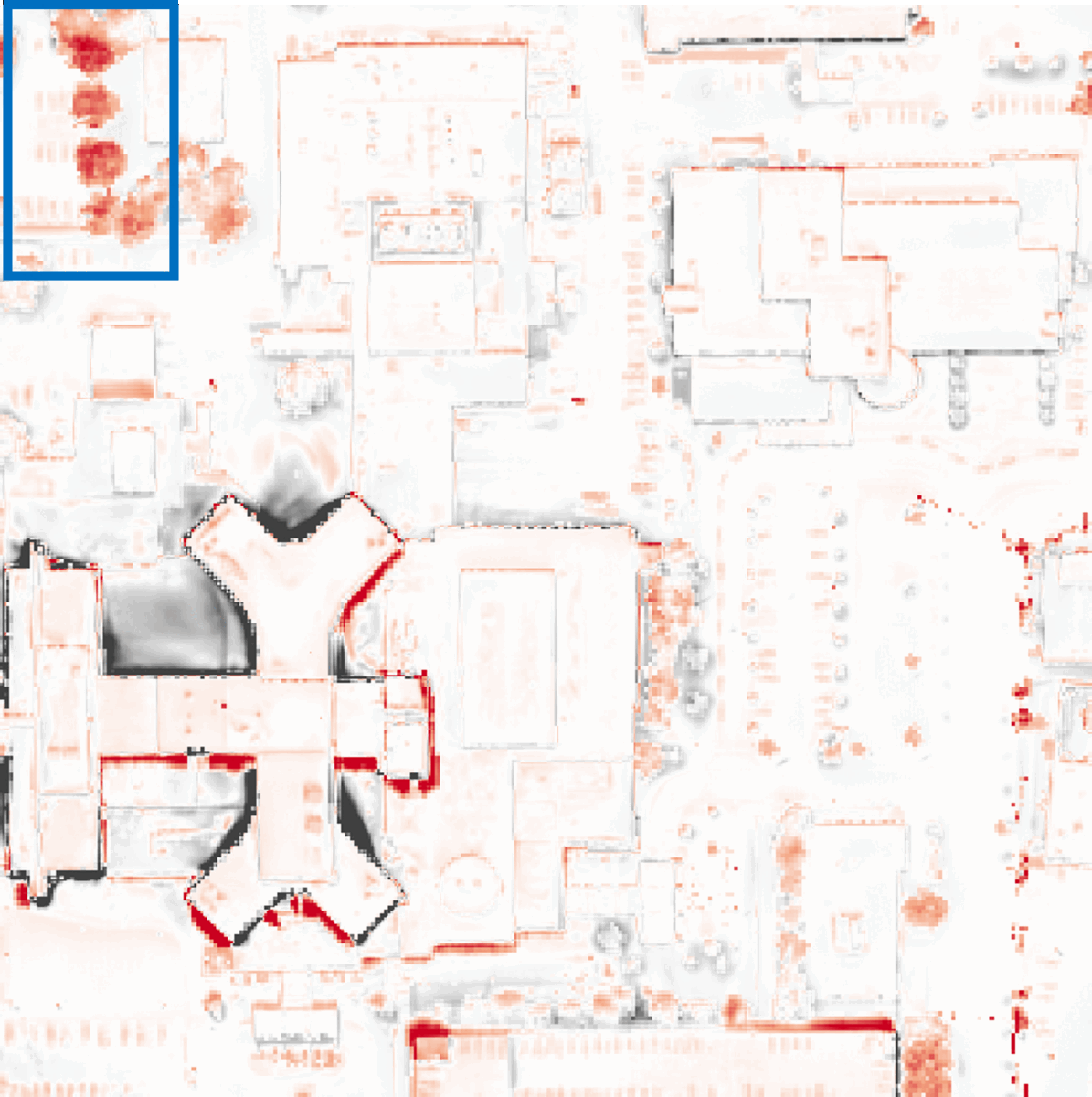} &
\includegraphics[width=0.11\linewidth, valign=m]{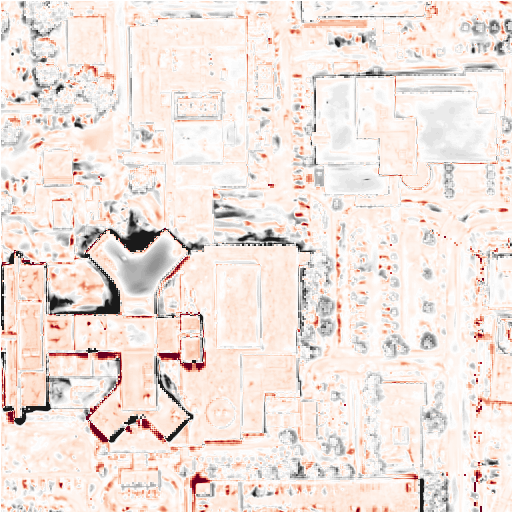} &
\includegraphics[width=0.11\linewidth, valign=m]{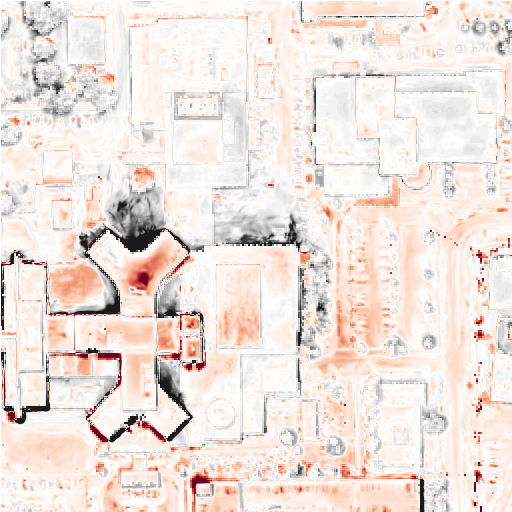} &
\includegraphics[width=0.11\linewidth, valign=m]{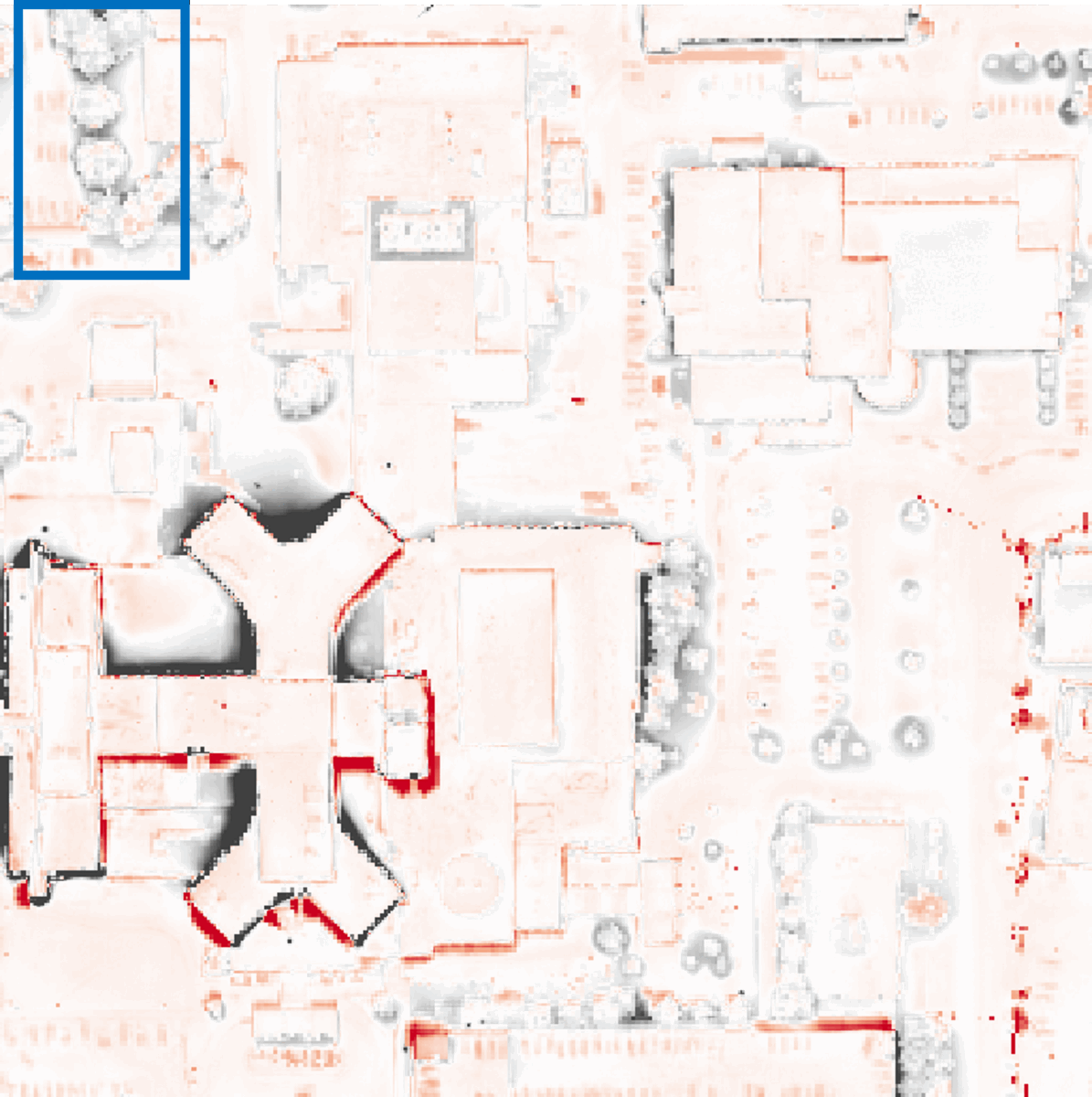} \\[2pt]

\leftlabeljax{JAX-214} &
\includegraphics[width=0.11\linewidth, valign=m]{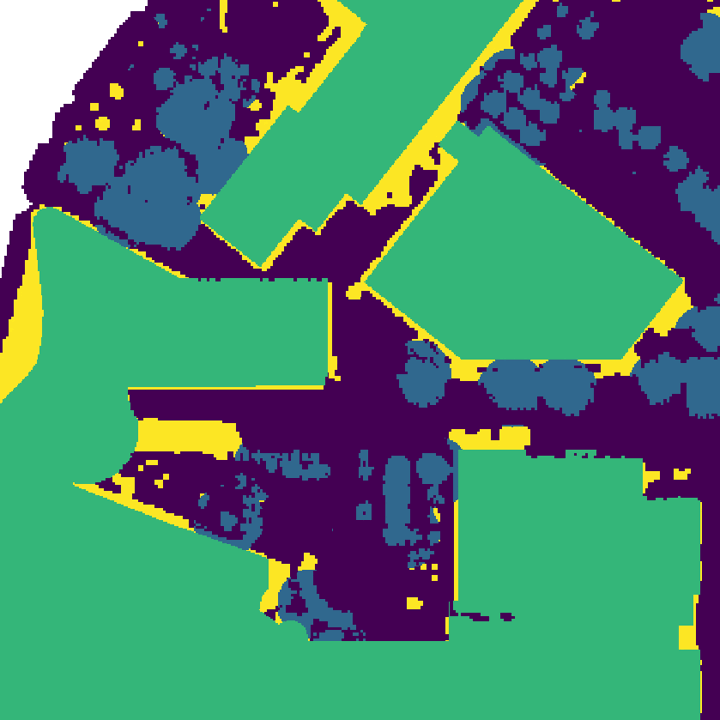} &
\includegraphics[width=0.11\linewidth, valign=m]{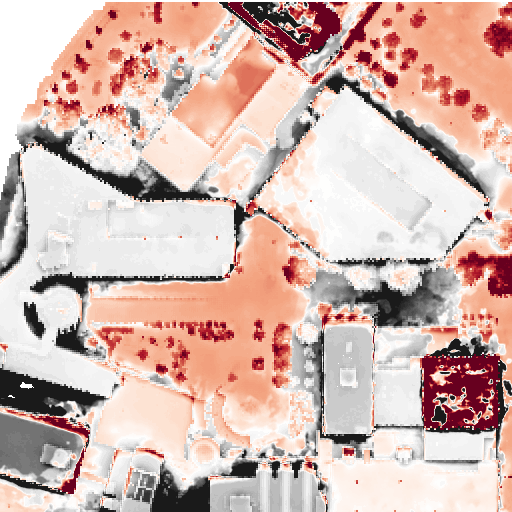} &
\includegraphics[width=0.11\linewidth, valign=m]{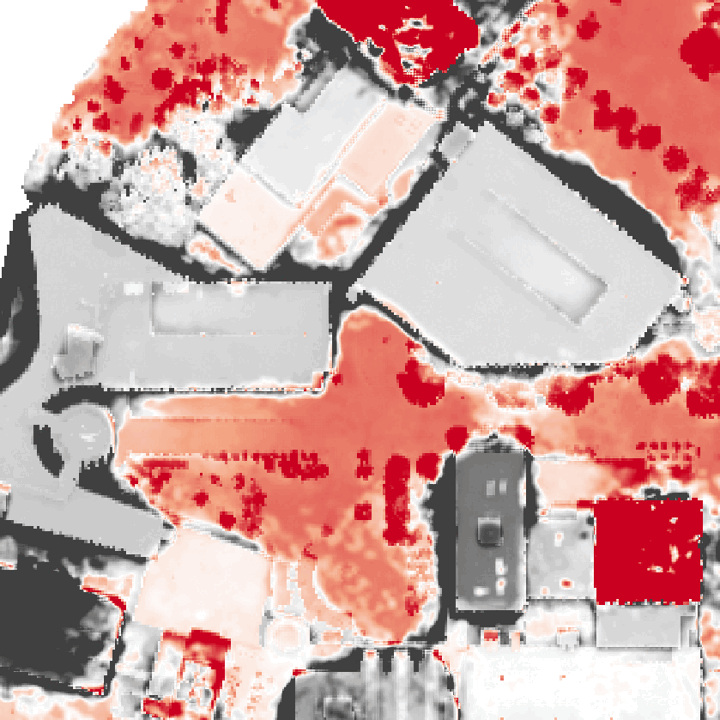} &
\includegraphics[width=0.11\linewidth, valign=m]{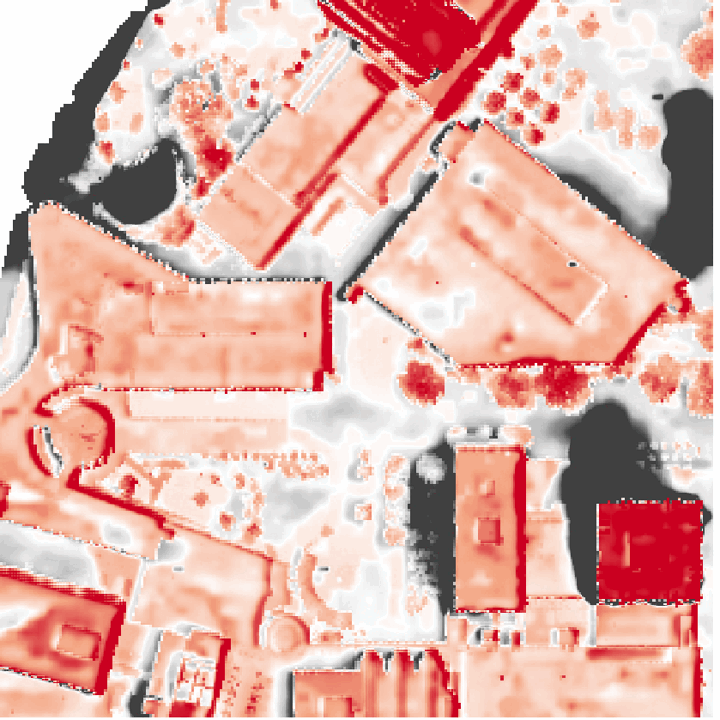} &
\includegraphics[width=0.11\linewidth, valign=m]{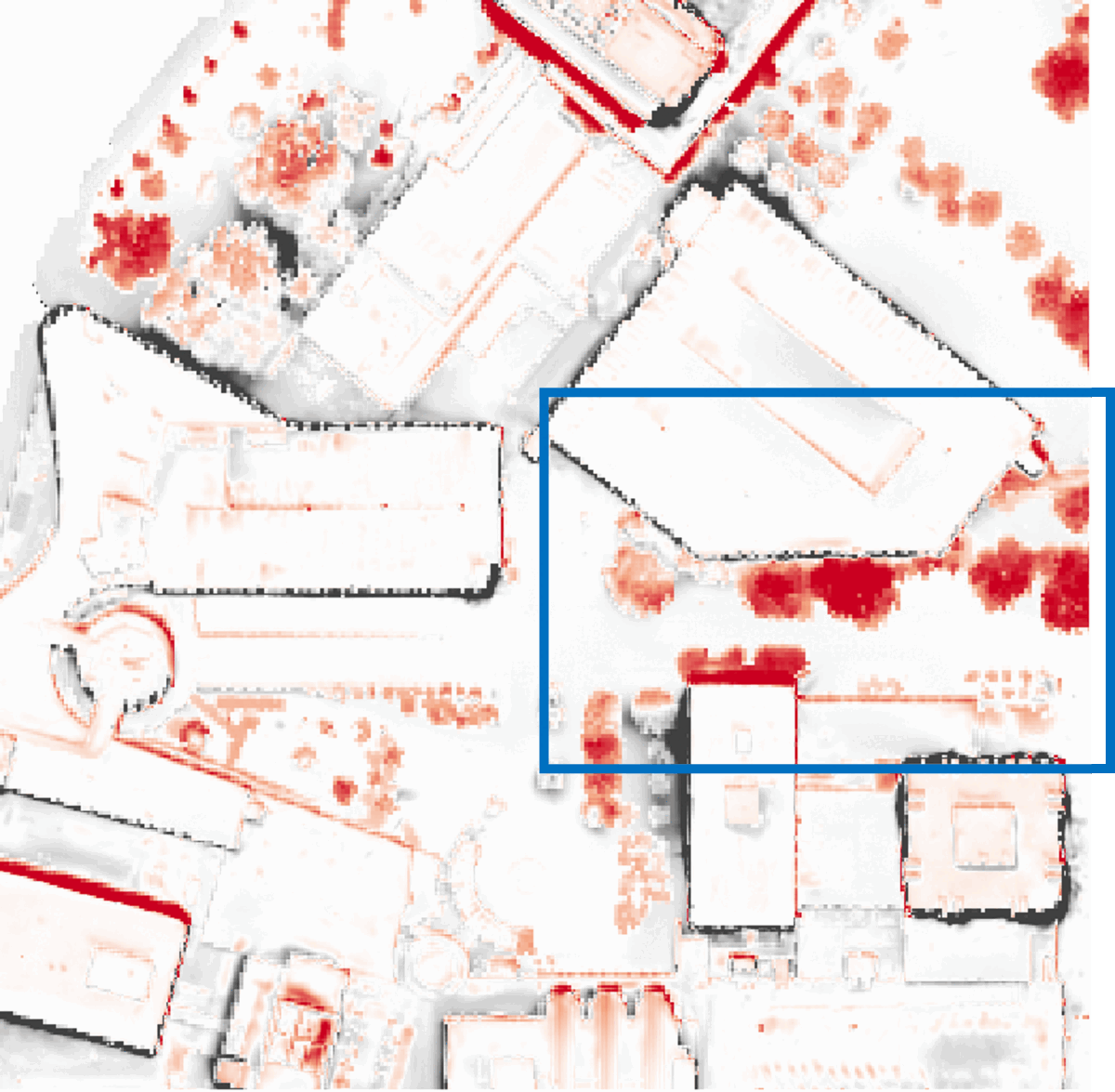} &
\includegraphics[width=0.11\linewidth, valign=m]{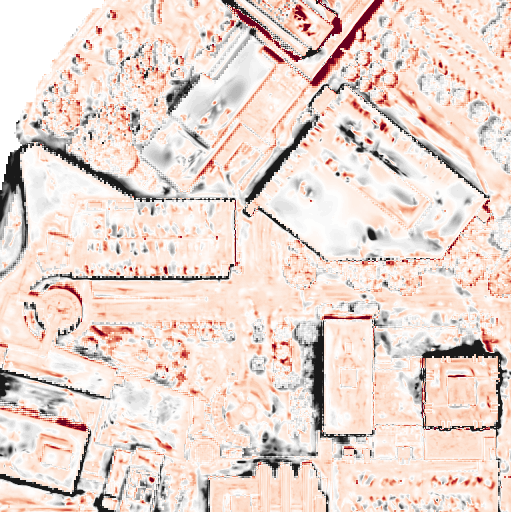} &
\includegraphics[width=0.11\linewidth, valign=m]{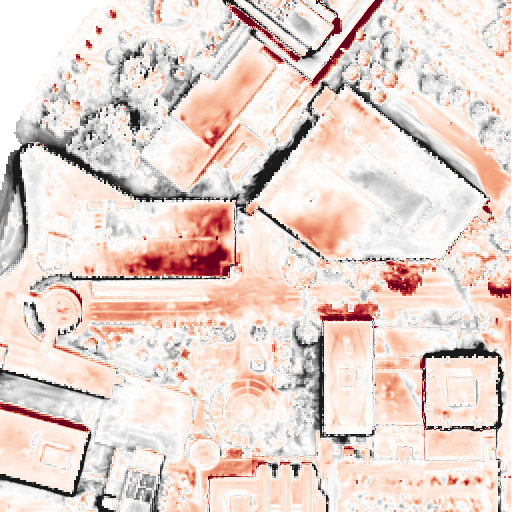} &
\includegraphics[width=0.11\linewidth, valign=m]{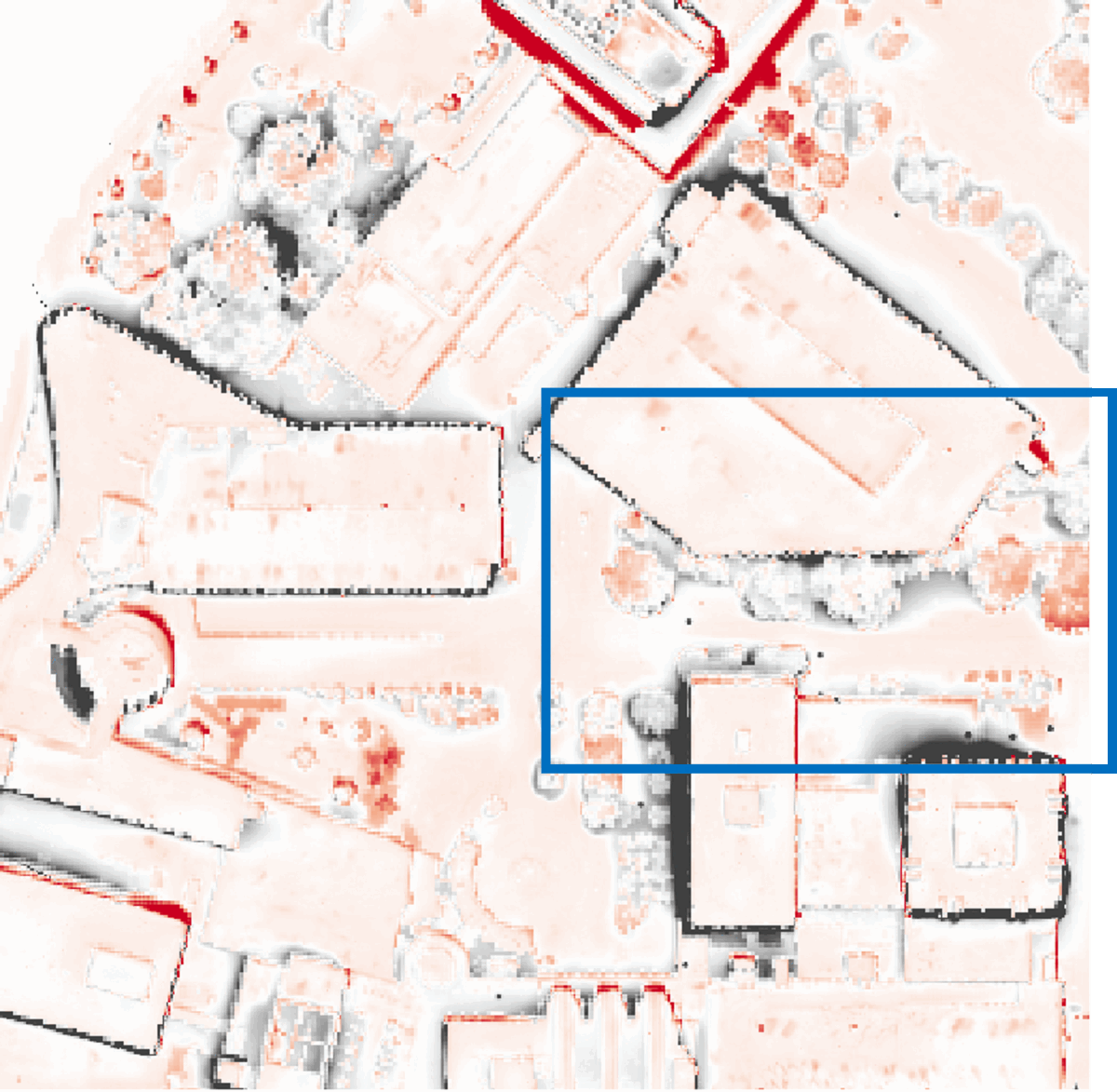} \\[2pt]

\leftlabeljax{JAX-260} &
\includegraphics[width=0.11\linewidth, valign=m]{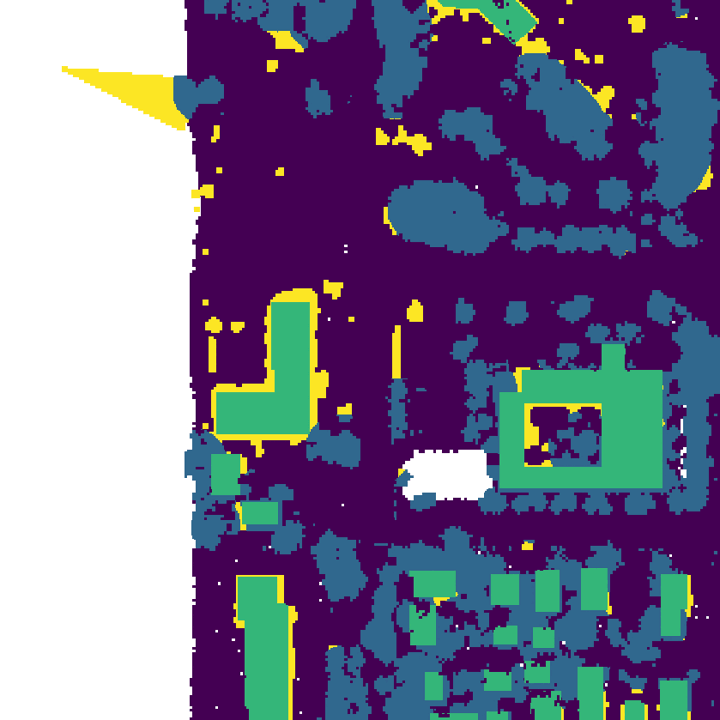} &
\includegraphics[width=0.11\linewidth, valign=m]{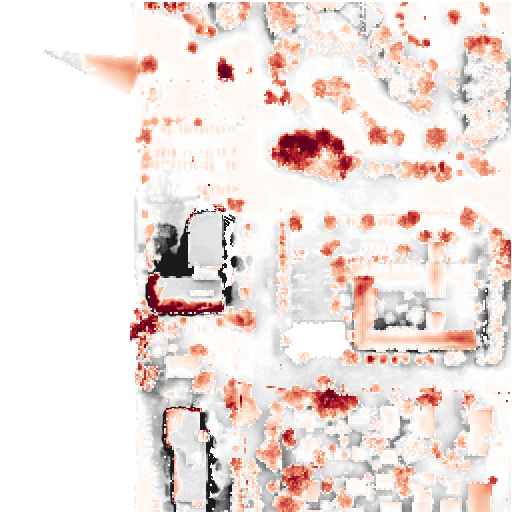} &
\includegraphics[width=0.11\linewidth, valign=m]{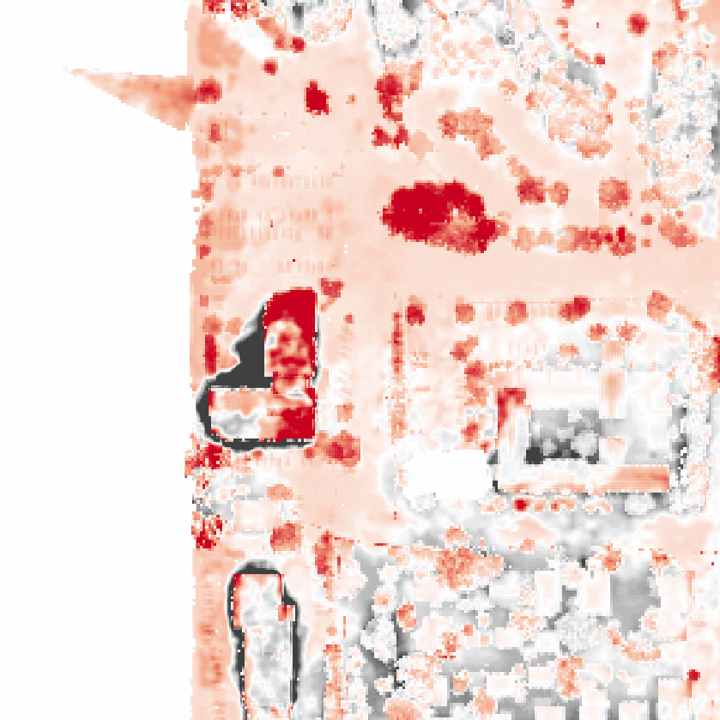} &
\includegraphics[width=0.11\linewidth, valign=m]{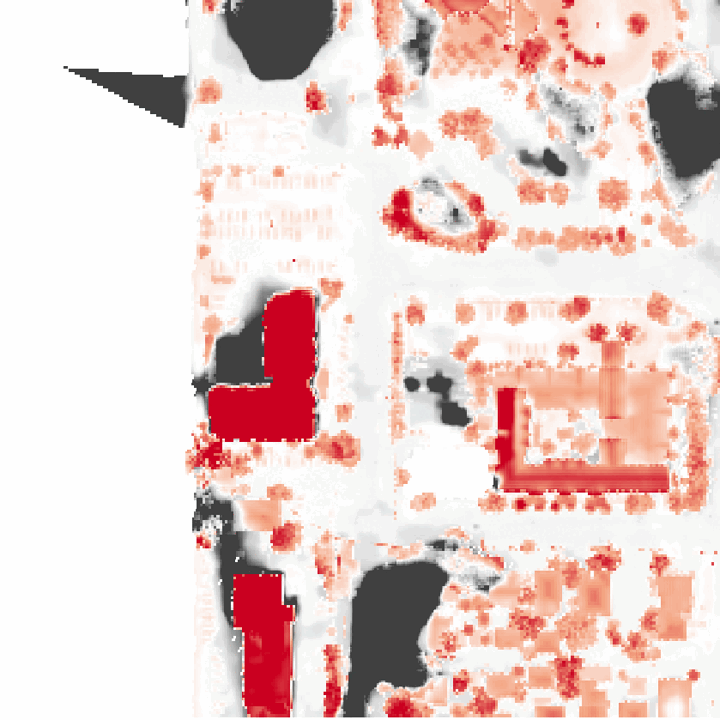} &
\includegraphics[width=0.11\linewidth, valign=m]{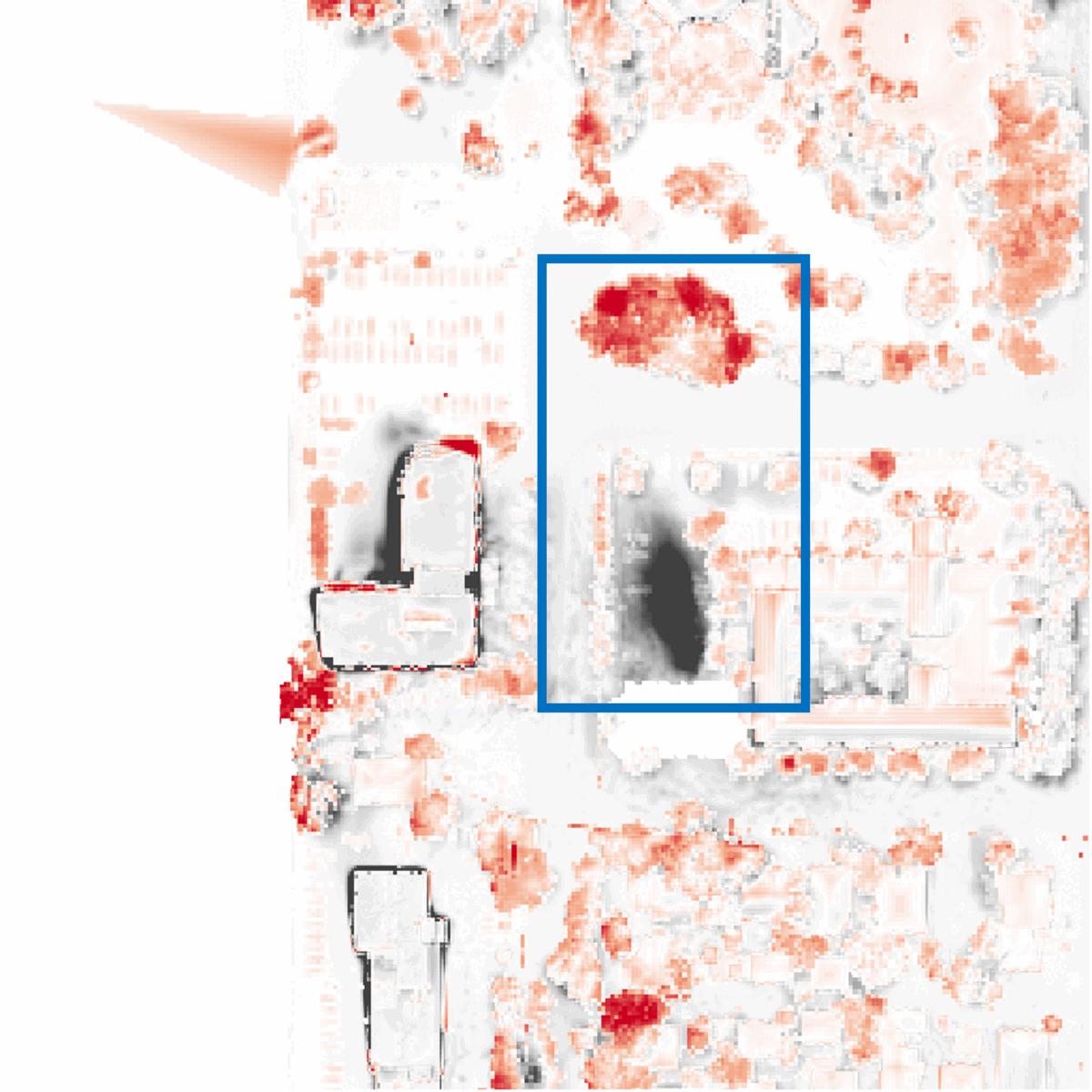} &
\includegraphics[width=0.11\linewidth, valign=m]{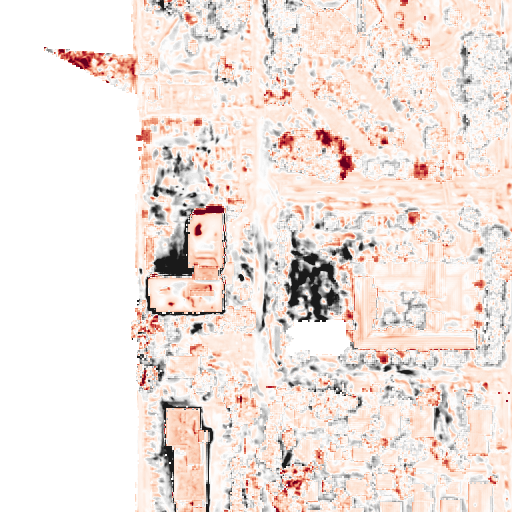} &
\includegraphics[width=0.11\linewidth, valign=m]{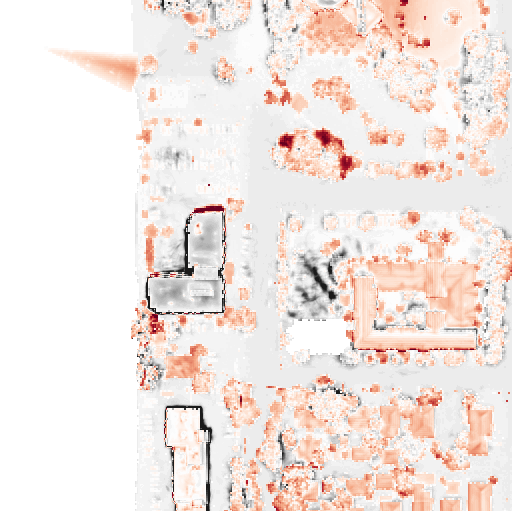} &
\includegraphics[width=0.11\linewidth, valign=m]{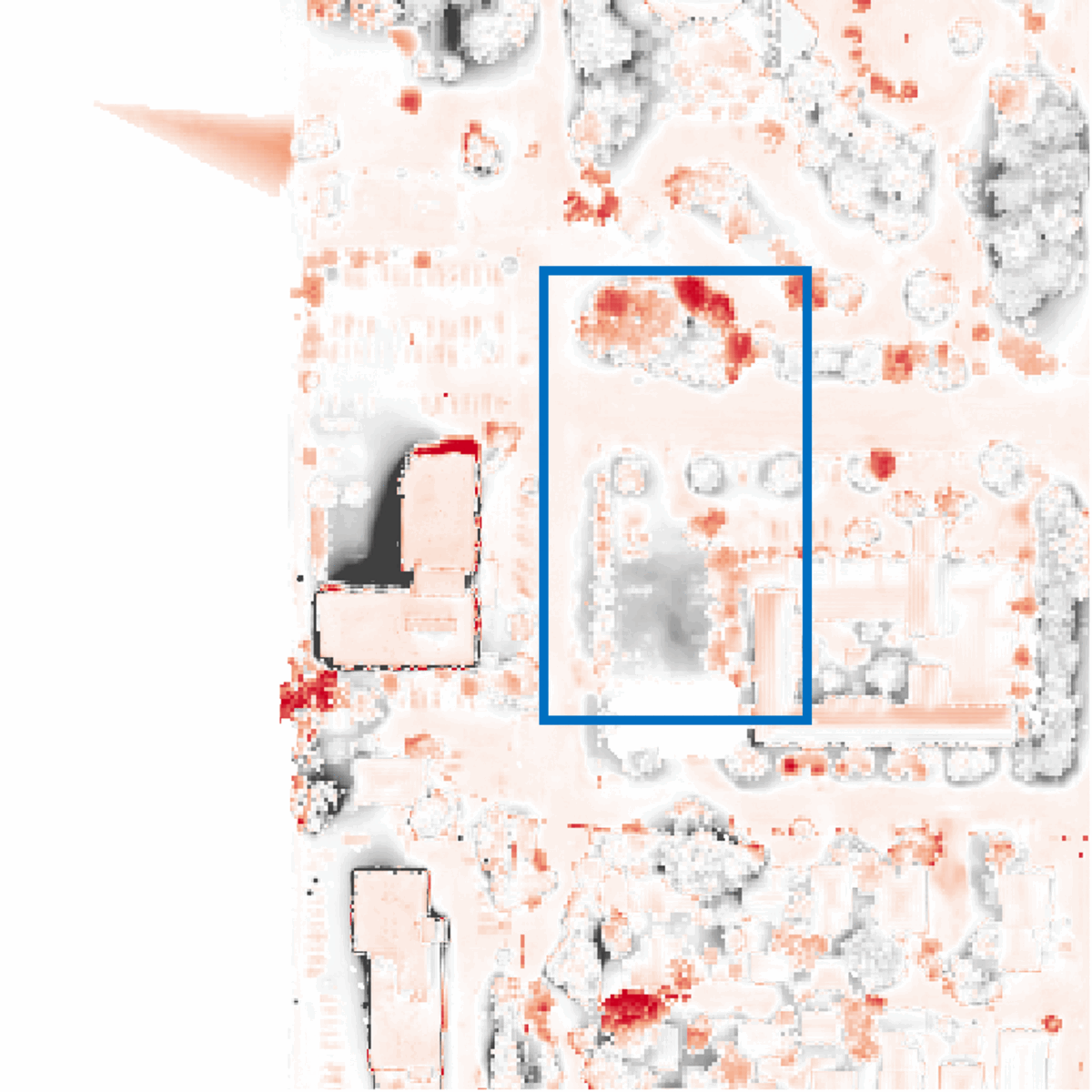} \\[2pt]

\leftlabeljax{JAX-168} &
\includegraphics[width=0.11\linewidth, valign=m]{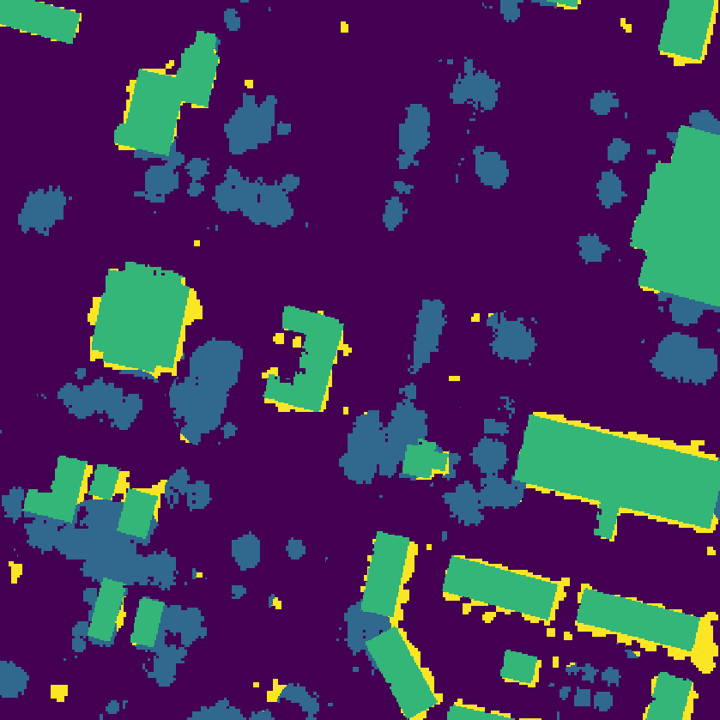} &
\includegraphics[width=0.11\linewidth, valign=m]{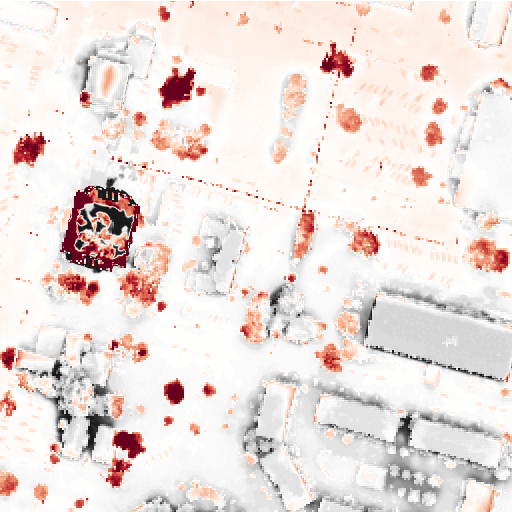} &
\includegraphics[width=0.11\linewidth, valign=m]{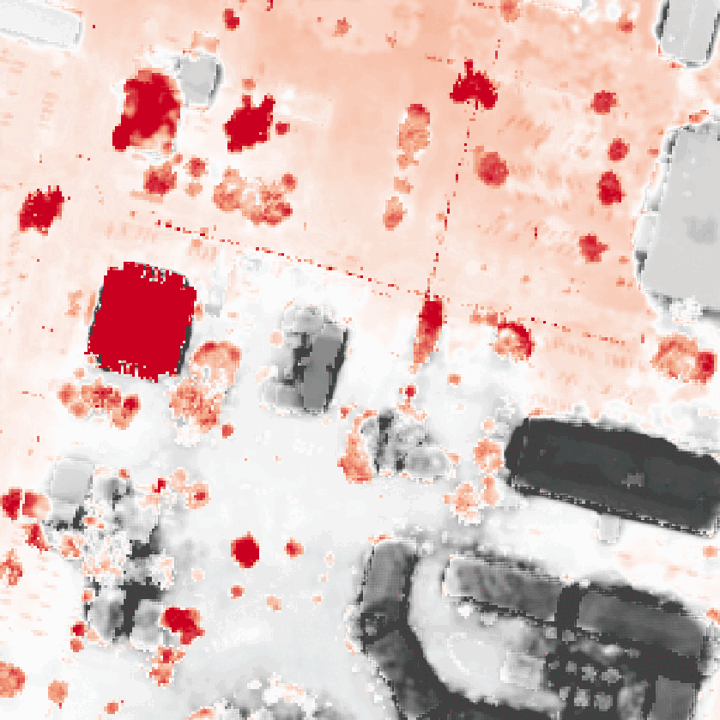} &
\includegraphics[width=0.11\linewidth, valign=m]{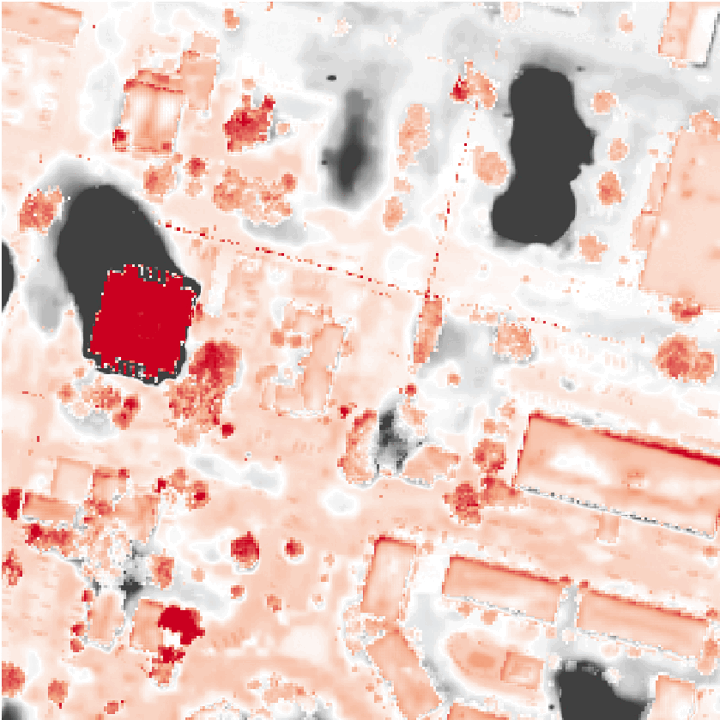} &
\includegraphics[width=0.11\linewidth, valign=m]{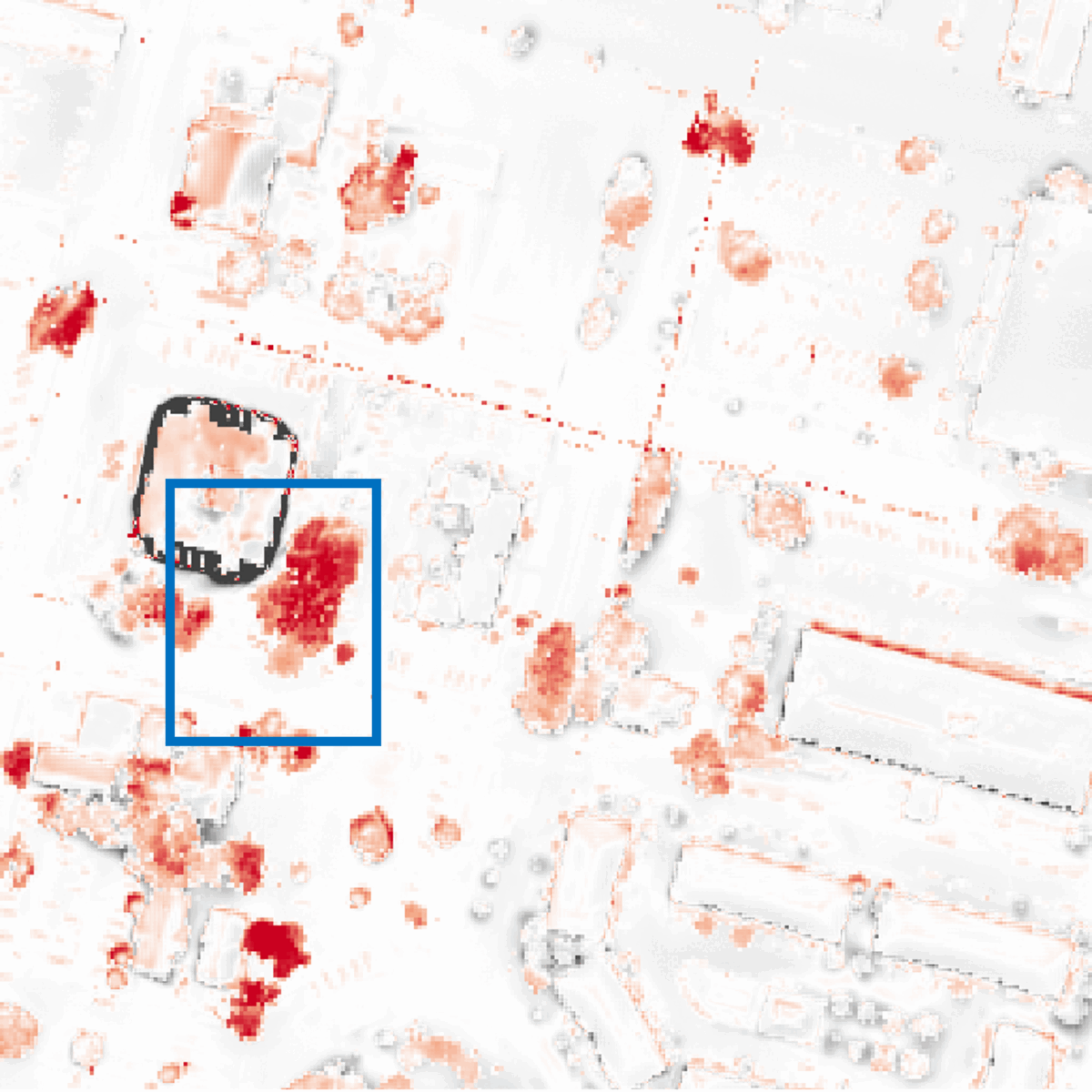} &
\includegraphics[width=0.11\linewidth, valign=m]{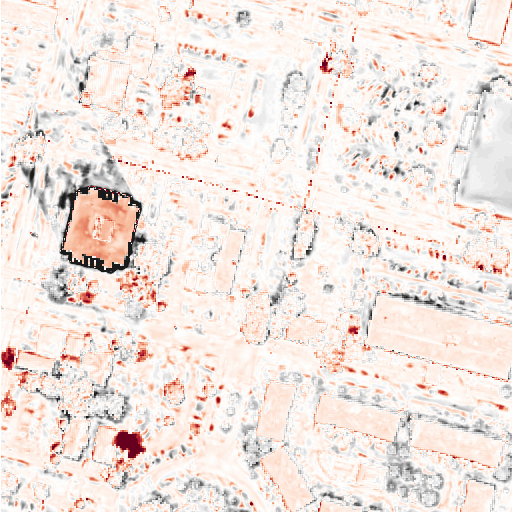} &
\includegraphics[width=0.11\linewidth, valign=m]{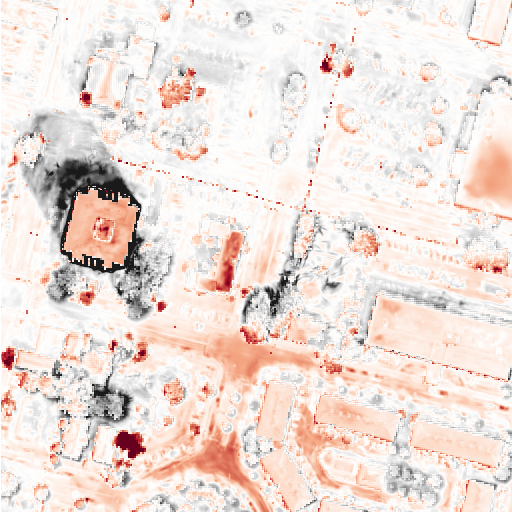} &
\includegraphics[width=0.11\linewidth, valign=m]{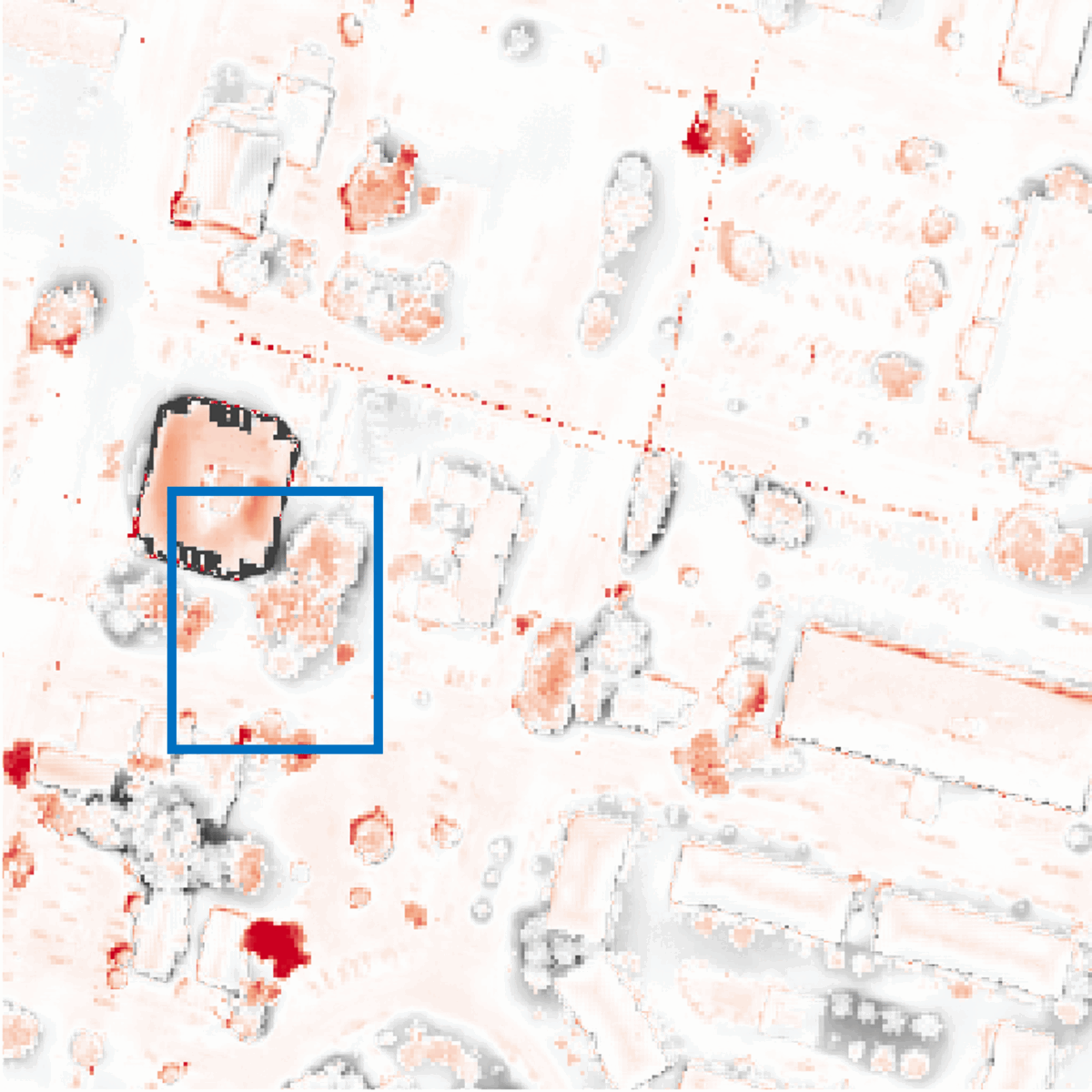} \\[2pt]

\leftlabeljax{JAX-251} &
\includegraphics[width=0.11\linewidth, valign=m]{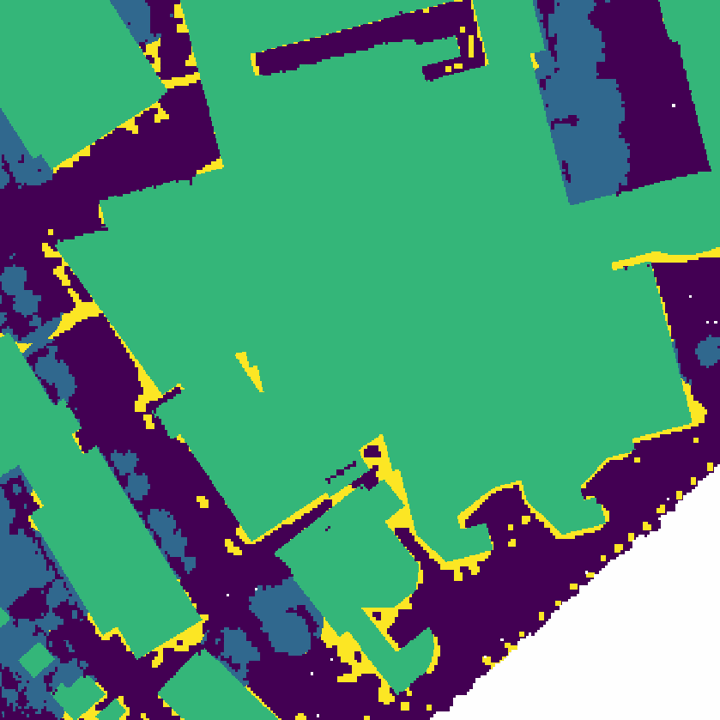} &
\includegraphics[width=0.11\linewidth, valign=m]{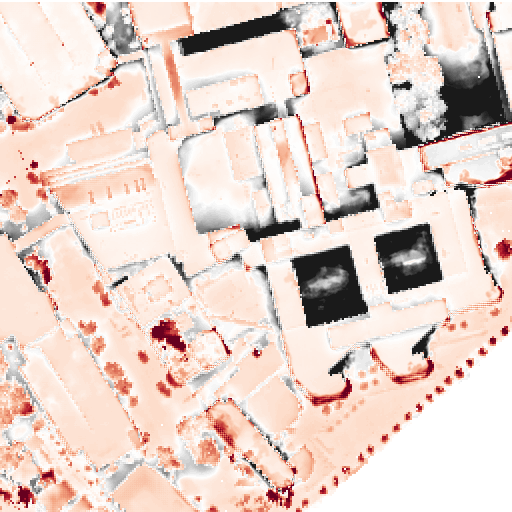} &
\includegraphics[width=0.11\linewidth, valign=m]{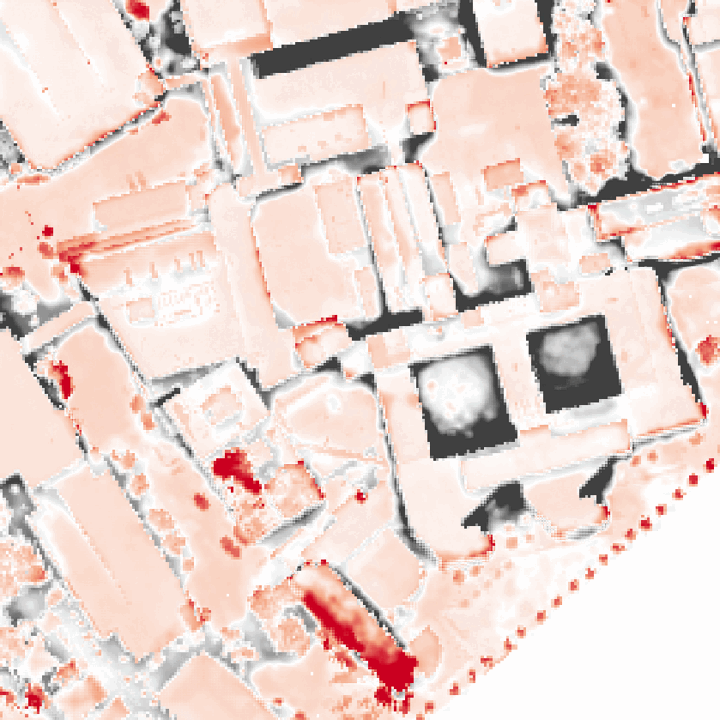} &
\includegraphics[width=0.11\linewidth, valign=m]{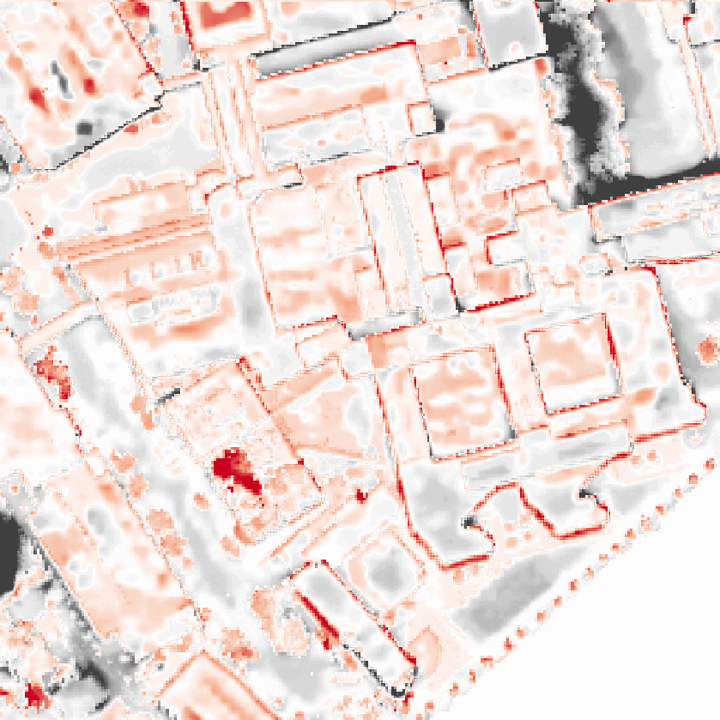} &
\includegraphics[width=0.11\linewidth, valign=m]{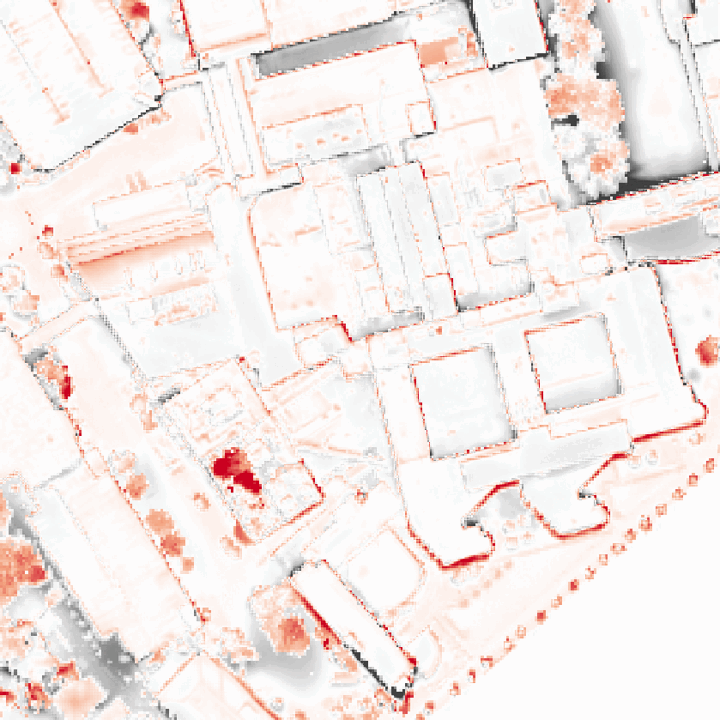} &
\includegraphics[width=0.11\linewidth, valign=m]{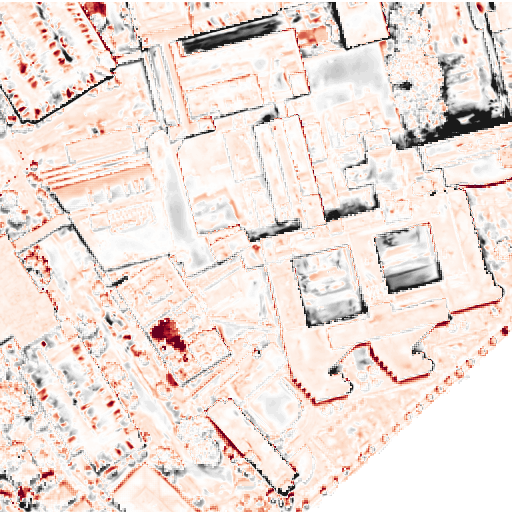} &
\includegraphics[width=0.11\linewidth, valign=m]{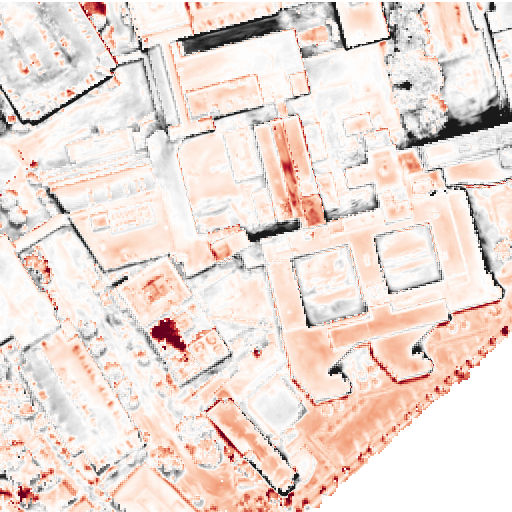} &
\includegraphics[width=0.11\linewidth, valign=m]{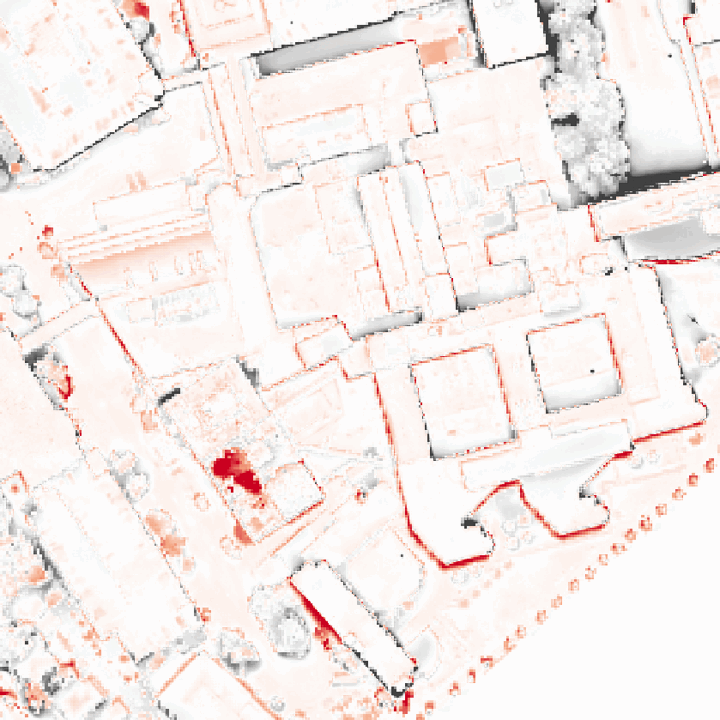} \\[2pt]

\leftlabeljax{JAX-280} &
\includegraphics[width=0.11\linewidth, valign=m]{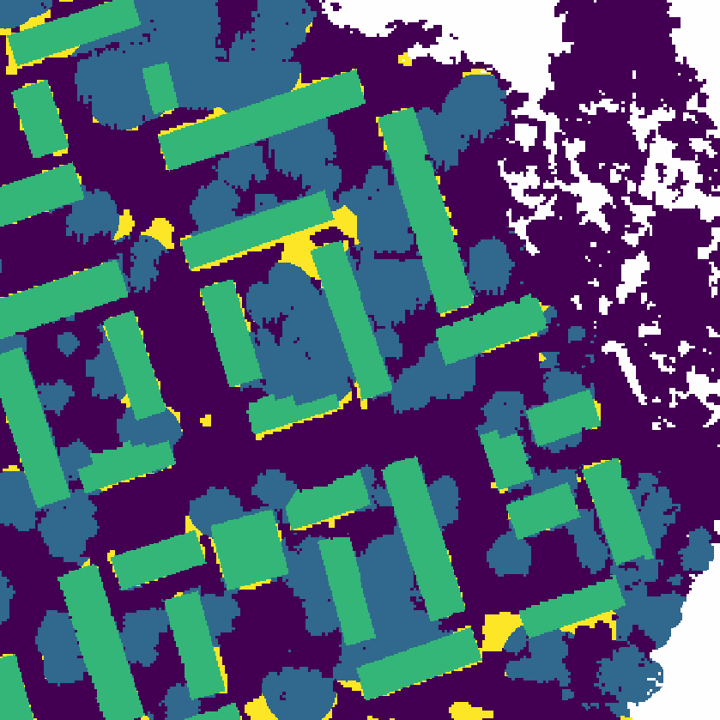} &
\includegraphics[width=0.11\linewidth, valign=m]{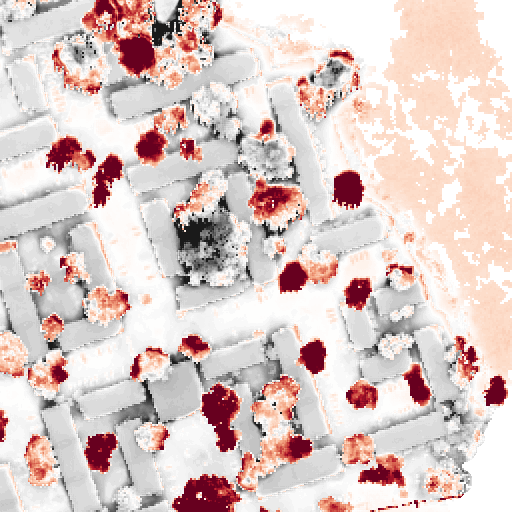} &
\includegraphics[width=0.11\linewidth, valign=m]{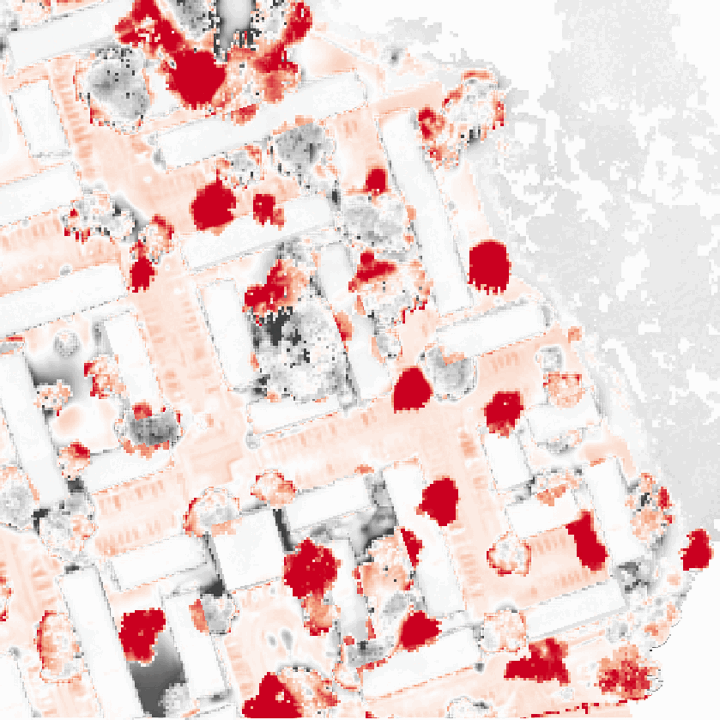} &
\includegraphics[width=0.11\linewidth, valign=m]{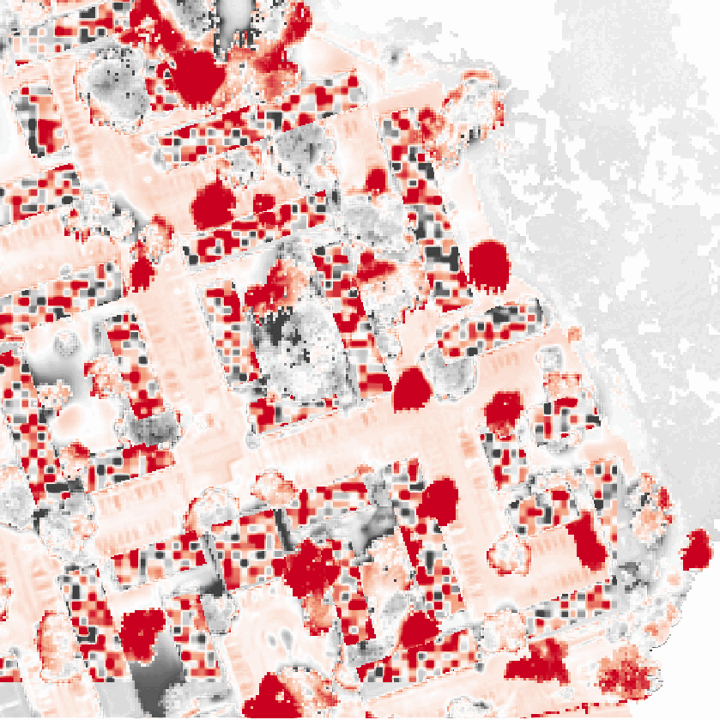} &
\includegraphics[width=0.11\linewidth, valign=m]{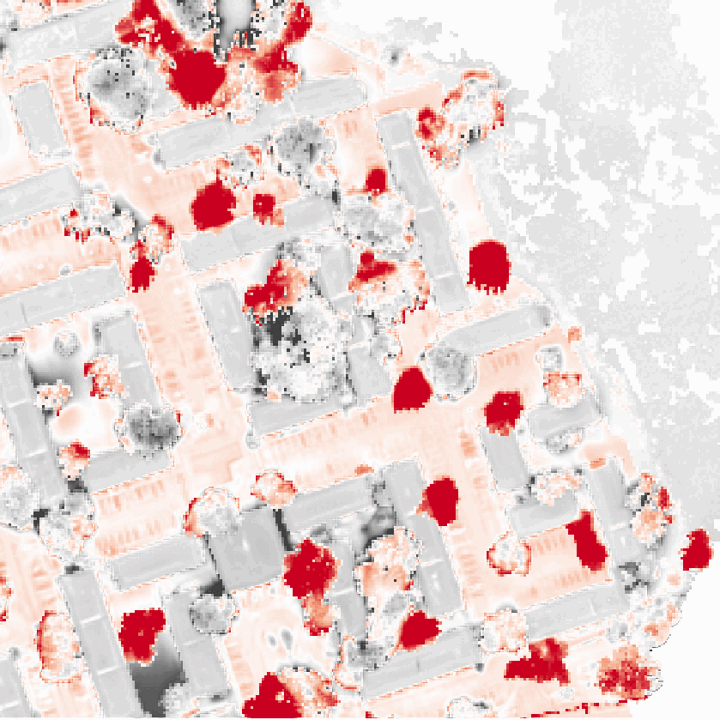} &
\includegraphics[width=0.11\linewidth, valign=m]{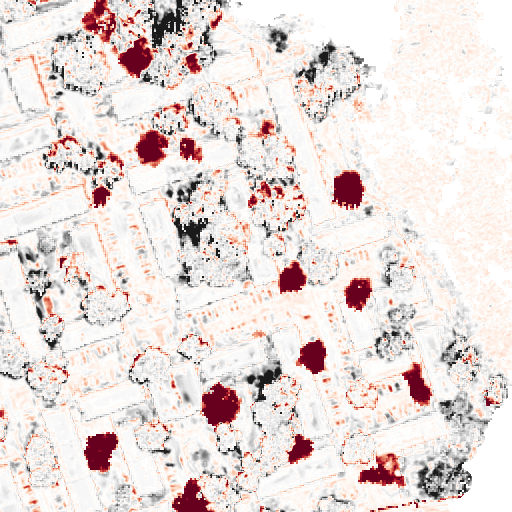} &
\includegraphics[width=0.11\linewidth, valign=m]{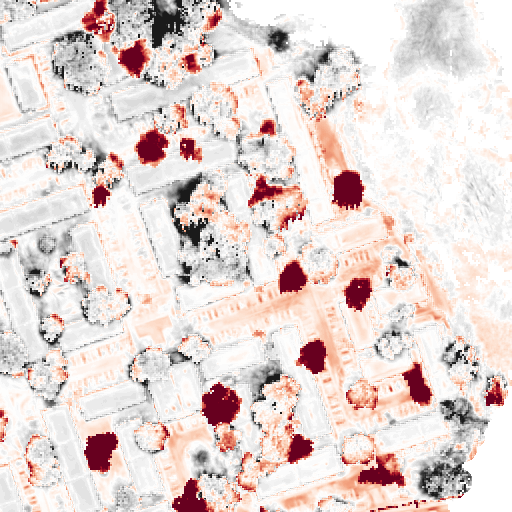} &
\includegraphics[width=0.11\linewidth, valign=m]{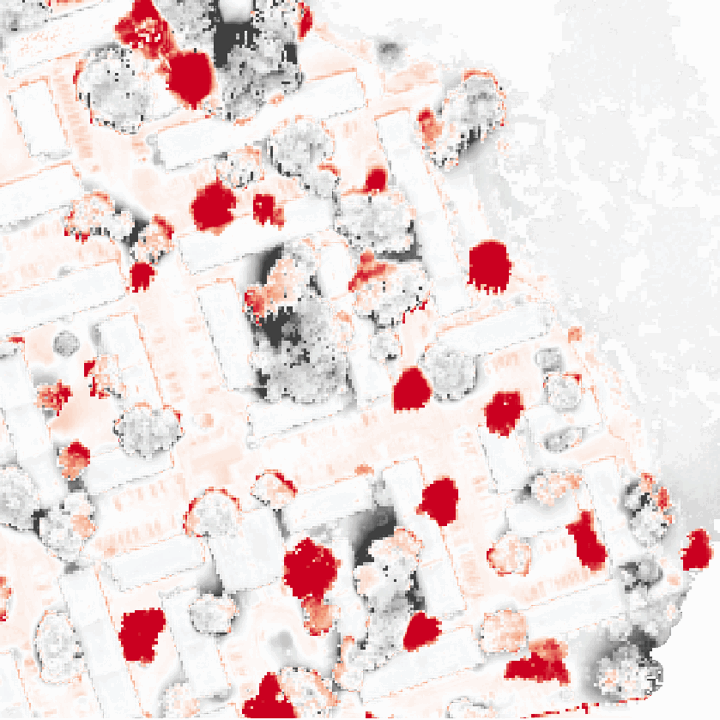} \\
& \small CLS & \small ASP & \small s2p & \small SAT-NGP & \small EOGS & \small Skyfall-1 & \small Skyfall-2 & \small SatSplat
\end{tabular}

\vspace{6pt}
\noindent
\begin{minipage}{0.15\linewidth}
\centering\includegraphics[width=\linewidth]{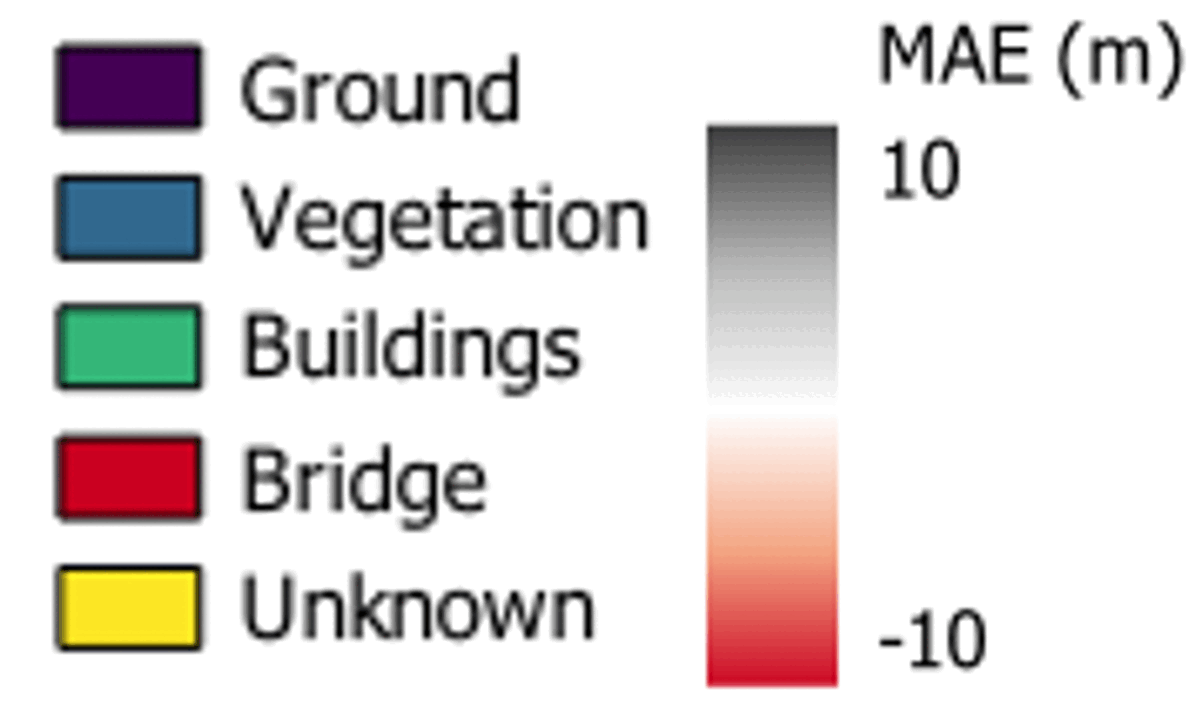}
\end{minipage}
\hfill
\begin{minipage}{0.82\linewidth}
\caption{Errormap comparison for Jacksonville (JAX) tiles. The errormap visualizes height residuals in the range of $-10$ to $10$\,m. For CLS, green indicates buildings, teal indicates vegetation, black represents ground, and white corresponds to water regions.}
\label{fig:errmap_jax}
\end{minipage}
\end{figure*}

\begin{figure*}[!htbp]
\centering
\newcommand{\leftlabelother}[1]{\adjustbox{valign=m}{\rotatebox{90}{\small #1}}}
\setlength{\tabcolsep}{1.5pt}
\begin{tabular}{m{7pt}cccccccc}
\renewcommand{\arraystretch}{1.1}
\leftlabelother{OMA-203} &
\includegraphics[width=0.11\linewidth, valign=m]{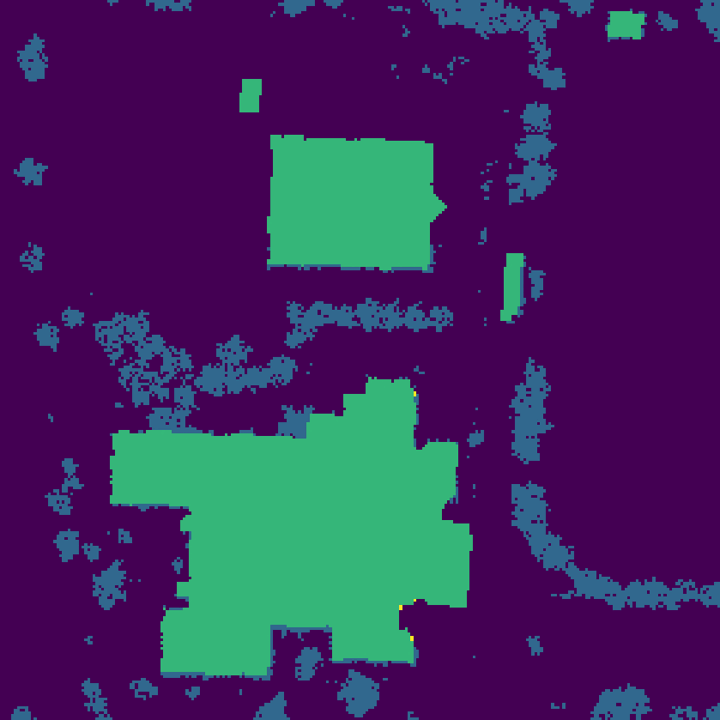} &
\includegraphics[width=0.11\linewidth, valign=m]{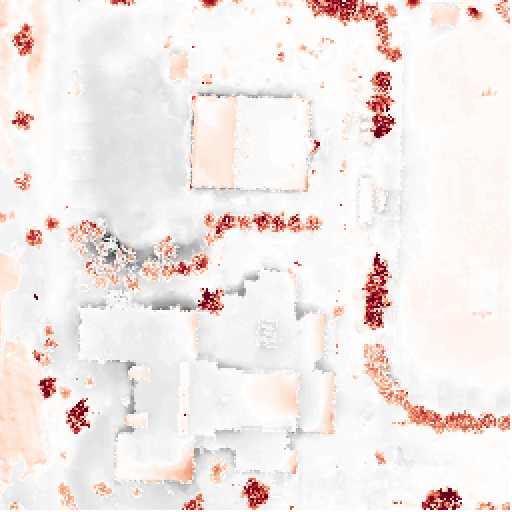} &
\includegraphics[width=0.11\linewidth, valign=m]{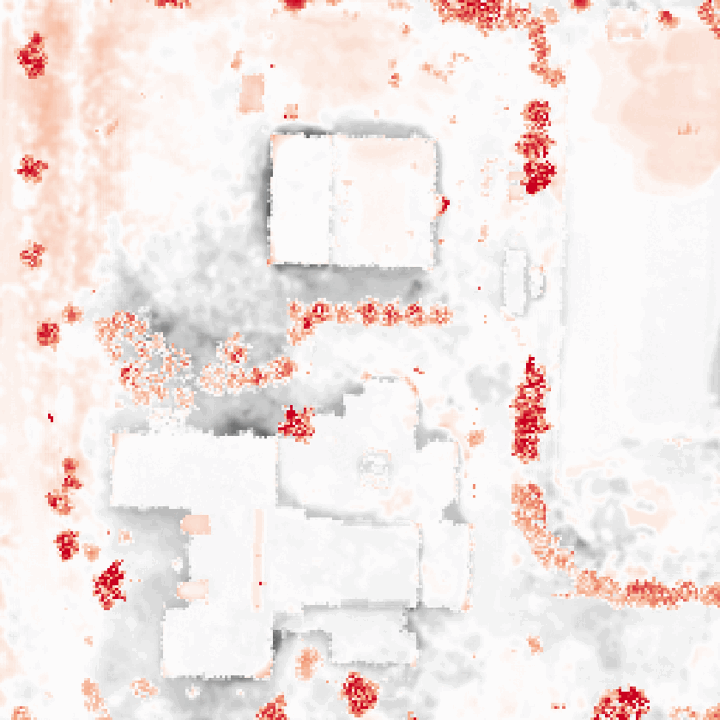} &
\includegraphics[width=0.11\linewidth, valign=m]{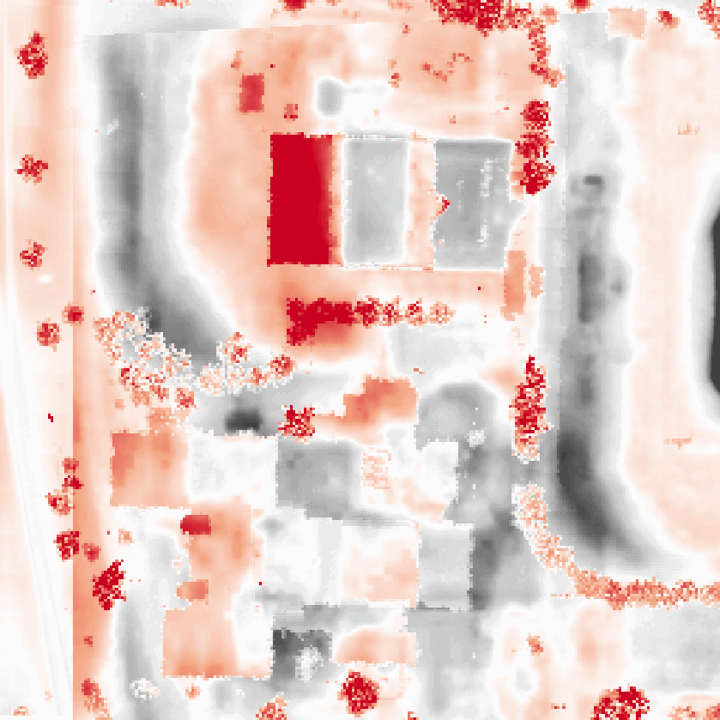} &
\includegraphics[width=0.11\linewidth, valign=m]{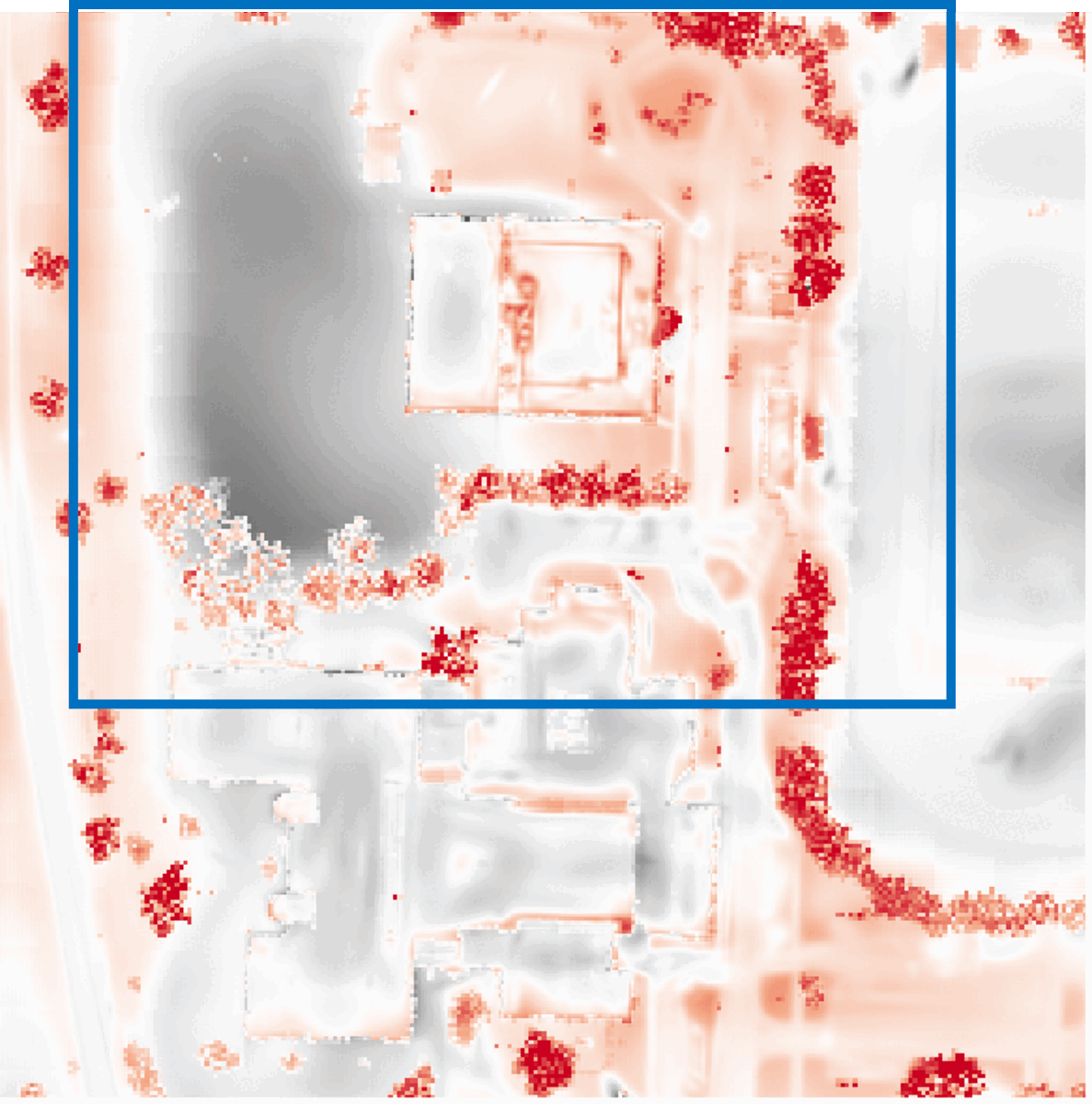} &
\includegraphics[width=0.11\linewidth, valign=m]{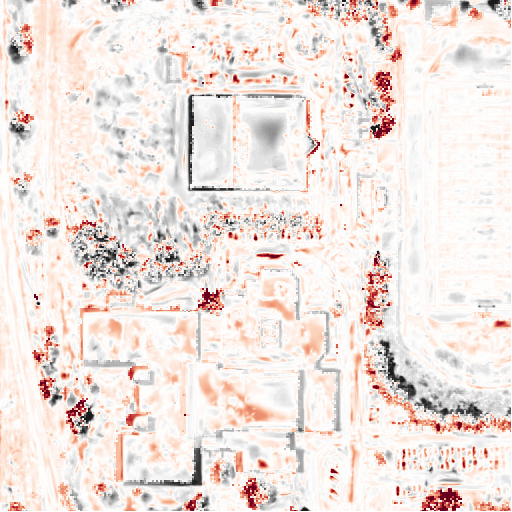} &
\includegraphics[width=0.11\linewidth, valign=m]{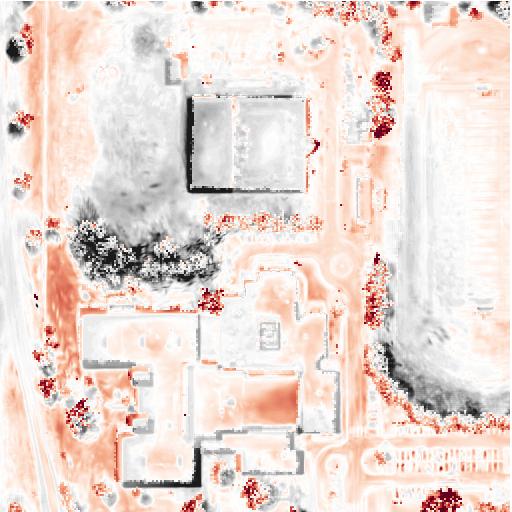} &
\includegraphics[width=0.11\linewidth, valign=m]{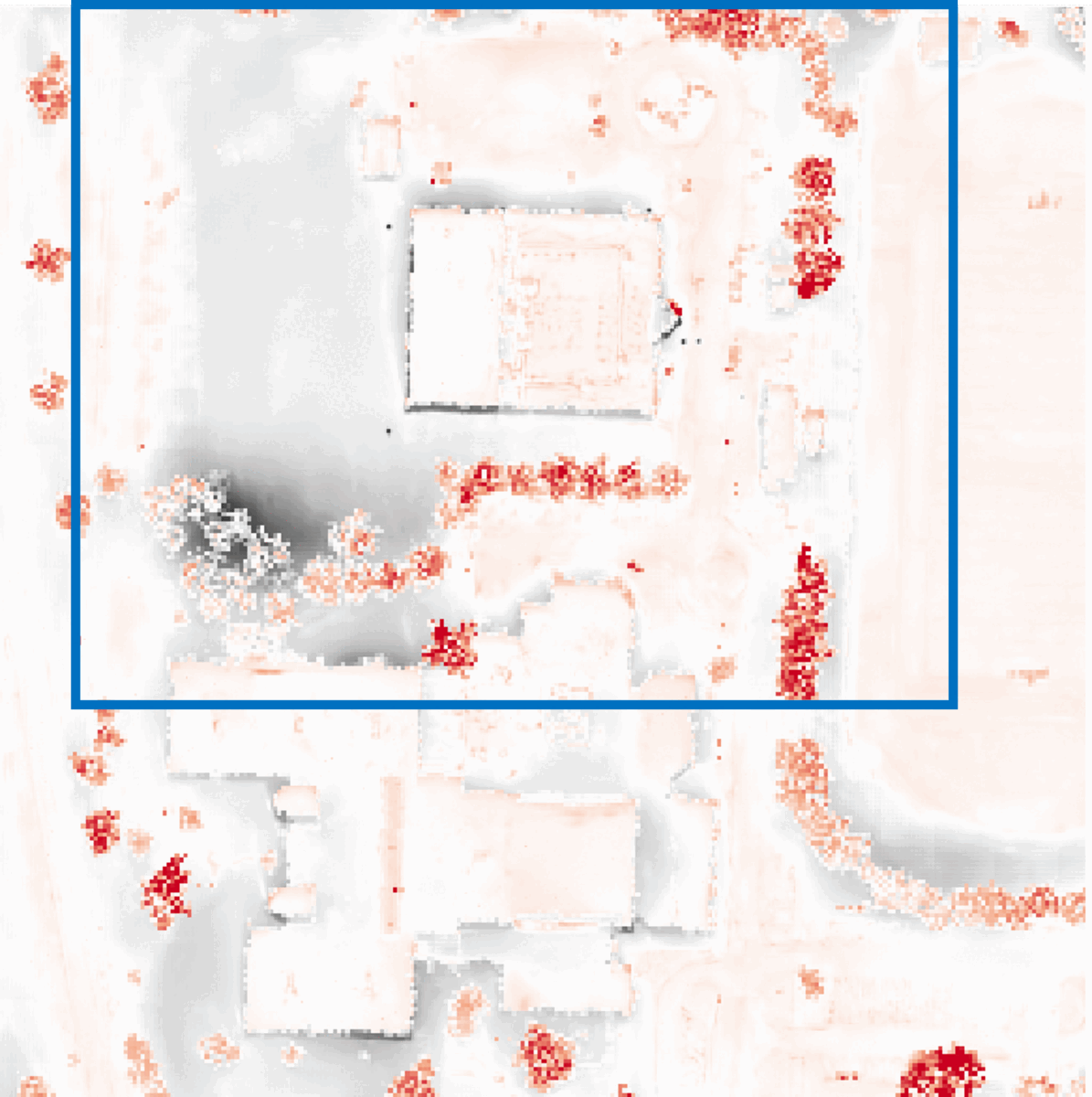} \\[2pt]

\leftlabelother{OMA-212} &
\includegraphics[width=0.11\linewidth, valign=m]{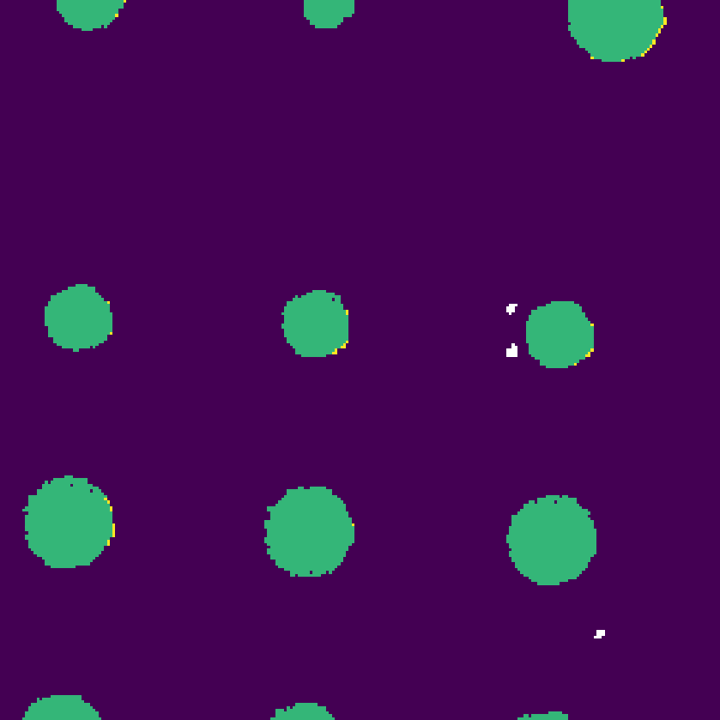} &
\includegraphics[width=0.11\linewidth, valign=m]{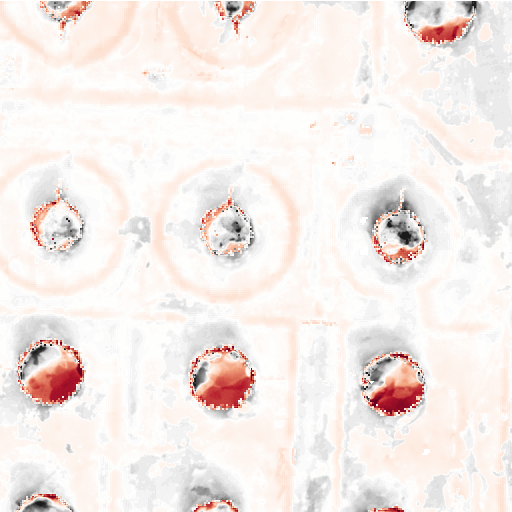} &
\includegraphics[width=0.11\linewidth, valign=m]{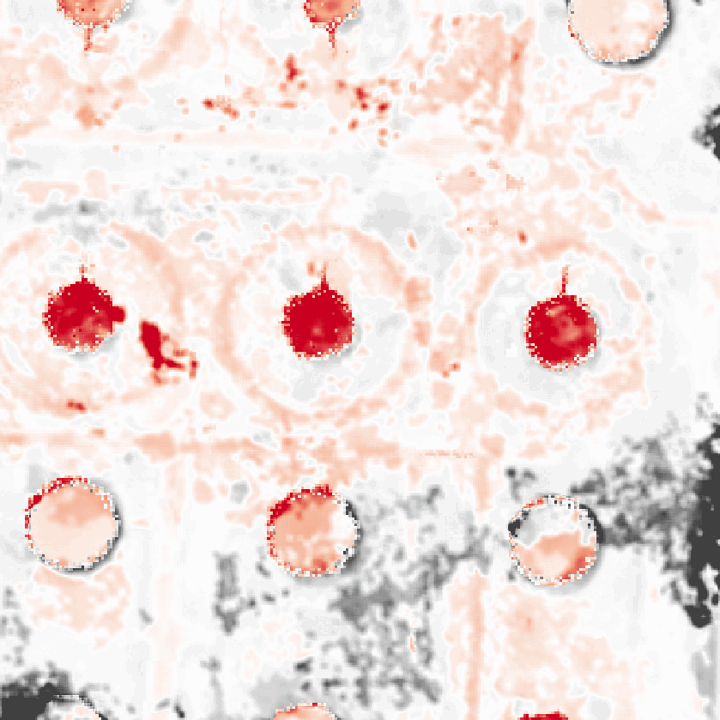} &
\includegraphics[width=0.11\linewidth, valign=m]{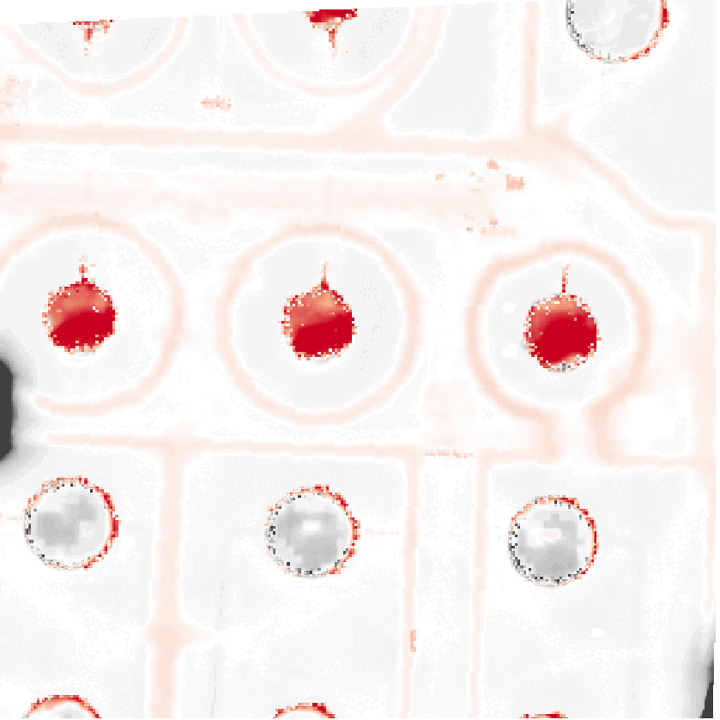} &
\includegraphics[width=0.11\linewidth, valign=m]{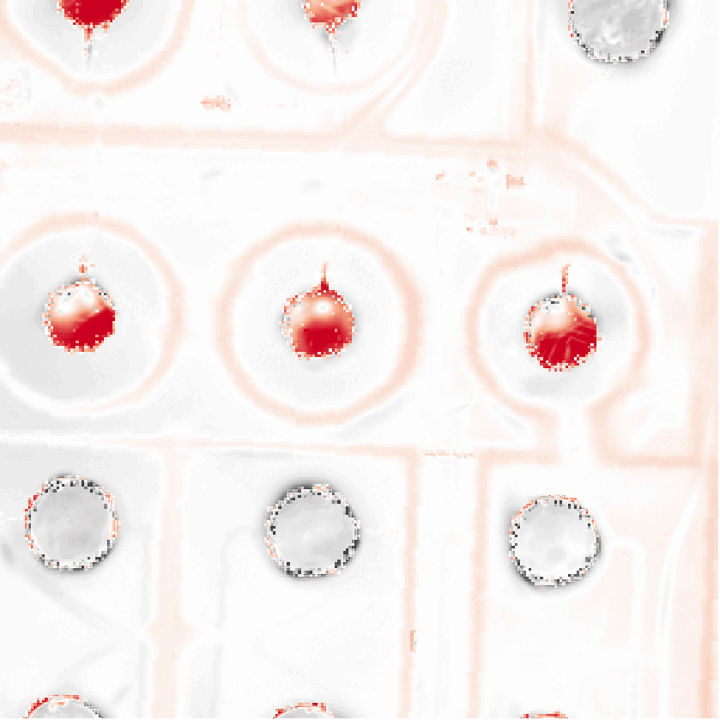} &
\includegraphics[width=0.11\linewidth, valign=m]{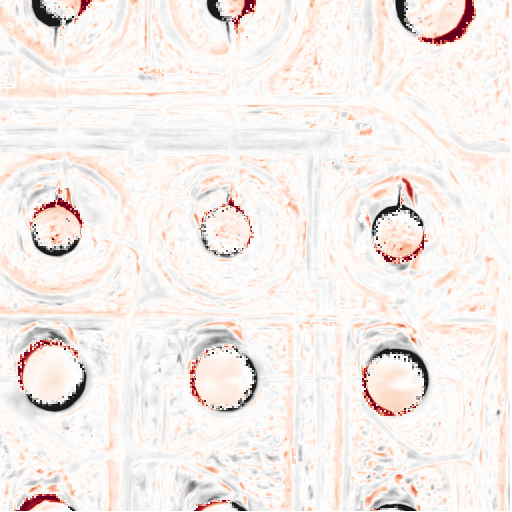} &
\includegraphics[width=0.11\linewidth, valign=m]{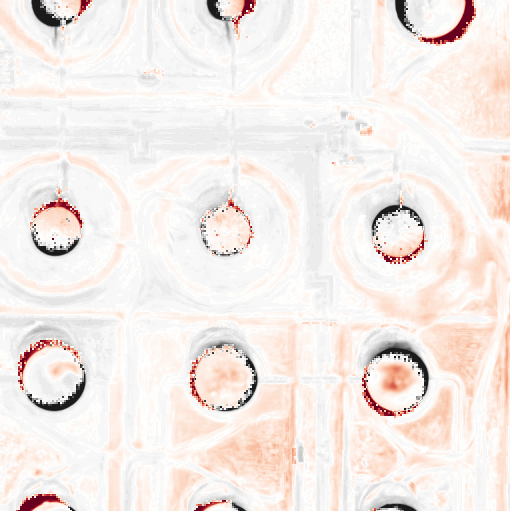} &
\includegraphics[width=0.11\linewidth, valign=m]{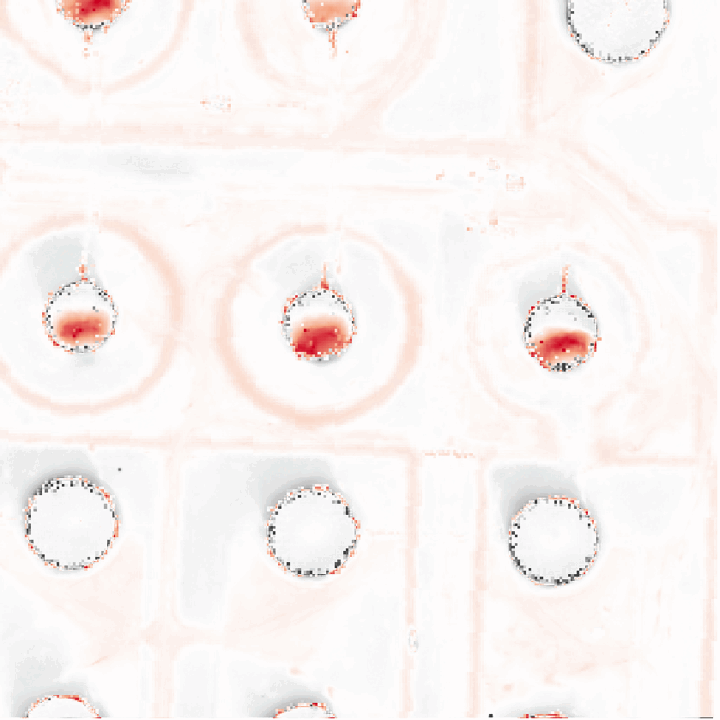} \\[2pt]

\leftlabelother{OMA-315} &
\includegraphics[width=0.11\linewidth, valign=m]{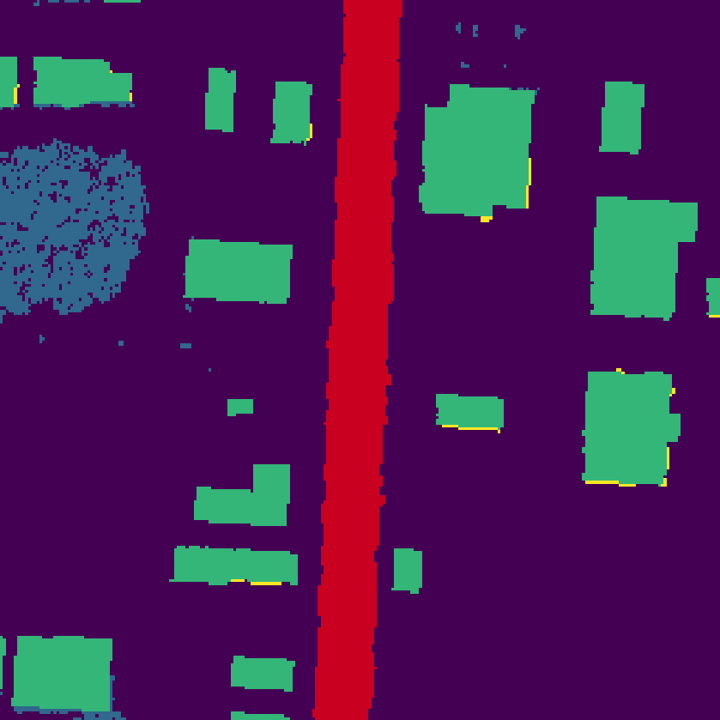} &
\includegraphics[width=0.11\linewidth, valign=m]{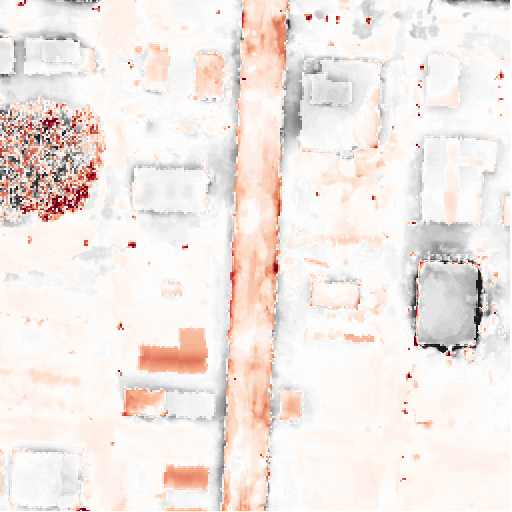} &
\includegraphics[width=0.11\linewidth, valign=m]{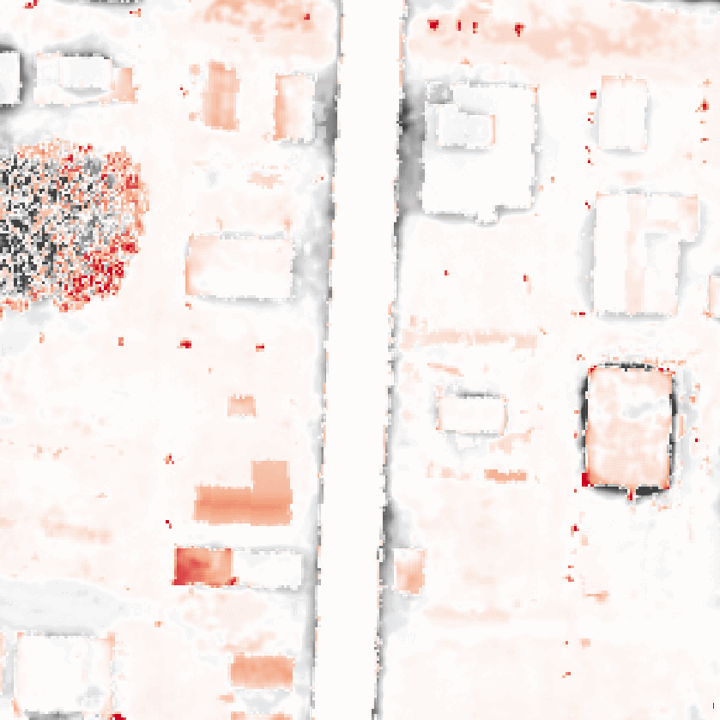} &
\includegraphics[width=0.11\linewidth, valign=m]{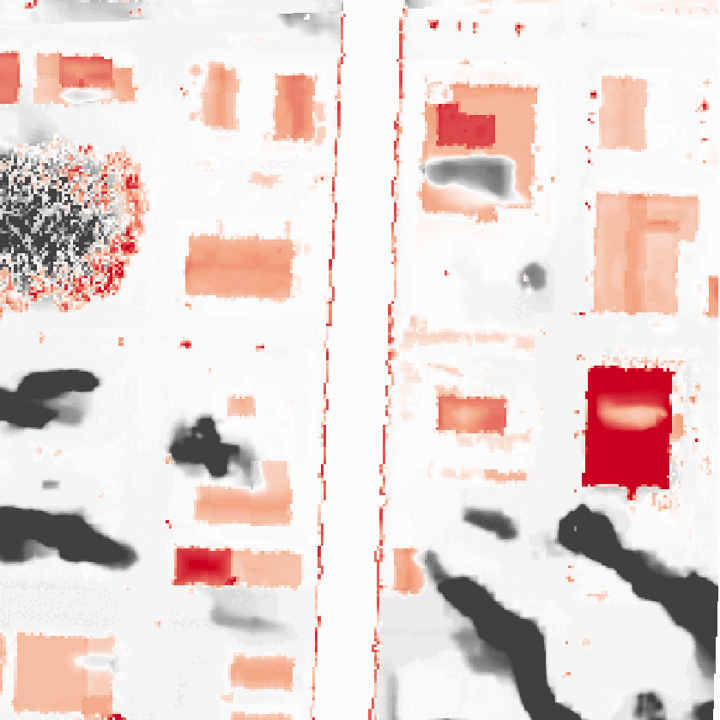} &
\includegraphics[width=0.11\linewidth, valign=m]{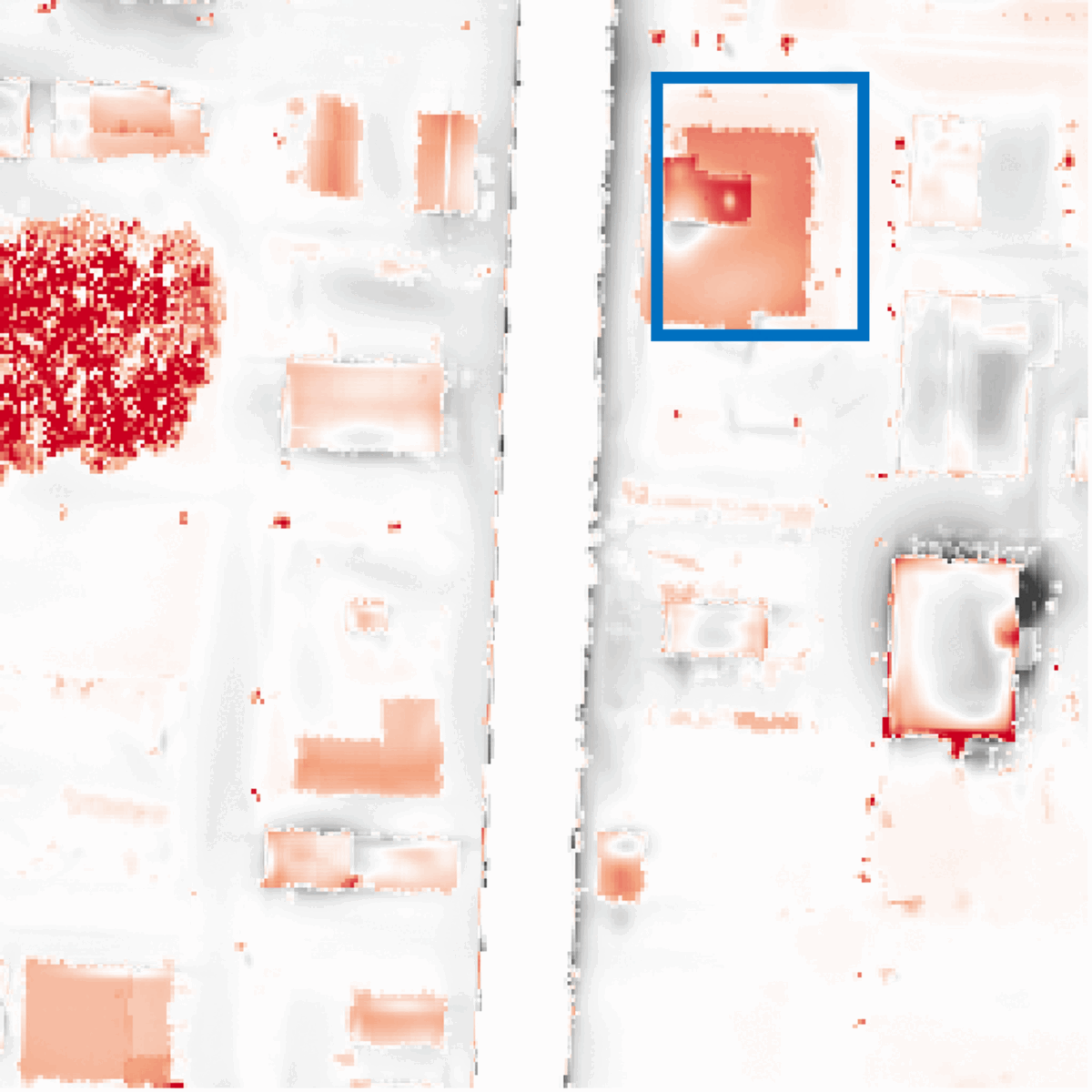} &
\includegraphics[width=0.11\linewidth, valign=m]{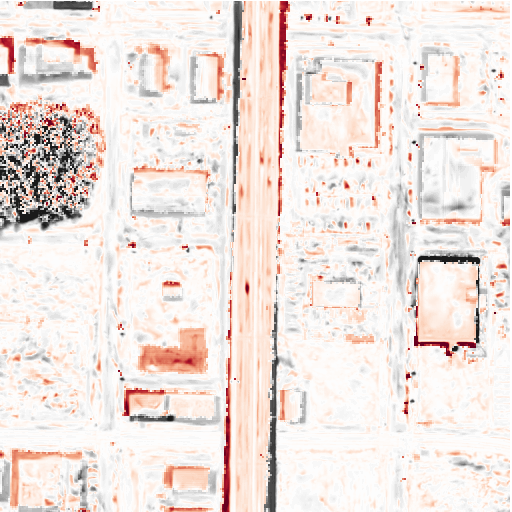} &
\includegraphics[width=0.11\linewidth, valign=m]{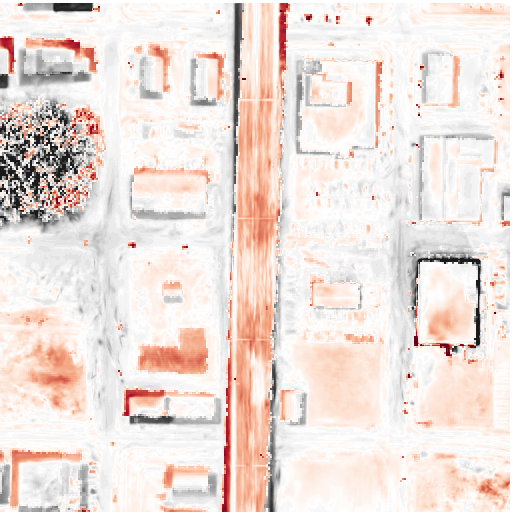} &
\includegraphics[width=0.11\linewidth, valign=m]{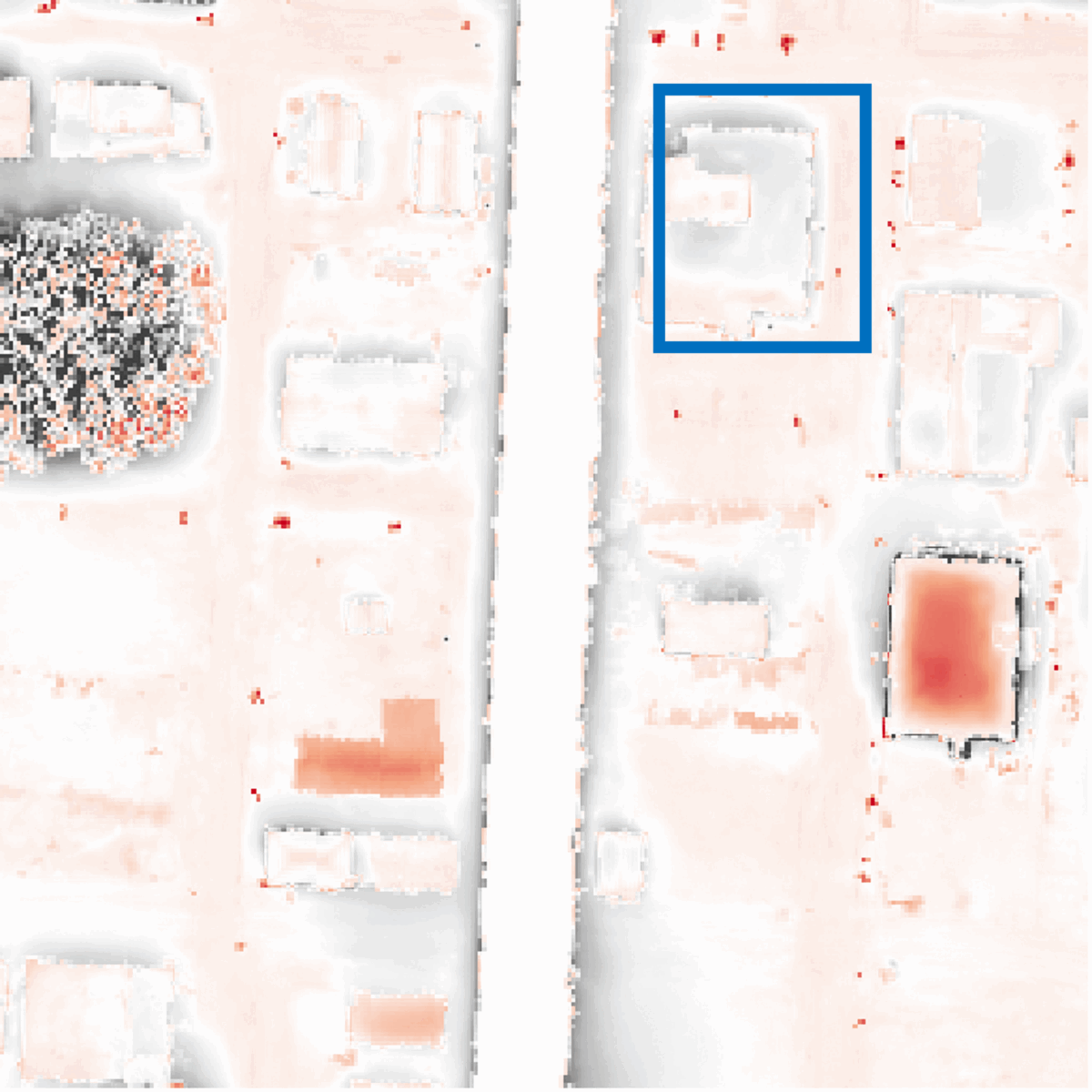} \\[2pt]

\leftlabelother{IARPA-001} &
\includegraphics[width=0.11\linewidth, valign=m]{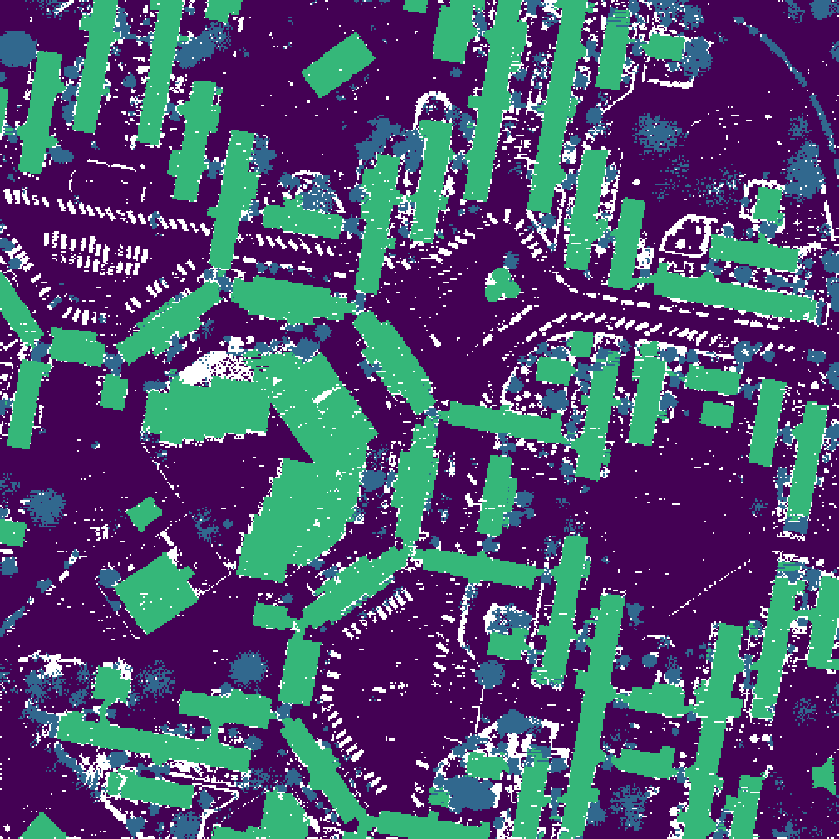} &
\includegraphics[width=0.11\linewidth, valign=m]{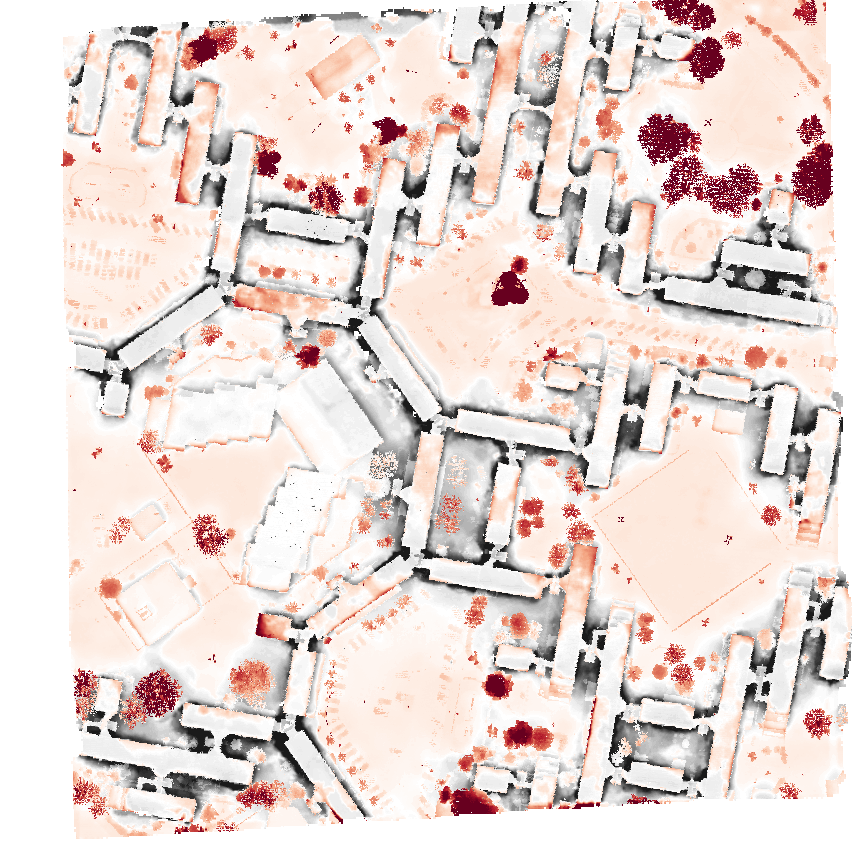} &
\includegraphics[width=0.11\linewidth, valign=m]{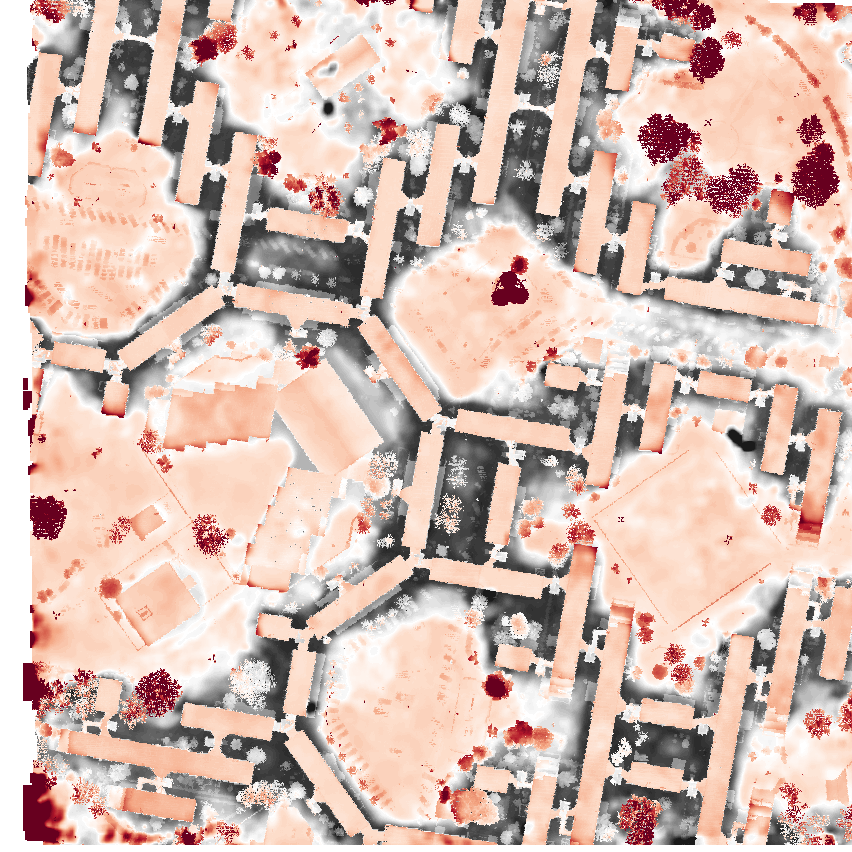} &
\includegraphics[width=0.11\linewidth, valign=m]{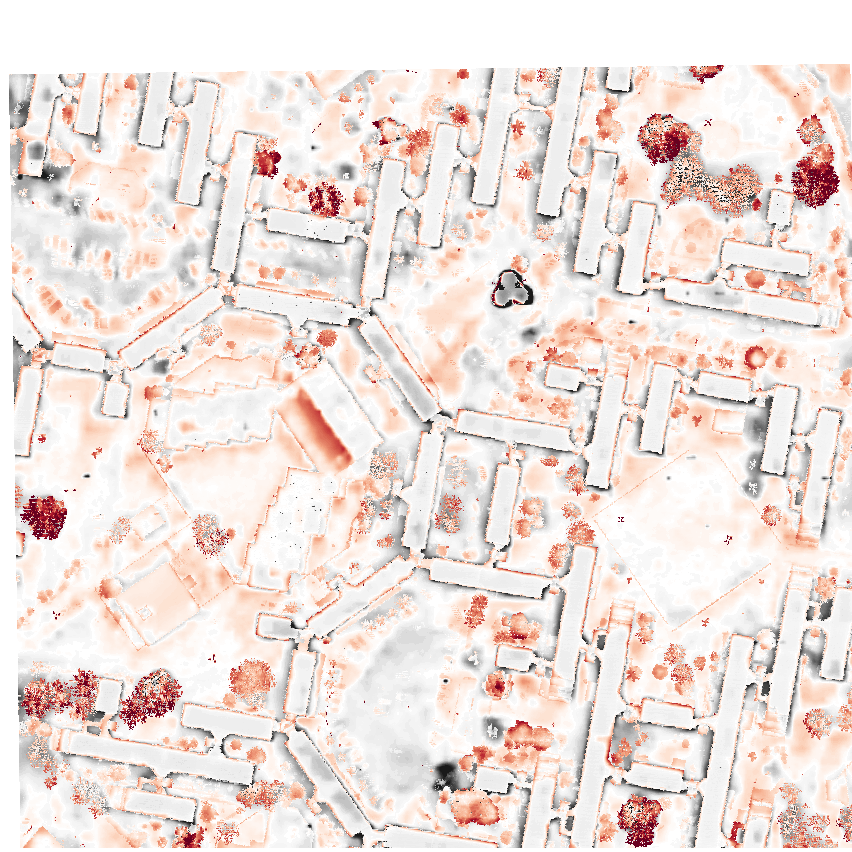} &
\includegraphics[width=0.11\linewidth, valign=m]{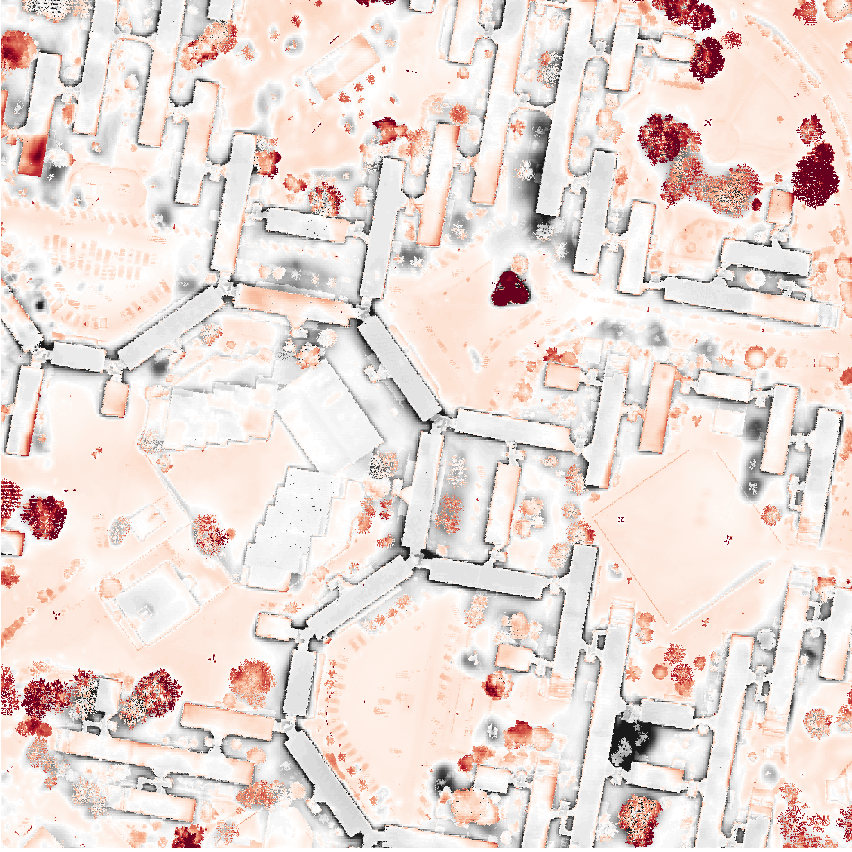} &
\includegraphics[width=0.11\linewidth, valign=m]{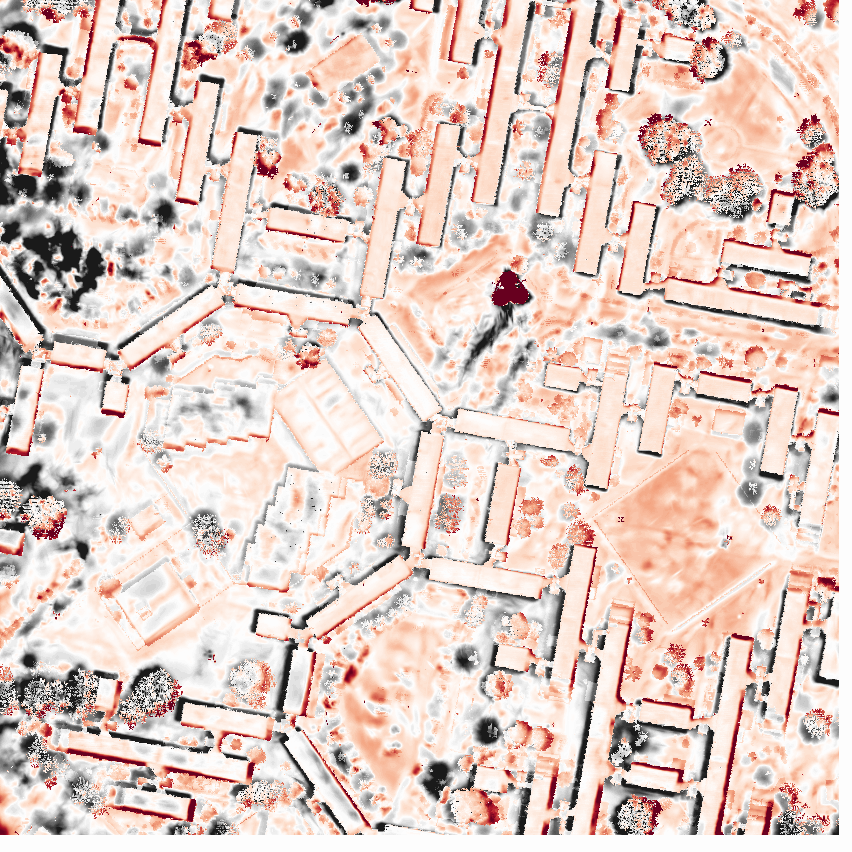} &
\includegraphics[width=0.11\linewidth, valign=m]{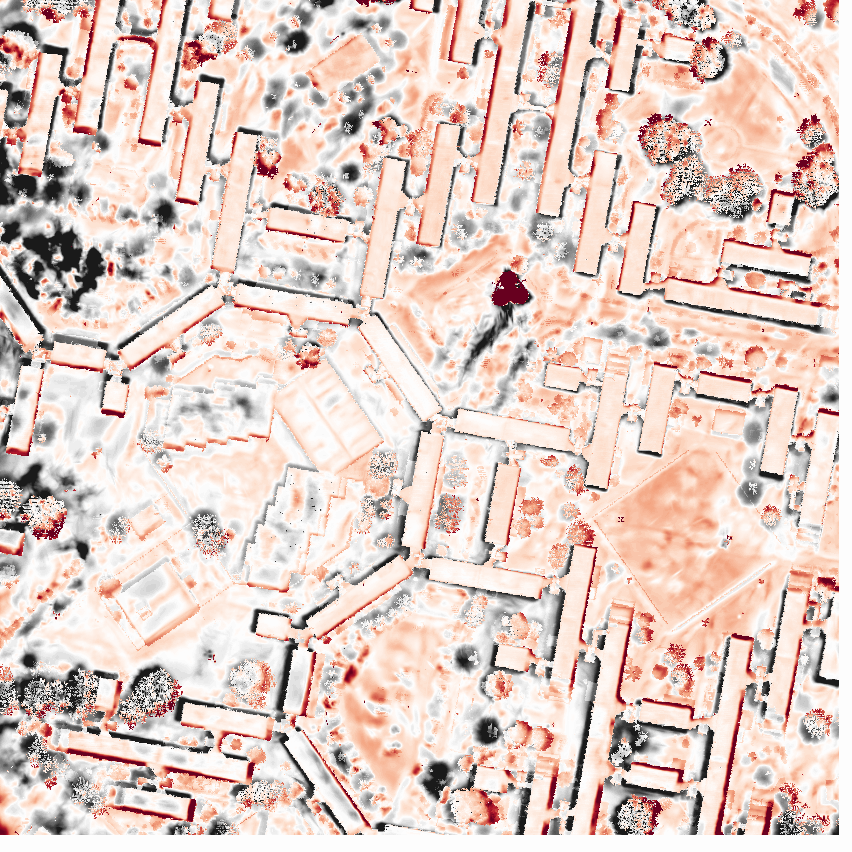} &
\includegraphics[width=0.11\linewidth, valign=m]{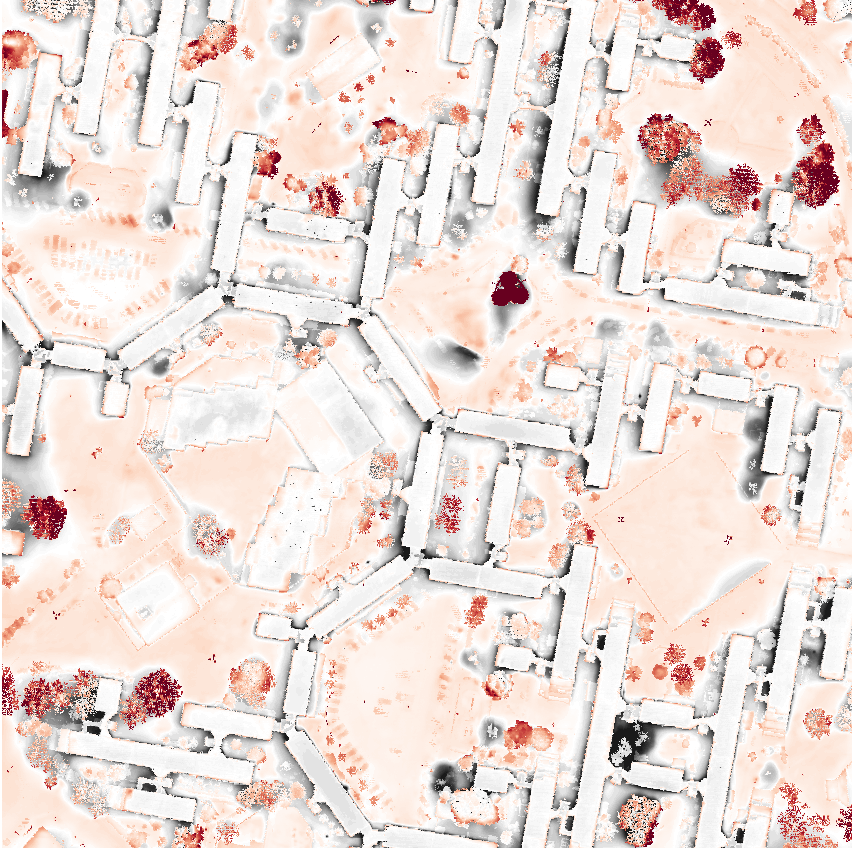} \\[2pt]

\leftlabelother{IARPA-002} &
\includegraphics[width=0.11\linewidth, valign=m]{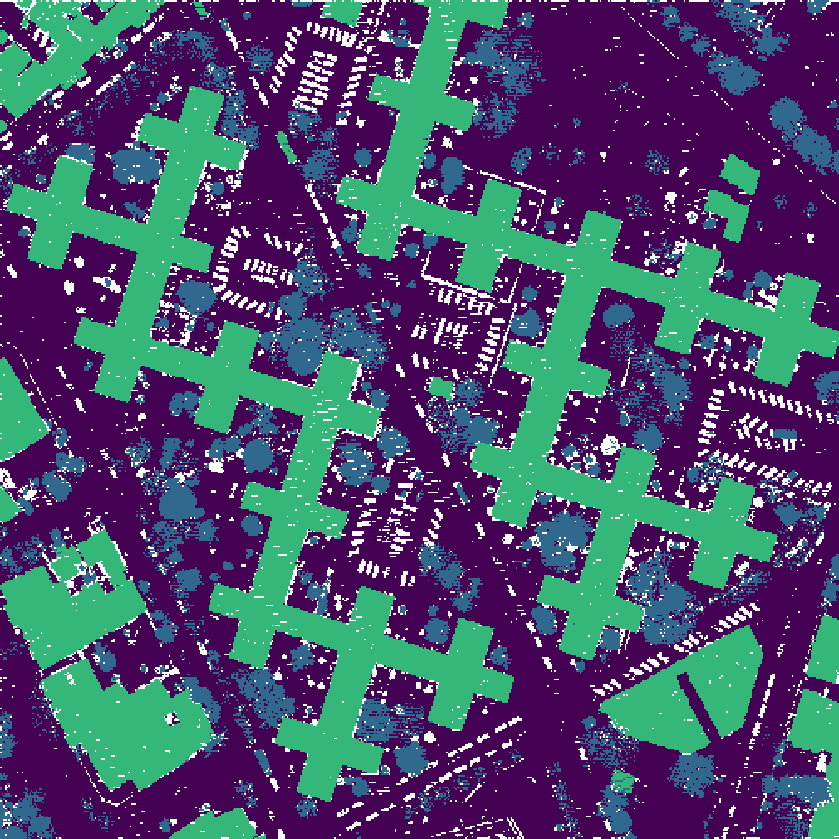} &
\includegraphics[width=0.11\linewidth, valign=m]{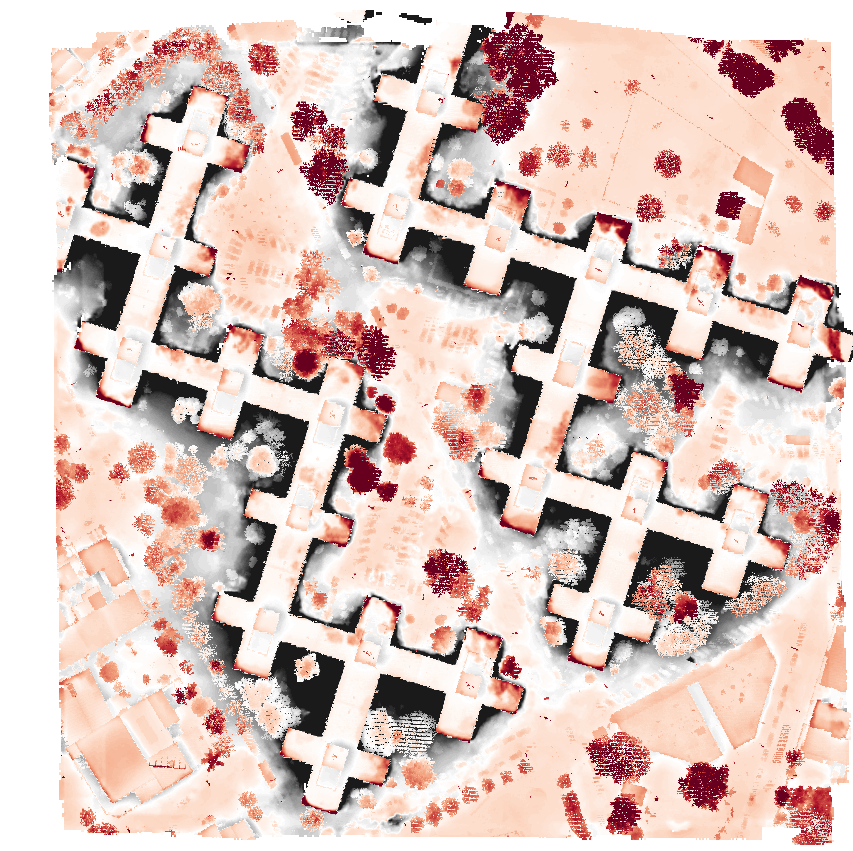} &
\includegraphics[width=0.11\linewidth, valign=m]{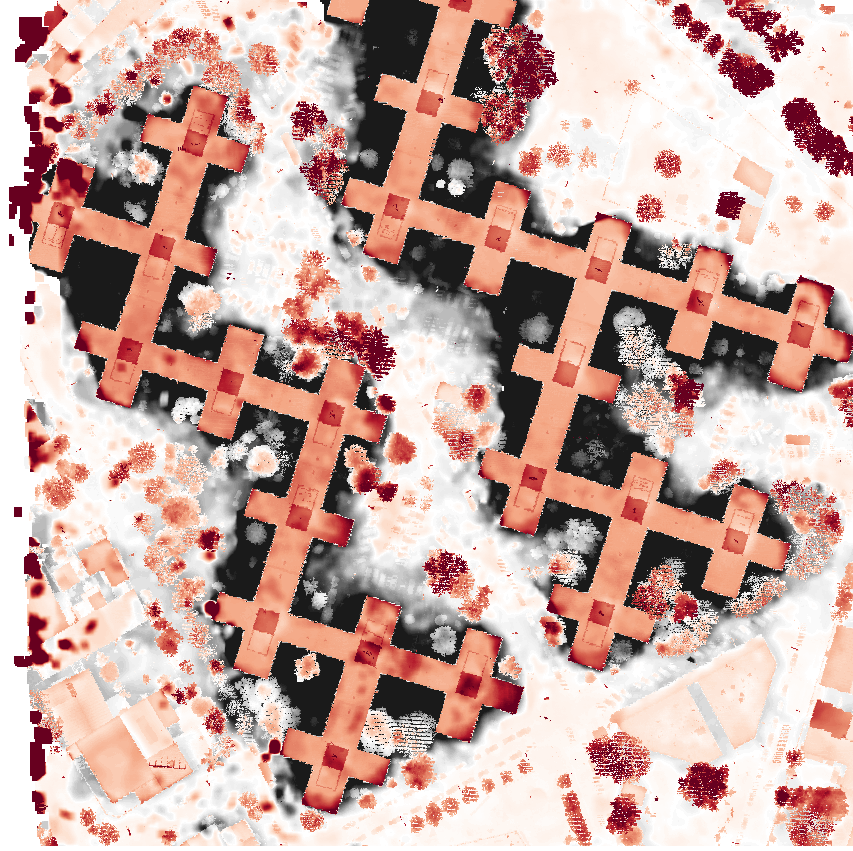} &
\includegraphics[width=0.11\linewidth, valign=m]{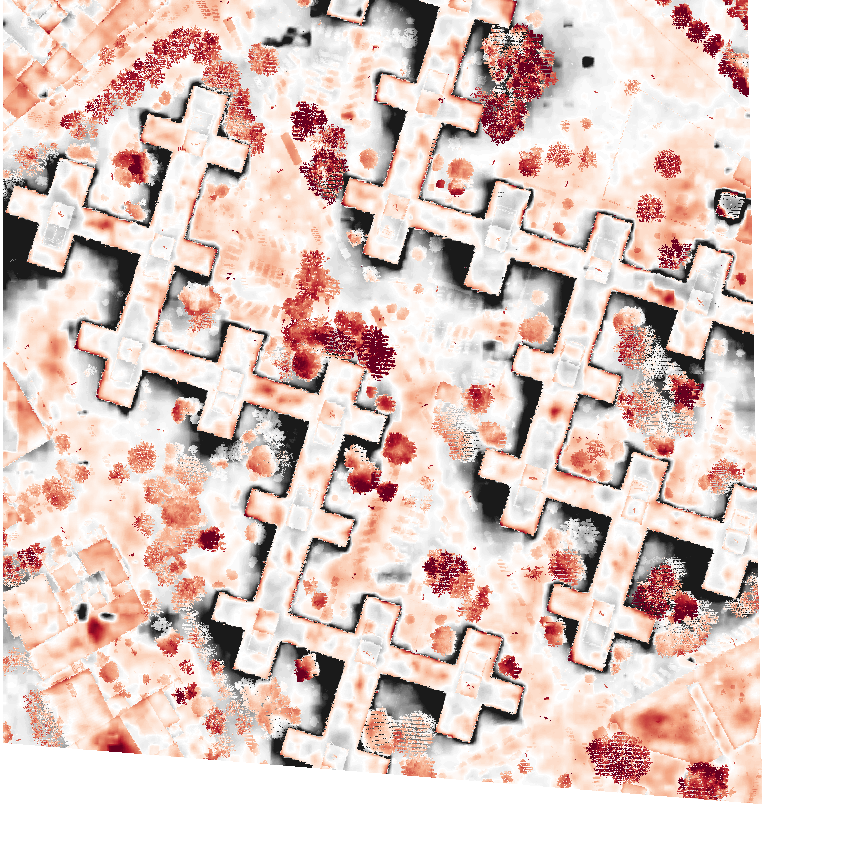} &
\includegraphics[width=0.11\linewidth, valign=m]{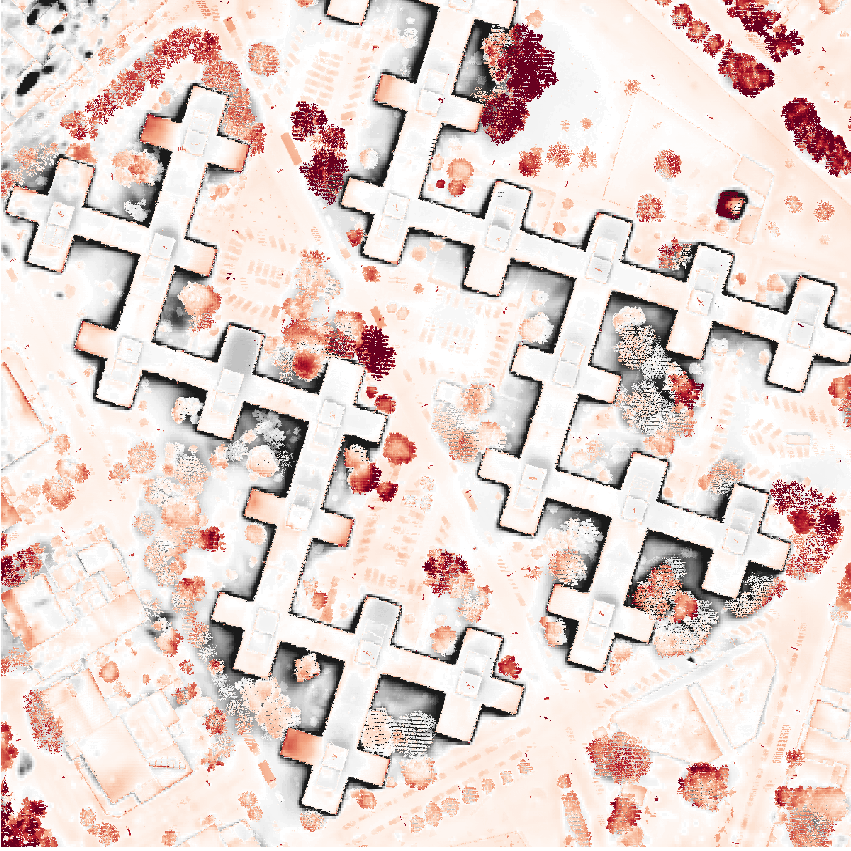} &
\includegraphics[width=0.11\linewidth, valign=m]{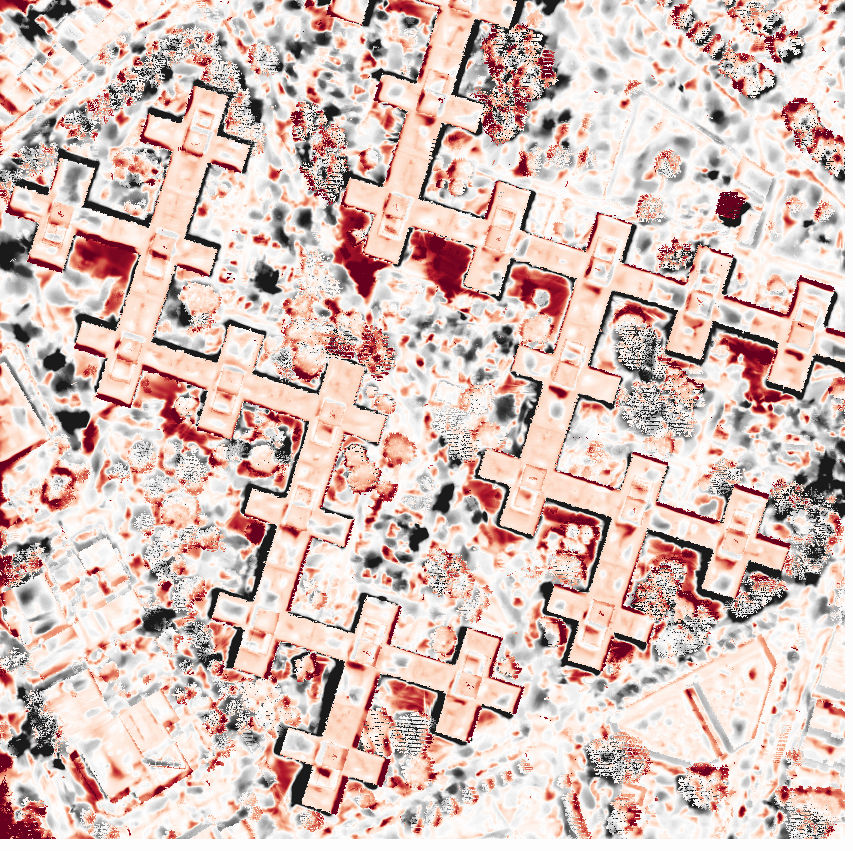} &
\includegraphics[width=0.11\linewidth, valign=m]{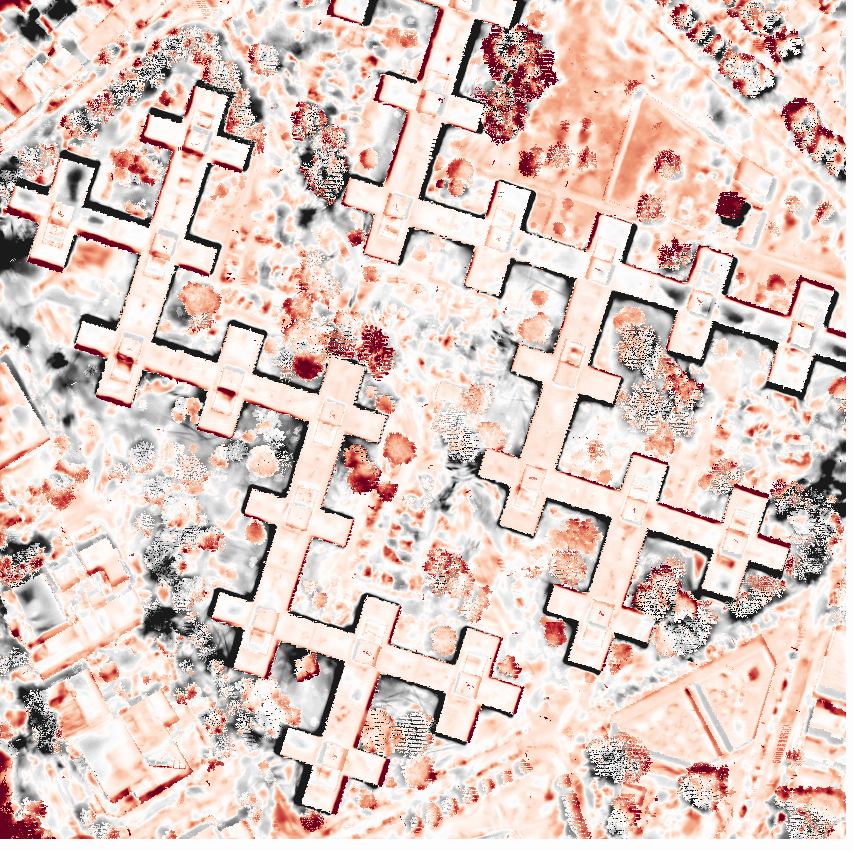} &
\includegraphics[width=0.11\linewidth, valign=m]{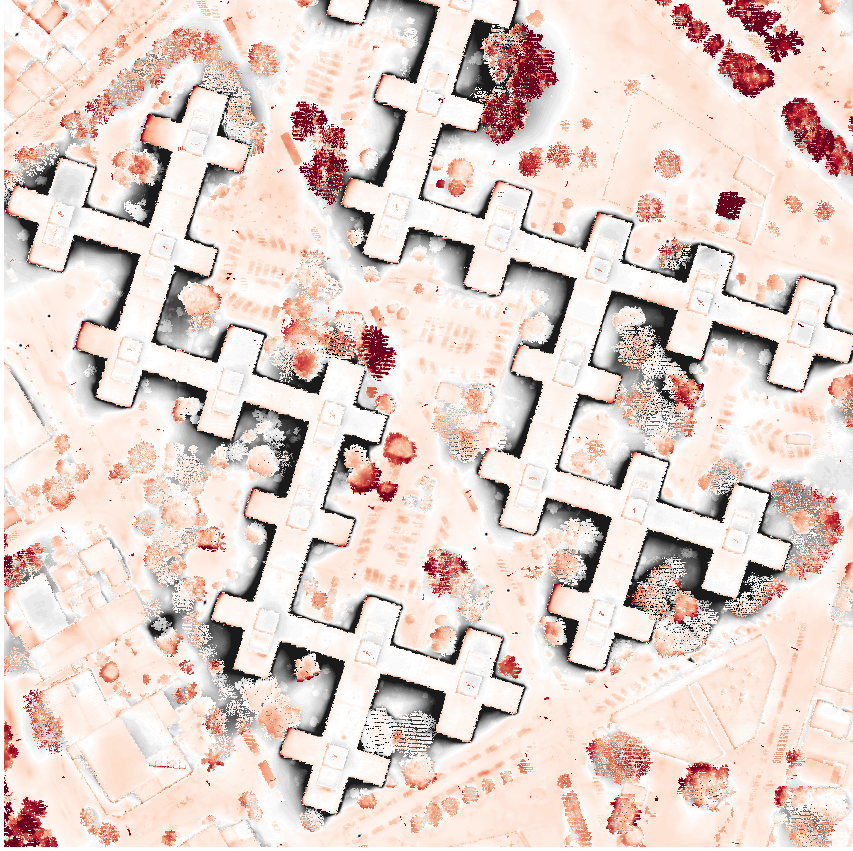} \\[2pt]

\leftlabelother{IARPA-003} &
\includegraphics[width=0.11\linewidth, valign=m]{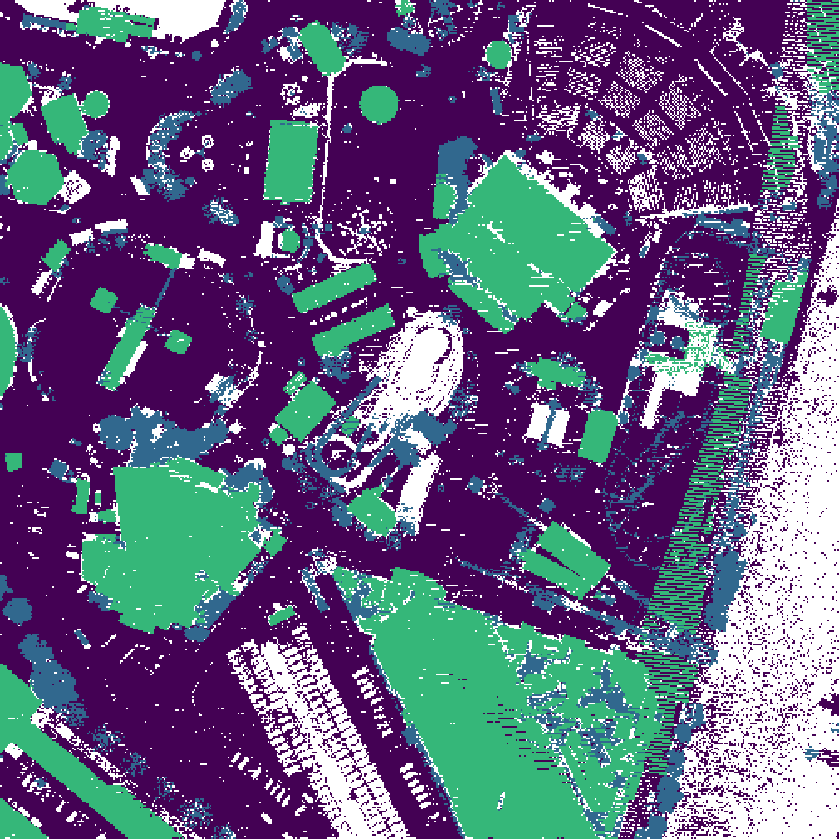} &
\includegraphics[width=0.11\linewidth, valign=m]{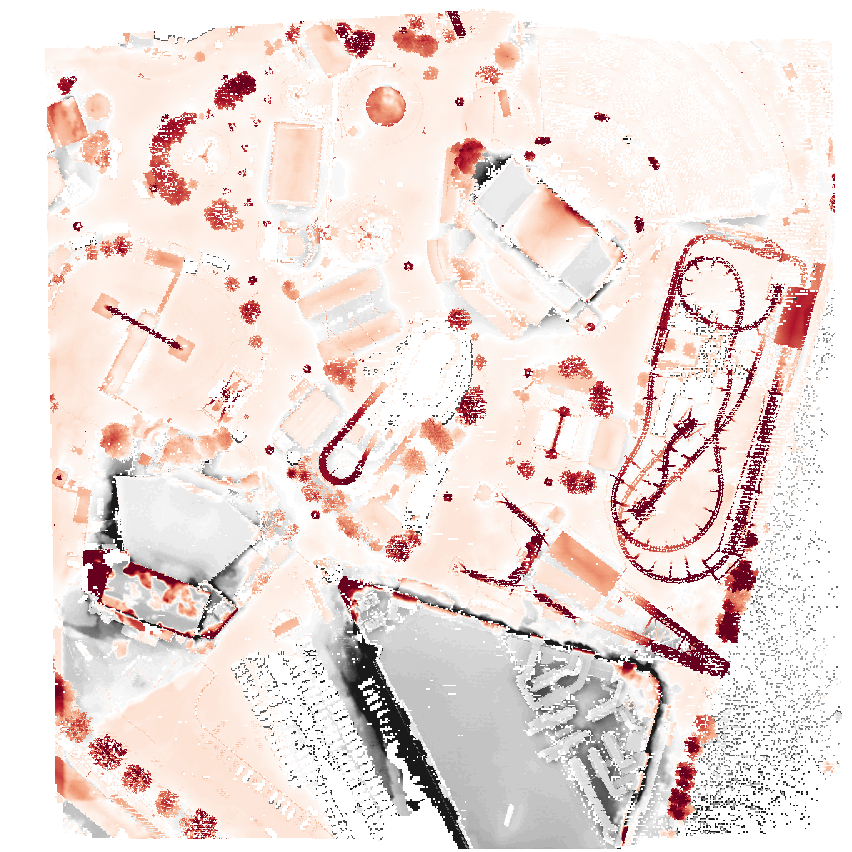} &
\includegraphics[width=0.11\linewidth, valign=m]{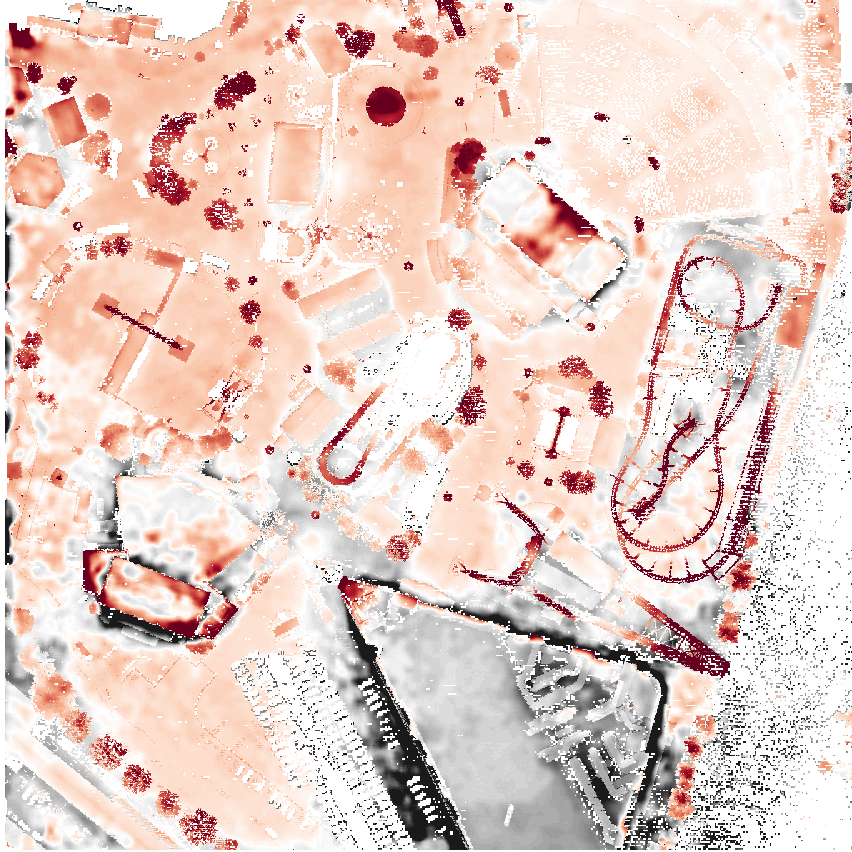} &
\includegraphics[width=0.11\linewidth, valign=m]{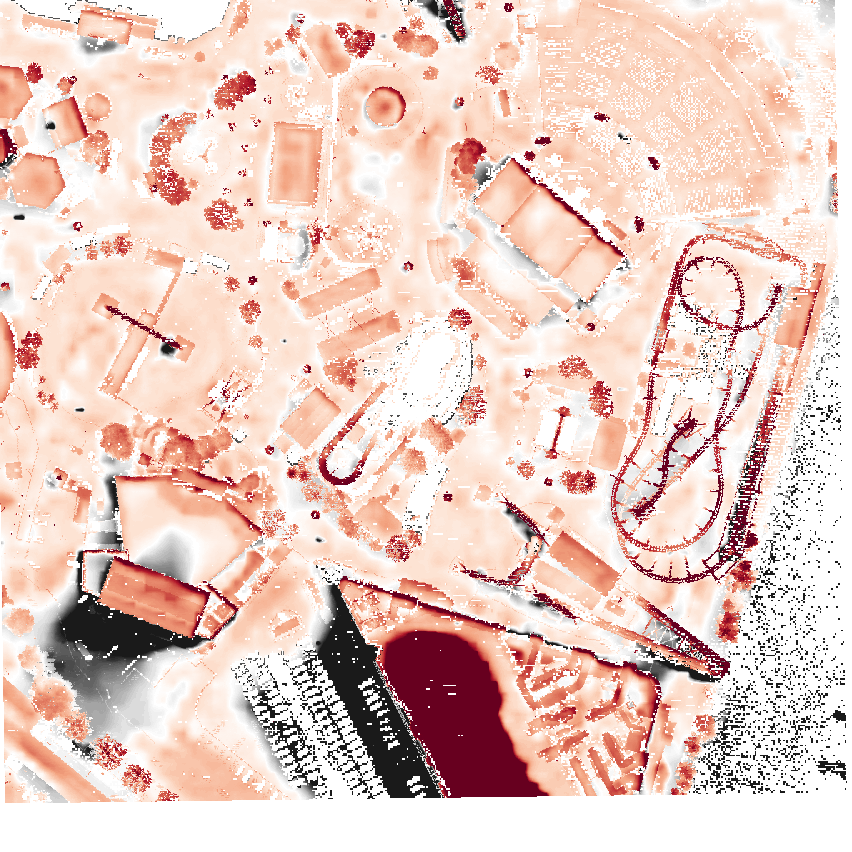} &
\includegraphics[width=0.11\linewidth, valign=m]{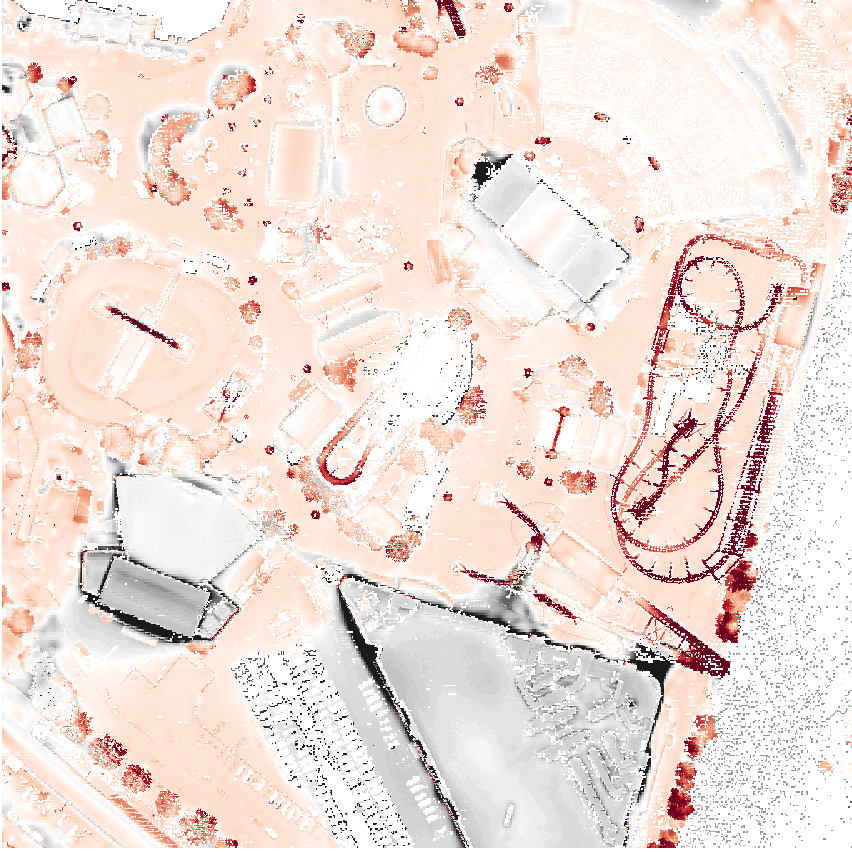} &
\includegraphics[width=0.11\linewidth, valign=m]{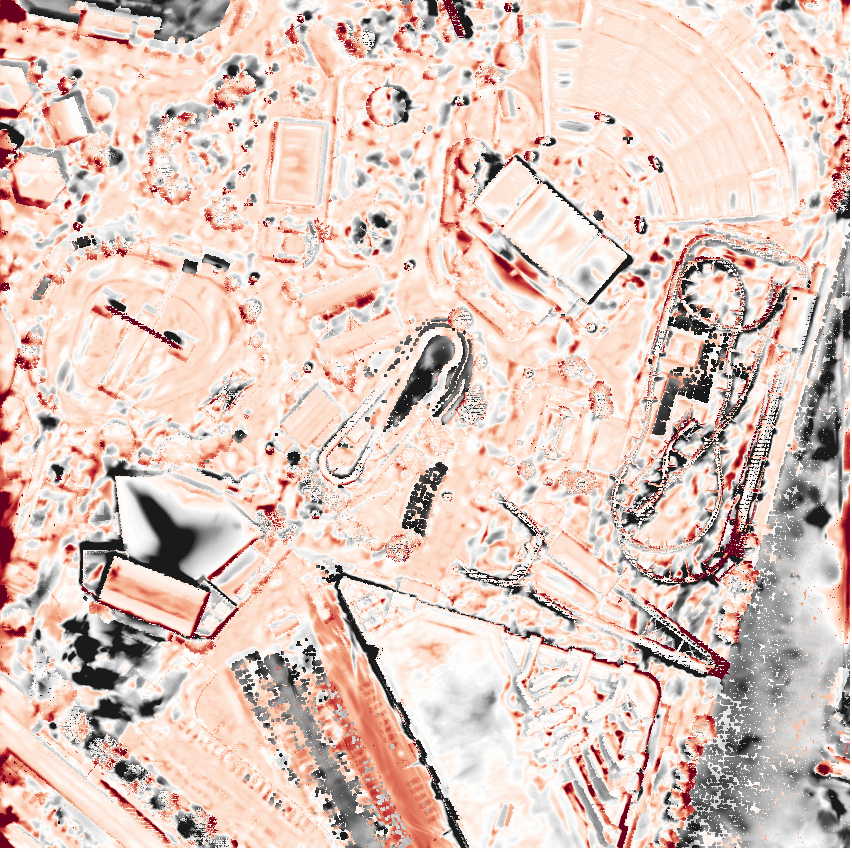} &
\includegraphics[width=0.11\linewidth, valign=m]{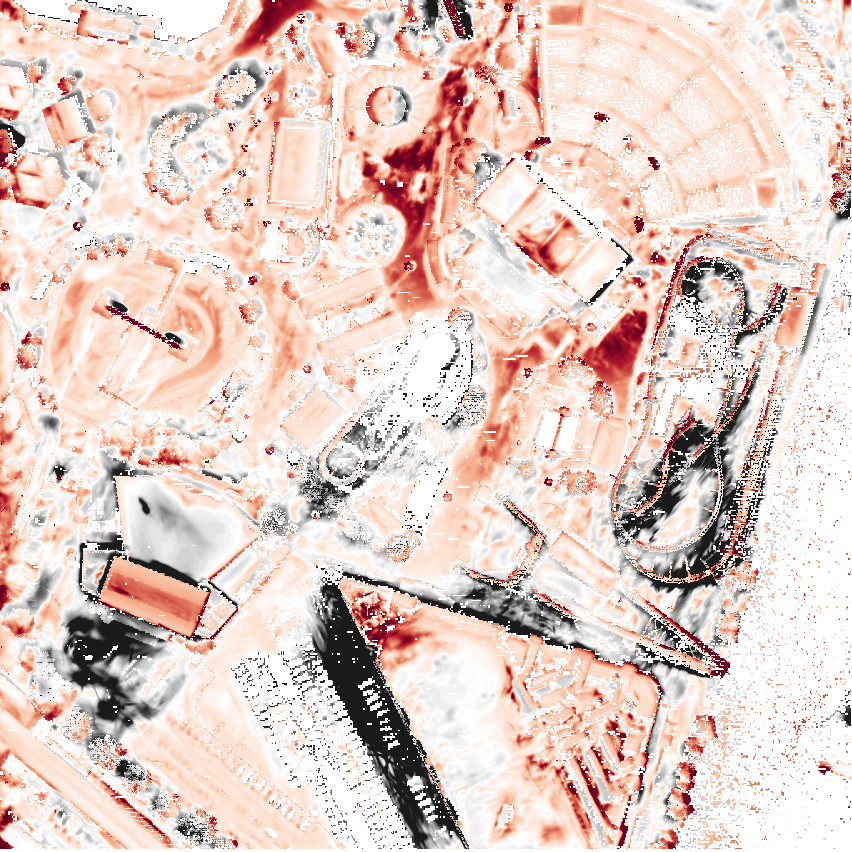} &
\includegraphics[width=0.11\linewidth, valign=m]{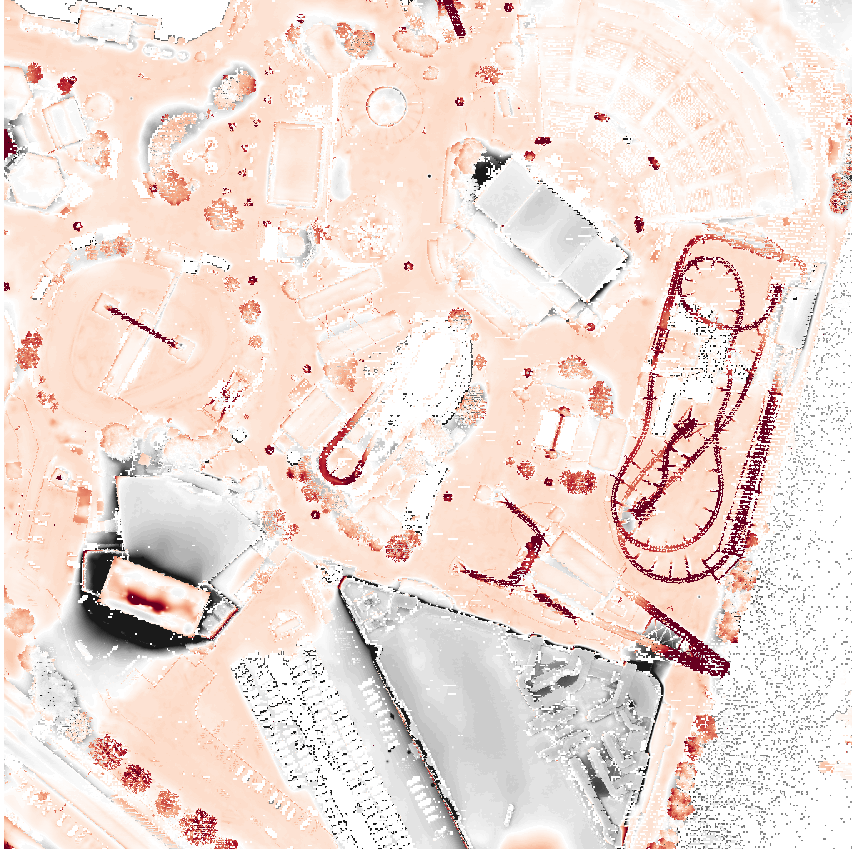} \\
& \small CLS & \small ASP & \small s2p & \small SAT-NGP & \small EOGS & \small Skyfall-1 & \small Skyfall-2 & \small SatSplat
\end{tabular}

\vspace{6pt}
\noindent
\begin{minipage}{0.15\linewidth}
\centering\includegraphics[width=\linewidth]{figure/err_map/CLS_ColorBar.png}
\end{minipage}
\hfill
\begin{minipage}{0.82\linewidth}
\caption{Errormap comparison for Omaha (OMA) and IARPA tiles. The errormap visualizes height residuals in the range of $-10$ to $10$\,m. For CLS, green indicates buildings, teal indicates vegetation, black represents ground, and white corresponds to water or regions with no ground truth altitude value.}
\label{fig:errmap_oma_iarpa}
\end{minipage}
\end{figure*}

In \cref{fig:errmap_jax,fig:errmap_oma_iarpa}, we provide extended qualitative comparisons across all JAX, OMA, and IARPA tiles in our benchmark. The errormap visualizations highlight that SatSplat consistently reduces large residuals and spurious artifacts (e.g., floaters and overly smoothed blobs) relative to competing methods, while better preserving sharp building outlines, road structures, and fine terrain detail.

% ==========================================================
% FIGURE 1: Jacksonville (JAX) Tiles
% ==========================================================

% ==========================================================
% FIGURE 2: Omaha (OMA) & IARPA Tiles
% ==========================================================

{\small
\bibliographystyle{ieeenat_fullname}
\bibliography{main,pers_wos_2024,pers_wos_2025}
}

\end{document}